\documentclass{IEEEoj}
\usepackage{cite}
\usepackage{amsmath,amssymb,amsfonts}
\usepackage{algorithmic}
\usepackage{graphicx,color}
\usepackage{textcomp}
\usepackage{amsmath,amsfonts}
\usepackage{algorithmic}
\usepackage{algorithm}
\usepackage{array}
\usepackage[caption=false,font=normalsize,labelfont=sf,textfont=sf]{subfig}
\usepackage{textcomp}
\usepackage{stfloats}
\usepackage{url}
\usepackage{verbatim}
\usepackage{svg}
\usepackage{tikz}
\usepackage{pgfplots}
\usepackage{pgfplotstable}
\pgfplotsset{compat=newest}
\usepackage[none]{hyphenat}
\usepackage{siunitx}
\usepackage[percent]{overpic}
\usepackage{hyperref}
\usepackage{makecell}
\usepackage{harveyballs}
\usepackage{cite}
\usepackage[acronym, nomain]{glossaries}
\newacronym{mlp}{MLP}{Multi-Layer Perceptron}
\newacronym{odd}{ODD}{Operational Design Domain}
\newacronym{nlp}{NLP}{Natural Language Processing}
\newacronym{fm}{FM}{Foundation Model}
\newacronym{llm}{LLM}{Large Language Model}
\newacronym{vlm}{VLM}{ Vision Language Model}
\newacronym{mllm}{MLLM}{Multimodal Large Language Model}
\newacronym{ccd}{CCD}{Car Crash Dataset}
\newacronym{fpv}{FPV}{First Person View}
\newacronym{dm}{DM}{Diffusion Model}
\newacronym{flops}{FLOPs}{Floating Point Operations Per Second}
\newacronym{wm}{WM}{World Model}
\newacronym{ml}{ML}{Machine Learning}
\newacronym{lstm}{LSTM}{Long Short-Term Memory}
\newacronym{gru}{GRU}{Gated Recurrent Unit}
\newacronym{svd}{SVD}{Stable Video Diffusion}
\newacronym{ad}{AD}{Autonomous Driving}
\newacronym{qa}{QA}{Question Answering}
\newacronym{ai}{AI}{Artificial Intelligence}
\newacronym{vqa}{VQA}{Visual Question Answering}
\newacronym{ros2}{ROS 2}{Robot Operating System}
\newacronym{adas}{ADAS}{Advanced Driver Assistance Systems}
\newacronym{dsl}{DSL}{domain-specific language}
\newacronym{ddpm}{DDPM}{Denoising Diffusion Probabilistic Model}
\newacronym{osm}{OSM}{OpenStreetMap}
\newacronym{vae}{VAE}{Variational Autoencoder}
\newacronym{ldm}{LDM}{Latent Diffusion Model}
\newacronym{vdm}{VDM}{Video Diffusion Model}
\newacronym{bev}{BEV}{bird's-eye view}
\newacronym{mdp}{MDP}{Markov Decision Process}
\newacronym{stl}{STL}{Signal Temporal Logic}
\newacronym{cot}{CoT}{Chain-of-Thought}
\newacronym{icl}{ICL}{In-Context Learning}
\newacronym{sc}{SC}{Self-Consistency}
\newacronym{rag}{RAG}{Retrieval-Augmented Generation}
\newacronym{cp}{CP}{Contextual Prompting}
\newacronym{fft}{FFT}{Full Fine-Tuning}
\newacronym{peft}{PEFT}{Parameter-Efficient Fine-Tuning}
\newacronym{lora}{LoRA}{Low-Rank Adaptation}
\newacronym{grpo}{GRPO}{Group Relative Policy Optimization}
\newacronym{mla}{MLA}{Multi-LLM Agent System}
\newacronym{ood}{OOD}{Out-of-Distribution}
\newacronym{sotif}{SOTIF}{Safety of the Intended Functionality}
\newacronym{vit}{ViT}{Vision Transformer}
\newacronym{clip}{CLIP}{Contrastive Language–Image Pre-training}
\newacronym{wd}{WD}{Wasserstein Distance}
\newacronym{qlora}{QLoRA}{Quantized Low-Rank Adaptation}
\newacronym{kld}{KLD}{Kullback–Leibler divergence}
\newacronym{dit}{DiT}{Diffusion Transformer}
\newacronym{hd}{HD}{High-Definition}
\newacronym{st}{ST}{Spatio-Temporal}
\newacronym{ma}{MA}{Modality Alignment}
\newacronym{qformer}{Q-Former}{Query Transformer}
\newacronym{sql}{SQL}{Structured Query Language}
\newacronym{mtr}{MTR}{Motion Transformer}
\newacronym{api}{API}{Application Programming Interface}
\newacronym{made}{mADE}{mean Average Displacement Error}
\newacronym{mfde}{mFDE}{mean Final Displacement Error}
\newacronym{mmd}{MMD}{Maximum Mean Discrepancy}
\newacronym{ttc}{TTC}{Time-to-Collision}
\newacronym{bleu}{BLEU}{Bilingual Evaluation Understudy}
\newacronym{cider}{CIDEr}{Consensus-based Image Description Evaluation}
\newacronym{meteor}{METEOR}{Metric for Evaluation of Translation with Explicit ORdering}
\newacronym{rougel}{ROUGE-L}{Recall-Oriented Understudy for Gisting Evaluation - Longest common subsequence}
\newacronym{sspd}{SSPD}{Symmetric Segment-Path Distance}
\newacronym{fid}{FID}{Frechet Inception Distance}
\newacronym{rmse}{RMSE}{Root Mean Squared Error}
\newacronym{fvd}{FVD}{Frechet Video Distance}
\newacronym{kvd}{KVD}{Kernel Video Distance}
\newacronym{iou}{IoU}{Intersection-over-Union}
\newacronym{miou}{mIoU}{mean Intersection-over-Union}
\newacronym{map}{mAP}{mean Average Precision}
\newacronym{po}{PO}{Preference Optimization}
\newacronym{rnn}{RNN}{Recurrent Neural Network}
\usepackage[table]{xcolor}
\usepackage{booktabs} 
\usepackage{multirow}
\usepackage{soul}
\usepackage{microtype} % used to safe space

\renewcommand{\arraystretch}{1.4}

\usepackage{graphicx}
\definecolor{Gray}{gray}{0.7}
\definecolor{LightGray}{gray}{0.92}
\definecolor{LightGreen}{RGB}{0, 111, 0}
\definecolor{LightRed}{RGB}{200, 0, 0}

\usepackage{amssymb}
\usepackage{threeparttable}

\newcommand{\cmark}{\checkmark} % checkmark
\newcommand{\xmark}{} % cross

\newcommand{\fullcirc}{\harveyBallFull}
\newcommand{\halfcirc}{\harveyBallHalf}
\newcommand{\emptycirc}{\harveyBallNone}

\newcommand{\image}{\includegraphics[height=0.35cm]{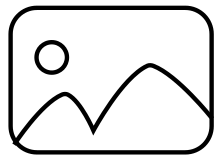}}
\newcommand{\video}{\includegraphics[height=0.35cm]{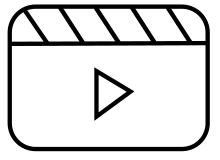}}
\newcommand{\trajectory}{\includegraphics[height=0.35cm]{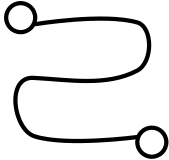}}
\newcommand{\script}{\includegraphics[height=0.35cm]{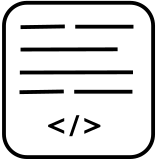}}

\def\BibTeX{{\rm B\kern-.05em{\sc i\kern-.025em b}\kern-.08em
    T\kern-.1667em\lower.7ex\hbox{E}\kern-.125emX}}
\usepackage{balance}
\usepackage{xcolor}

\newcommand{\cc}{\cellcolor{LightGray}}

\usepackage{scalerel}
\usepackage{tikz}
\usetikzlibrary{svg.path}
\definecolor{orcidlogocol}{HTML}{A6CE39}
\tikzset{
  orcidlogo/.pic={
    \fill[orcidlogocol] svg{M256,128c0,70.7-57.3,128-128,128C57.3,256,0,198.7,0,128C0,57.3,57.3,0,128,0C198.7,0,256,57.3,256,128z};
    \fill[white] svg{M86.3,186.2H70.9V79.1h15.4v48.4V186.2z}
                 svg{M108.9,79.1h41.6c39.6,0,57,28.3,57,53.6c0,27.5-21.5,53.6-56.8,53.6h-41.8V79.1z M124.3,172.4h24.5c34.9,0,42.9-26.5,42.9-39.7c0-21.5-13.7-39.7-43.7-39.7h-23.7V172.4z}
                 svg{M88.7,56.8c0,5.5-4.5,10.1-10.1,10.1c-5.6,0-10.1-4.6-10.1-10.1c0-5.6,4.5-10.1,10.1-10.1C84.2,46.7,88.7,51.3,88.7,56.8z};
  }
}
\newcommand\orcidicon[1]{\href{https://orcid.org/#1}{\mbox{\scalerel*{
\begin{tikzpicture}[yscale=-1,transform shape]
\pic{orcidlogo};
\end{tikzpicture}
}{|}}}}

\usepackage{hyperref} %<--- Load after everything else
\newcommand{\rev}[1]{\textcolor{black}{#1}}
\newcommand{\revB}[1]{\textcolor{black}{#1}}

\def\BibTeX{{\rm B\kern-.05em{\sc i\kern-.025em b}\kern-.08em
    T\kern-.1667em\lower.7ex\hbox{E}\kern-.125emX}}
\AtBeginDocument{\definecolor{ojcolor}{cmyk}{0.93,0.59,0.15,0.02}}
\def\OJlogo{}% No logo for arXiv version
\makeatletter
\gdef\@IEEEacceptancenotice{%
  \begin{center}
  \fbox{\parbox{0.85\linewidth}{\centering\small\normalfont
  This paper has been accepted for publication at IEEE Open Journal of Intelligent Transportation Systems 2026 \textcopyright\,IEEE
  \par}}\\[0.8em]
  \end{center}}
\makeatother
\begin{document}

\title{Foundation Models in Autonomous Driving: A Survey on Scenario Generation and Scenario Analysis}

\author{Yuan Gao$^{\orcidicon{0009-0004-9158-7202}1}$, Mattia Piccinini$^{\orcidicon{0000-0003-0457-8777}1}$, Yuchen Zhang$^{\orcidicon{0009-0003-0800-9878}1}$, Dingrui Wang$^{\orcidicon{0009-0003-7546-2226}1}$,\\ Korbinian Moller$^{\orcidicon{0000-0001-7120-0796}1}$, Roberto Brusnicki$^{\orcidicon{0000-0001-7223-9895}1}$, Baha Zarrouki$^{\orcidicon{0009-0007-1071-4487}1}$, Alessio Gambi$^{\orcidicon{0000-0002-0132-6497}2}$,\\Jan Frederik Totz$^{\orcidicon{0000-0003-4961-1630}3}$, Kai Storms$^{\orcidicon{0000-0002-9551-5466}4}$, Steven Peters$^{\orcidicon{0000-0003-3131-1664}4}$, Andrea Stocco$^{\orcidicon{0000-0001-8956-3894}5}$,\\ Bassam Alrifaee$^{\orcidicon{0000-0002-5982-021X}6}$, Marco Pavone$^{\orcidicon{0000-0002-0206-4337}7}$, Johannes Betz$^{~\orcidicon{0000-0001-9197-2849}1}$
% <-this % stops a space

\affil{Y. Gao,  M. Piccinini, Y. Zhang, D. Wang, K. Moller, R. Brusnicki, B. Zarrouki, and J. Betz are with the Professorship of Autonomous Vehicle Systems and Munich Institute of Robotics and Machine Intelligence (MIRMI), Technical University of Munich, 85748 Garching, Germany  (e-mail: johannes.betz@tum.de)}
\affil{A. Gambi is with the Austrian Institute of Technology (AIT), Vienna, Austria (e-mail: alessio.gambi@ait.ac.at)}%
\affil{J.F. Totz is with AUDI AG, Germany (e-mail: jan.frederik.totz@audi.de)}
\affil{K. Storms, S. Peters are with the Institute of Automotive Engineering (FZD) at TU Darmstadt, 64289 Darmstadt, Germany (e-mail: steven.peters@tu-darmstadt.de)}
\affil{A. Stocco is with the Technical University of Munich and fortiss GmbH, Munich, Germany (e-mail: andrea.stocco@tum.de|stocco@fortiss.org)}
\affil{B. Alrifaee is with the Department of Aerospace Engineering, University of the Bundeswehr
Munich, Germany, (e-mail: bassam.alrifaee@unibw.de)}
\affil{M. Pavone is with Stanford University and Nvidia Research, Stanford, CA
94305, USA (e-mail: pavone@stanford.edu)}
}

\corresp{CORRESPONDING AUTHOR: Yuan Gao (e-mail: yuan\_avs.gao@tum.de).}
% \authornote{This work was supported by the Natural Sciences and Engineering Research Council (NSERC) of Canada.}
\markboth{Foundation Models in Autonomous Driving: A Survey on Scenario Generation and Scenario Analysis}{Y. Gao \textit{et al.}}

\begin{abstract}
\rev{Ensuring the safety of autonomous vehicles in real-world environments requires handling a wide spectrum of diverse and rare driving scenarios. Scenario-based testing addresses this need by offering a scalable and controlled approach to develop and validate autonomous driving systems. However, traditional scenario generation methods relying on rule-based logic, knowledge-driven models, or data-driven synthesis often yield limited diversity and unrealistic cases.
With the emergence of foundation models, which represent a new generation of pre-trained, general-purpose \gls{ai} models, developers can process heterogeneous inputs (e.g., natural language, sensor data, maps, and control actions), enabling the synthesis, interpretation, analysis of complex driving scenarios.
In this paper, we review the use of foundation models for scenario generation and scenario analysis in autonomous driving.
Our survey presents a unified taxonomy that includes large language models, vision language models, multimodal large language models, diffusion models, and world models for the generation and analysis of autonomous driving scenarios, outlining their fundamental principles, applications, and corresponding evaluation metrics. In addition, we review the methodologies, open-source datasets, simulation platforms, and benchmark challenges. Finally, the survey concludes by highlighting the open challenges, research questions and promising future directions in applying foundation models to scenario generation and analysis in autonomous driving. 
All reviewed papers are listed in a continuously maintained repository, which is publicly available and updated with new research: \href{https://github.com/TUM-AVS/FM-for-Scenario-Generation-Analysis}{GitHub.com/TUM-AVS/FM-for-Scenario-Generation-Analysis}.}

% \textcolor{red}{The safety of autonomous vehicles depends on their ability to navigate diverse and complex driving scenarios. While real-world testing is often prohibitively costly and unsafe, scenario based testing provides a scalable and controlled alternative. However, traditional rule-based  methods struggle to produce the diversity and realism required for comprehensive validation. The emergence of Foundation Models (FMs) now offers a paradigm shift, enabling data-driven approaches that can generate complex scenarios and provide deeper semantic analysis. In this paper, we conduct a survey about the application of FMs to the dual pillars of scenario-based testing: \textbf{scenario generation}, which focuses on creating realistic and controllable driving scenarios, and \textbf{scenario analysis}, which concerns the evaluation of these scenarios in terms of safety, risk, and behavioral understanding. We review five major categories of FMs, including Large Language Models (LLMs), Vision-Language Models (VLMs), Multimodal Large Language Models (MLLMs), Diffusion Models (DMs), and World Models (WMs), outlining their fundamental principles, applications, and corresponding evaluation metrics. The survey concludes with a discussion of open challenges and future research directions. All papers reviewed are cataloged in a continuously maintained GitHub Repository, which is publicly available and updated with new research: \href{https://github.com/TUM-AVS/FM-for-Scenario-Generation-Analysis}{GitHub.com/TUM-AVS/FM-for-Scenario-Generation-Analysis}.}

\end{abstract}


\maketitle

\section{Introduction}
\label{sec:introduction}

\IEEEPARstart{A}{utonomous driving} has seen rapid advancements in recent years, reaching a stage where human intervention is minimal or entirely unnecessary within specific \glspl{odd}~\cite{Betz2024}. 
Companies such as Waymo have successfully deployed fully autonomous robotaxi services operating at SAE Level 4~\cite{SAE2021} since 2018, demonstrating the feasibility of driverless mobility in specific urban environments. 
As of 2025, Waymo serves $250,000$ commercial rides per week~\cite{cnbc_waymo}. These advancements have been driven by the development and rigorous validation of highly reliable modular \gls{ad} software functions, including perception, prediction, planning, and control~\cite{pendleton2017perception}.
In addition to the traditional modular architecture, end-to-end learning-based approaches~\cite{betz2022autonomous, chen2024end} have emerged, leveraging deep learning to process raw sensor data and directly generate trajectories or control actions~\cite{Zheng2024}. 
\begin{figure}[!t]
    \centering
    \includegraphics[width=0.95\linewidth]{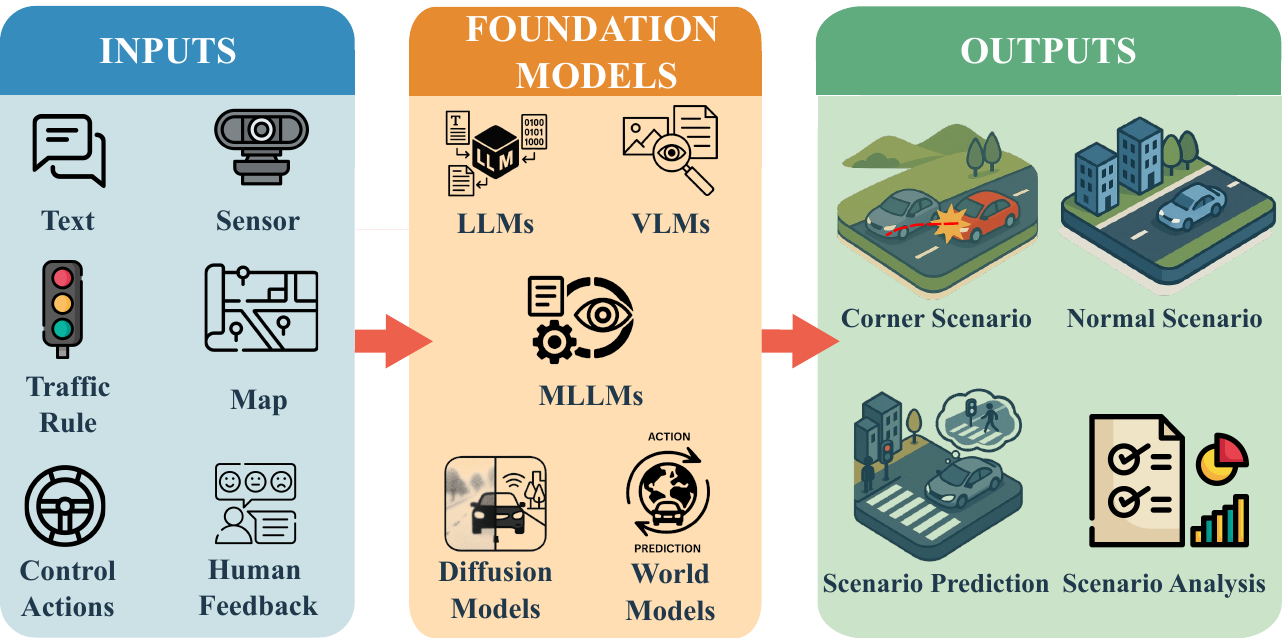}
    \caption{This survey critically analyzes existing FMs across LLMs, VLMs, MLLMs, DMs, and WMs for scenario generation and scenario analysis in autonomous driving.}
    \label{fig:enter-label_b}
\end{figure}

Scenario-based testing in simulations is a key element for evaluating and validating the safety and performance of \gls{ad} systems~\cite{menzel2018scenarios}. As a cost-efficient alternative to physical testing, it enables the simulation of realistic, reproducible, and controllable driving environments~\cite{riedmaier2020survey}, and is particularly effective in replicating safety-critical situations, including rare corner cases that are often absent in real-world datasets~\cite{gambi2022generating, karle2022scenario}. 
Therefore, the ability to systematically generate and analyze driving scenarios is crucial to scenario-based testing \cite{Cherubini2024}. \rev{More specifically, generation focuses on creating diverse, safety-critical, and controllable driving situations for \gls{ad} testing, while analysis concerns the evaluation of these scenarios in terms of safety, risk, and behavioral understanding to classify or select scenarios for testing, or to enhance overall \gls{ad} performance.}

Recent advances in \gls{ml}, especially the emergence of large-scale \glspl{fm}, offer new opportunities to enhance the realism, diversity, and scalability of scenario-based testing in autonomous driving.
\glspl{fm} were introduced by the Stanford Institute for Human-Centered Artificial Intelligence (HAI) in August 2021~\cite{bommasani2021opportunities} to describe a class of models trained on large-scale, diverse datasets typically using self-supervised learning. Unlike traditional \gls{ml} models, which are often trained for specific, narrowly defined tasks, \glspl{fm} can be adapted to a wide range of tasks through techniques such as prompting or fine-tuning. These models have demonstrated strong performance across various domains, including \gls{nlp}~\cite{devlin2019bert}, visual understanding~\cite{dosovitskiy2020image}, and code generation~\cite{jiang2024survey}. In the context of autonomous driving, \glspl{fm} have recently garnered significant attention, as they combine general knowledge learned through large-scale pre-training with efficient adaptability to specific \gls{ad} tasks \rev{like perception, palnning, control ~\cite{huang2023applications, gao2024survey, wang2025dualad, wang2025generative}}.

\begin{figure*}
    \centering
    \newcommand{\plotheight}{5.5cm}
\newcommand{\plotwidth}{15.0cm}
\newcommand{\enlargexlimits}{0.025}

\begin{tikzpicture}[font=\footnotesize]

% Blue – Computer Vision
\definecolor{ADColor}{HTML}{0065bd}
\definecolor{CVColor}{HTML}{98c6ea}
\definecolor{MLColor}{HTML}{e37222}
\definecolor{POColor}{HTML}{999999}
\definecolor{ROColor}{HTML}{a2ad00}

\begin{axis}[
    width=\plotwidth,
    height=\plotheight,
    ybar stacked,
    bar width=4pt,
    enlarge x limits=\enlargexlimits,
    xtick=data,
    xticklabel style={rotate=60, anchor=east},
    xticklabels from table={figs/LLM_Timeline_data.csv}{Date},
    ylabel={Number of Papers},
    xlabel={Publication Month},
    legend style=
    {
        at={(0.5,-0.42)}, 
        anchor=north,
        legend columns=2,
        cells={anchor=west},
        draw=none,
        column sep=0.4em,
        row sep=0.1em,
        %draw
    },
    table/col sep=semicolon,
    axis y line*=left,
    ymin=0
]

\addplot [fill=ADColor, draw=none] table [x expr=\coordindex, y=AutonomousDriving, meta=Date] {figs/LLM_Timeline_data.csv};
\addlegendentry{\textbf{Autonomous Driving:} AIAT, IAVVC, ITSC, IV, T-IV, T-SMC}

\addplot [fill=CVColor, draw=none] table [x expr=\coordindex, y=ComputerVision, meta=Date] {figs/LLM_Timeline_data.csv};
\addlegendentry{\textbf{Computer Vision:} CVPR, ECCV, ICCV, ICJV, ICME, ICPR, WACV}

\addplot [fill=ROColor, draw=none] table [x expr=\coordindex, y=Robotics, meta=Date] {figs/LLM_Timeline_data.csv};
\addlegendentry{\textbf{Robotics:} CoRL, ICRA, IROS, RAL, RSS, UR}

\addplot [fill=POColor, draw=none] table [x expr=\coordindex, y=PreprintsOthers, meta=Date] {figs/LLM_Timeline_data.csv};
\addlegendentry{\textbf{Preprints \& Others:} arXiv, MDPI, ELECO, IEICE, SSRN}

\addplot [fill=MLColor, draw=none] table [x expr=\coordindex, y=MachineLearning, meta=Date] {figs/LLM_Timeline_data.csv};
\addlegendentry{\textbf{Machine Learning:}} %AAAI, AIAHPC, ASE, FLLM, FORGE, ICLR, ICML, ICTAI, NeurIPS}

\end{axis}

% Right axis and data
\pgfplotstableread{
Date       Cumulative
2023-05	   4
2023-06	   5
2023-07	   6
2023-08	   7
2023-09	   13
2023-10	   17
2023-11	   27
2023-12	   31
2024-01	   34
2024-02	   38
2024-03	   41
2024-04	   44
2024-05	   50
2024-06	   56
2024-07	   60
2024-08	   65
2024-09	   73
2024-10	   75
2024-11    86
2024-12	   95
2025-01    103
2025-02	   112
2025-03	   127
2025-04	   134
2025-05	   143
}\cumulativeData

\begin{axis}[
    height=\plotheight,
    width=\plotwidth,
    symbolic x coords={
        2023-05,2023-06,2023-07,2023-08,2023-09,2023-10, 2023-11,2023-12,2024-01,2024-02,2024-03,2024-04,2024-05,2024-06,2024-07,2024-08,2024-09,2024-10,2024-11,2024-12,2025-01,2025-02,2025-03,2025-04,2025-05},
    axis x line=none,
    axis y line*=right,
    ymin=0,
    ymax=160,
    ylabel={Cumulative Paper Count},
    enlarge x limits=\enlargexlimits,
]

\addplot+[black, thick, mark=*, ,mark options={fill=black,scale=0.75}] table [x=Date, y=Cumulative] {\cumulativeData};

\end{axis}

\node[anchor=west, font=\footnotesize, align=left] at (1.2, -2.83) 
{AAAI, AIAHPC, ASE, FLLM, FORGE, ICLR, ICML, ICTAI, NeurIPS};

\end{tikzpicture}
    \vspace{-0.2cm}
    \caption{Timeline of foundation model-related publications in scenario generation and analysis, across selected journals, conferences and platforms between May 2023 and May 2025. Each bar represents the monthly count of papers, grouped by thematic category. The black line shows the cumulative number of papers over time (right axis). Note that the grouping refers to the general scope of the conference or journal, not to the content of the individual papers. For example, preprints listed on \textit{arXiv} are categorized under \textit{Preprints \& Others}, although they may address topics from other categories.}
    \label{fig:enter-label}
\end{figure*}
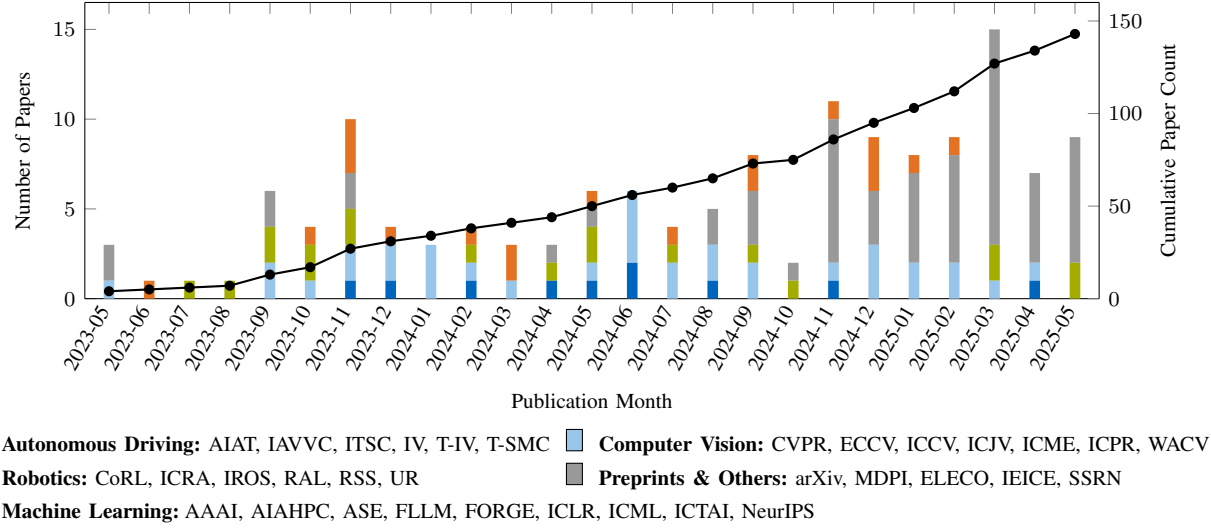

\subsection{Scope of the Considered Literature}

In this survey, we focus on publications addressing \textit{scenario generation} and \textit{scenario analysis} in the context of autonomous driving with Foundation Models (see \autoref{fig:enter-label_b}). 
Our survey is based on keyword searches in Google Scholar. The full list of keywords, as well as an overview of all referenced papers, is available \rev{ in the GitHub repository of this paper}\footnote{\href{https://github.com/TUM-AVS/FM-for-Scenario-Generation-Analysis}{https://github.com/TUM-AVS/FM-for-Scenario-Generation-Analysis}}.

To ensure both breadth and relevance, we included peer-reviewed conference and journal papers as well as preprints from arXiv. Although arXiv publications are not formally peer-reviewed, they often present timely and impactful research, especially in rapidly developing areas such as \gls{fm} applications.
Our survey covers papers published between October 2022 and May 2025, with a primary focus on venues in autonomous driving, computer vision, machine learning, and robotics. 
\autoref{fig:enter-label} shows monthly trends in publication counts and their distribution by the thematic focus of the publication venues, e.g., conferences, journals, or preprint platforms.
%
%The complete list of publication venues for each paper, along with the corresponding open-source code (when available), is included in the paper's GitHub repository.

\subsection{Structure of the Survey}

\begin{figure*}[ht]
    \centering
    \begin{overpic}[width=0.98\textwidth]{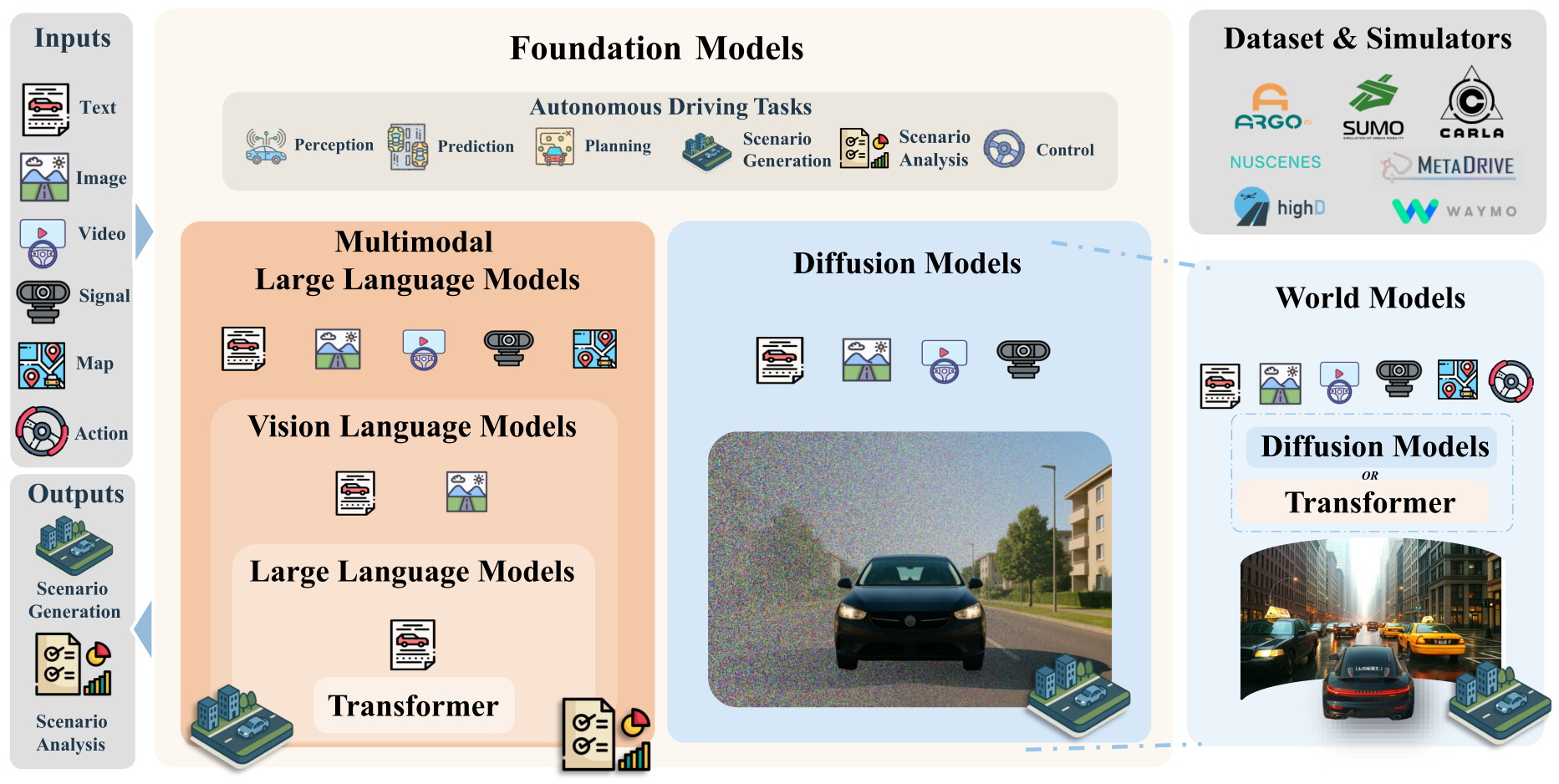}
        % Add clean text labels (linked to sections)
        \put(38.3,44.8){\hyperlink{sec:fm}{\textcolor{black}{\small \textbf{~Section II}~\includegraphics[height=1.5ex]{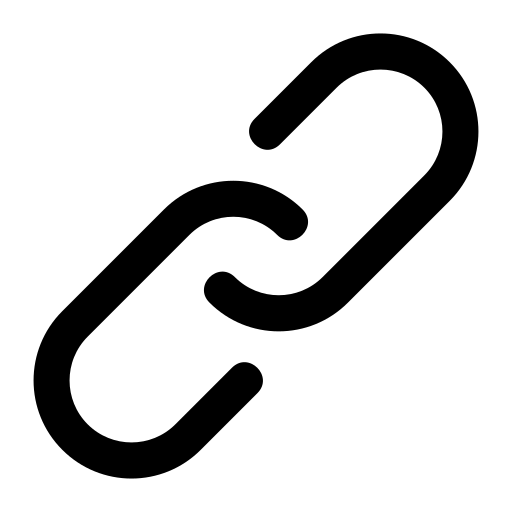}}}}       % Foundation Models
        \put(21.5,11.3){\hyperlink{sec:llm}{\textcolor{black}{\small \textbf{~Section III}~\includegraphics[height=1.5ex]{figs/icons/hyperlink.png}}}}     % LLMs
        \put(21.5,20.5){\hyperlink{sec:vlm}{\textcolor{black}{\small \textbf{~Section IV}~\includegraphics[height=1.5ex]{figs/icons/hyperlink.png}}}}     % VLMs
        \put(21.5,29.7){\hyperlink{sec:mllm}{\textcolor{black}{\small \textbf{~Section V}~\includegraphics[height=1.5ex]{figs/icons/hyperlink.png}}}}    % MLLMs
        \put(52.6,29.5){\hyperlink{sec:dm}{\textcolor{black}{\small \textbf{~Section VI}~\includegraphics[height=1.5ex]{figs/icons/hyperlink.png}}}}      % Diffusion
        \put(82.2,28.3){\hyperlink{sec:wm}{\textcolor{black}{\small \textbf{~Section VII}~\includegraphics[height=1.5ex]{figs/icons/hyperlink.png}}}}      % World Models
        \put(82.2,45.6){\hyperlink{sec:dataset}{\textcolor{black}{\small \textbf{~Section VIII}~\includegraphics[height=1.5ex]{figs/icons/hyperlink.png}}}} % Datasets & Simulators
    \end{overpic}
    \caption{Overview of \glspl{fm} applied to scenario generation and analysis for autonomous driving, and the corresponding structure of this survey.}
    \label{fig:FMs}
\end{figure*}

The structure of this survey is outlined in \autoref{fig:FMs}. Section~\ref{sec:background} provides an introduction to Foundation Models and a critical review of related surveys on scenario generation and analysis, encompassing both classical approaches and recent advances with Foundation Models. Sections~\ref{sec:llm}, \ref{sec:vlm}, and \ref{sec:mllm} systematically examine language-based Foundation Models, beginning with fundamental concepts and followed by an in-depth discussion of recent applications of \glspl{llm}, \glspl{vlm}, and \glspl{mllm} in scenario generation and analysis. Sections~\ref{sec:dm} and \ref{sec:wm} address vision-centric Foundation Models, detailing the principles of \glspl{dm} and \glspl{wm} and their relevance to scenario generation. Section~\ref{sec:dataset} surveys \rev{commen evaluation metrics},  publicly available datasets, and simulation benchmarks pertinent to scenario generation and analysis in autonomous driving, and highlights major competition challenges in the field. Finally, Section~\ref{sec:discussion} and Section~\ref{sec:directions} identify open research questions and outline prospective research directions, while Section~\ref{sec:conclusion} summarizes the key findings of this survey.
\hypertarget{sec:fm}{}
\section{Related Work \& Contributions}
\label{sec:background}

\subsection{Development of Foundation Models}

% \commAS{this section is too generic; I would drop it, or shorten it substantially to keep the paper focused on FMs for AD, as the paper is already very long}

\rev{%As demonstrated through a wide array of generative applications, foundation models have shown considerable flexibility. 
The term \textit{Foundation Models}, introduced in 2021~\cite{bommasani2021opportunities}, refers to general-purpose models trained on large-scale unlabeled data, designed to operate and generalize across a wide range of applications. Their cross-modal adaptability has the potential to enable tasks like \gls{qa}, image captioning, sentiment analysis, information extraction, object recognition, and instruction following, combining generative abilities with deep contextual understanding. Although \glspl{fm} and generative AI are related, they represent distinct concepts: \glspl{fm} are broad, adaptable systems, whereas generative AI focuses specifically on content creation.}

\rev{In 2020, OpenAI released GPT-3~\cite{brown2020language}, a major milestone that popularized \textbf{\glspl{llm}}. The success of GPT-3 built upon the Transformer Architecture~\cite{vaswani2017attention}, whose self-attention mechanism facilitated efficient parallel training on long sequences. Subsequent models further refined this design, including BERT~\cite{devlin2019bert} (encoder-only for masked language modeling), GPT~\cite{radford2018improving} (decoder-only for autoregressive generation), and T5~\cite{raffel2020exploring} (encoder-decoder for text-to-text transfer). Each employs self-supervised pre-training and serves as a backbone for downstream adaptation.}

\rev{The principles of the transformer architecture were quickly} extended beyond \gls{nlp}, and enabled visual understanding~\cite{dosovitskiy2020image}, speech~\cite{liu2020mockingjay}, tabular~\cite{yin2020tabert}, and multimodal learning~\cite{luo2020univl}. The extension of transformer architectures across different domains led to the development of \textbf{\glspl{vlm}} such as \gls{clip}~\cite{radford2021clip} and \textbf{\glspl{mllm}} such as LLaVA~\cite{liu2023visual}, that combine linguistic reasoning from \glspl{llm} with visual representations to produce cross-modal alignment and grounded understanding. \rev{More specifically, compact \glspl{vlm} typically use an \gls{llm} as a backbone, augmented with a text tokenizer and a dedicated vision encoder to extract visual features. \glspl{mllm} further extend this paradigm by incorporating additional modality-specific encoders, such as audio, depth, or sensory inputs, and employ alignment modules to fuse these heterogeneous representations with the \gls{llm} backbone. The strong reasoning capability inherited from \glspl{llm} enables these models to perform complex multimodal inference, while the added visual and sensory encoders allow \glspl{vlm} and \glspl{mllm} to operate effectively in real-world settings, where understanding both linguistic instructions and perceptual inputs is essential.}

\rev{At the same time, advances in generative modeling developed \textbf{\glspl{dm}}, and specifically \glspl{ddpm}~\cite{ho2020denoising} that generate high-quality samples using learned denoising processes. Subsequent variants, including improved \glspl{ddpm}~\cite{song2020denoising}, \glspl{ldm}~\cite{rombach2022high}, and \glspl{vdm}~\cite{ho2022video}, further enhanced generation efficiency, controllability, and temporal coherence. Extending beyond image synthesis, \glspl{dm} naturally support multimodal conditioning, enabling text-, audio-, and video-guided generation. Their high generation fidelity and flexible conditioning mechanisms make them powerful complements to transformer-based architectures, particularly in multimodal learning and world-modeling applications.}

\rev{Finally, \textbf{\glspl{wm}}~\cite{ha2018world} were developed to learn compact representations of interactive environments. Classical \glspl{wm} combine vision encoders (e.g., \glspl{vae}) with recurrent memory modules (e.g., \glspl{rnn}) and lightweight controllers (e.g., Evolution Strategies) to enable \emph{future prediction}, such as forecasting video frames or rolling out latent state trajectories. Recent \glspl{wm} designs integrate \glspl{fm} into their components, e.g., by replacing encoders or memory modules with \glspl{dm}~\cite{wang2024drivedreamer} or \glspl{llm}~\cite{DriveDreamer2_2024}, thus potentially unifying perception, reasoning, and generation in the same framework. Overall, this inter-connected evolution represents a progression from modality-specific \glspl{fm} to general multimodal systems for holistic environmental understanding and interaction.}

%

% \begin{figure*}[ht]
% \centering
% \begin{overpic}[width=0.85\textwidth]{figs/Figure3_flow.pdf}
%         \put(1.8,25.2){\hyperlink{sec:fm}{\textcolor{white}{\small \textbf{~Section II}~\includegraphics[height=1.5ex]{figs/icons/hyperlink.png}}}}       % Foundation Models
%         \put(19,27){\hyperlink{sec:llm}{\textcolor{white}{\small \textbf{~Section III}~\includegraphics[height=1.5ex]{figs/icons/hyperlink.png}}}}     % LLMs
%         \put(35,26.5){\hyperlink{sec:vlm}{\textcolor{white}{\small \textbf{~Section IV}~\includegraphics[height=1.5ex]{figs/icons/hyperlink.png}}}}     % VLMs
%         \put(52,24.8){\hyperlink{sec:mllm}{\textcolor{white}{\small \textbf{~Section V}~\includegraphics[height=1.5ex]{figs/icons/hyperlink.png}}}}    % MLLMs
%         \put(68.5,28.2){\hyperlink{sec:dm}{\textcolor{white}{\small \textbf{~Section VI}~\includegraphics[height=1.5ex]{figs/icons/hyperlink.png}}}}      % Diffusion
%         \put(82.9,28){\hyperlink{sec:wm}{\textcolor{white}{\small \textbf{~Section VII}~\includegraphics[height=1.5ex]{figs/icons/hyperlink.png}}}}      % World Models
%         \put(0.7,1.1){\hyperlink{sec:dataset}{\textcolor{white}{\small \textbf{~Section VIII}~\includegraphics[height=1.5ex]{figs/icons/hyperlink.png}}}} % Datasets & Simulators
%     \end{overpic}
%     \caption{Overview of \glspl{fm} applied to scenario generation and analysis for autonomous driving, and the corresponding structure of this survey.}
%     \label{fig:FMs}
% \end{figure*}

\subsection{Foundation Models in Autonomous Driving}\label{Sec2B:fm}
Recent studies have explored the integration of \glspl{fm} into \gls{ad} systems, exploiting their adaptability and multimodality across both modular and end-to-end architectures. Comprehensive surveys such as~\cite{huang2023applications, gao2024survey} offer a broad overview of the current landscape, covering \glspl{llm}, \glspl{vlm}, \glspl{mllm}, \glspl{dm}, and \glspl{wm}. 

\textbf{\glspl{llm} in Autonomous Driving:} The survey by Zhu et al.~\cite{zhu2024will} reviews the integration of \glspl{llm} into modular autonomous driving systems, and focuses on perception, decision-making, control, and end-to-end approaches.
Similarly, Wu et al.\cite{wu2025multi} investigate the use of \glspl{llm} for multi-agent perception, decision-making, and simulation. 
% However, both surveys primarily focus on the broader scope of autonomous driving tasks and offer only limited coverage of scenario generation. For instance, Zhu et al.\cite{zhu2024will} briefly mention dataset generation, while Wu et al.\cite{wu2025multi} address scenario generation only in the context of vehicle–assistant interaction.
%
Finally, Li et al.~\cite{li2024large} review the role of \glspl{llm} in enabling human-like reasoning in both modular and end-to-end AD systems and also emphasize training and integration strategies, which is not relevant to our tasks.

\textbf{\glspl{vlm} in Autonomous Driving:} The survey~\cite{zhou2024vision} explores the use of \glspl{vlm} across a range of \gls{ad} tasks, where diffusion and world models are involved in scene understanding, visual reasoning, and dataset generation. 
% While it mentions several works related to scenario generation, the discussion lacks categorization and technical depth, offering no detailed analysis of the datasets, model architectures, or generation techniques involved.

\textbf{\glspl{mllm} in Autonomous Driving:} Cui et al.'s survey~\cite{cui2024survey} focuses on \glspl{mllm} architectures, modality fusion, and their applications across \gls{ad} tasks. 
Fourati et al.'s~\cite{fourati2024xlm}  survey introduces XLMs as a combination of \glspl{llm}, \glspl{vlm}, and \glspl{mllm}, providing a review of their use in \gls{ad} that covers concepts, workflows, and techniques. Finally, Li et al's~\cite{li2025applications} survey explores \gls{llm} and \gls{mllm} applications across different \gls{ad} modules, covering integration and training techniques. %Scenario generation is mentioned briefly under data augmentation; 
% However, in the three surveys \cite{cui2024survey,fourati2024xlm,li2025applications}, scenario generation is mentioned briefly and is not in the main scope.

\textbf{\glspl{dm} and \glspl{wm} in Autonomous Driving:} Guan et al.~\cite{guan2024world} provide an overview of world models in \gls{ad}, focusing on their applications in scenario generation, motion prediction, and control. 
Driving \glspl{wm} are further explored by Tu et al.~\cite{tu2025role}, which categorize them into 2D scene evolution, 3D occupancy prediction, and scene-free paradigms. 
% This work introduces generative tasks, including simulation, anticipatory driving, and 4D pre-training. 
%
% However, both surveys lack a clear classification of practical applications for scenario generation and do not differentiate between \glspl{dm} and \glspl{wm}, nor do they provide an in-depth technical analysis of the underlying methods.

\rev{Regarding the overlap with \textbf{generative AI}, Wang et al.~\cite{wang2025generative} review generative models across the \gls{ad} stack. While broad in scope, their survey adopts a model-centric perspective and lacks a focused discussion on scenario generation.}

\rev{In summary, although the above works cover perception, planning, decision-making, simulation, and testing in \gls{ad}, they do not explicitly address the roles of \glspl{fm} in scenario generation or analysis, as this was not their primary focus. Our review aims to fill this gap.}

\subsection{Scenario Generation in Autonomous Driving}
\begin{figure*}[ht]
    \centering
    \includegraphics[width=1\textwidth]{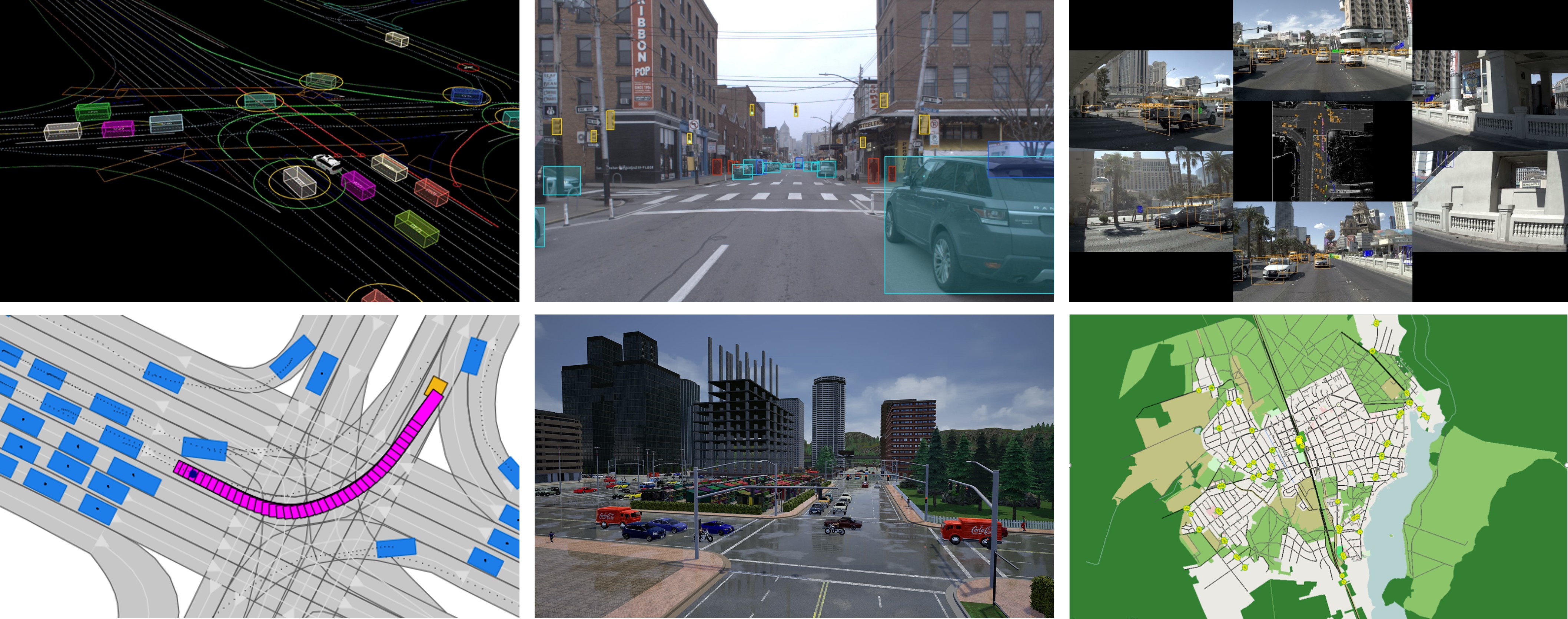} 
    \caption{Examples of driving scenarios in autonomous driving: datasets and simulations used for scenario-based testing. Sensor data such as camera images, videos, and LiDAR point clouds derived from these scenarios can be used to evaluate perception algorithms. Concurrently, simulator-specific scenario formats support rigorous testing of planning and control modules.\\ 
\textbf{Top row (left to right):} Waymo Open motion~\cite{ettinger2021large} dataset; Argoverse2~\cite{wilson2023argoverse} dash camera video; NuPlan~\cite{karnchanachari2024nuplan} multi-camera views with map overlays. \\ 
\textbf{Bottom row (left to right):} CommonRoad~\cite{althoff2017commonroad} motion planning scenario; CARLA~\cite{dosovitskiy2017carla} simulated urban scenario; SUMO~\cite{Lopez2018-sumo} large-scale traffic scenario.}
    \label{fig:driving_scenarios}
\end{figure*}

Scenario formats in \gls{ad} range from annotated sensor data and multi-camera streams to map-based layouts, simulated urban scenes, and traffic-level environments, e.g., OpenScenario\footnote{\url{https://www.asam.net/standards/detail/openscenario/}}. \autoref{fig:driving_scenarios} shows examples of driving scenarios in different formats. In the following, we review the existing surveys on scenario generation with classical and \gls{fm}-based methods.

\textbf{Surveys with Classical Approaches:} Most of the existing reviews deal with classical methods (i.e., not \gls{fm}-based) for scenario generation.
Nalic et al ~\cite{nalic2020scenario} introduce knowledge-driven and data-driven generation approaches, and discuss safety metrics for scenario assessment. They also propose a six-layer model, which captures all essential components of a scenario.
Ding et al.~\cite{ding2023survey} categorize scenario generation methods into data-driven, adversarial, and knowledge-based approaches, providing detailed insights into the mechanisms underlying each. They also highlight the role of deep generative models for synthesizing image- and video-based scenarios with several papers across different models.
In alignment with the ISO/WD PAS 21448 standard, \gls{sotif}, Schutt et al.~\cite{schutt20231001} examine scenario generation across functional, logical, and concrete levels of abstraction. Their review includes machine learning-based generation, optimization-driven scenario exploration, scenario extraction from driving data, and manual scenario design.

\textbf{Survey with FMs:} Huang et al.~\cite{huang2023applications} provide an overview of various types of foundation models and briefly discuss scenario generation. However, their analysis is limited to input modalities and model types, without addressing specific techniques or evaluation strategies.

\textbf{Surveys with VLMs:} Yang et al.~\cite{yang2023llm4drive} examine the use of \glspl{llm} and \glspl{vlm} in tasks such as perception, question answering, and generation. They mention scenario generation using \glspl{vlm}, \glspl{dm}, and \glspl{wm} but provide no clear distinction between these model types. While several evaluation metrics are cited, these are not organized by task or application.
Tian et al.~\cite{tianlarge} present a more structured review of VLMs in autonomous driving across \glspl{llm}, \glspl{vlm}, and \glspl{wm}, focusing particularly on traffic simulation via VLM-guided generation and their integration with diffusion models. However, the survey lacks information about input modalities, scenario generation techniques, and the distinction of model types.

\textbf{Surveys with DMs and WMs:} Fu et al.~\cite{fu2024exploring} review video generation and \glspl{wm}, covering diffusion-based, autoregressive, and reinforcement learning approaches. 
Feng et al.~\cite{feng2025survey} focus on \glspl{wm}, categorizing outputs into images, bird's-eye views, and 3D point clouds, and discuss evaluation metrics such as semantic segmentation and occupancy prediction.
However, neither survey directly connects model outputs to scenario generation tasks. They also fail to distinguish between standalone DMs and WMs, and lack discussion of concrete techniques and evaluation strategies. 

\subsection{Scenario Analysis in Autonomous Driving}
Scenario analysis involves the systematic evaluation of driving scenarios (\autoref{fig:driving_scenarios}). It encompasses tasks such as scenario evaluation, scene understanding, risk assessment, anomaly detection, and accident prediction. Further, it includes identifying safety-critical situations, evaluating system robustness, and supporting informed decision-making in both simulation and real-world environments.

\textbf{Surveys with Classical Approaches:} Riedmaier et al.~\cite{riedmaier2020survey} propose a taxonomy of scenario-based safety assessment methods, including knowledge-based, data-driven, and falsification-based approaches. They emphasize the use of key performance indicators (e.g., time-to-collision, required deceleration) as proxies for accident risk and advocate for the integration of formal methods into safety validation.

Mahmud et al.~\cite{mahmud2017application} review proximal surrogate indicators such as time-to-collision, post-encroachment time, and deceleration rate to avoid a crash. 
They categorize these metrics and identify key research challenges, including metric standardization, real-world validation, and integration into simulation-based scenario analysis frameworks.

\textbf{Survey with FMs:} Huang et al.~\cite{huang2023applications} briefly mention scenario analysis under the term ``perception data annotation'', but do not categorize tasks based on their goals. Additionally, they neither associate datasets with individual studies nor discuss modality transformations; as such, their review does not focus on pre-trained \glspl{fm}.

\textbf{Surveys with VLMs:} Yang et al.~\cite{yang2023llm4drive} address scenario analysis in the context of question answering, focusing mainly on dataset descriptions. Their analysis remains limited, as it lacks discussion of input modalities, methodological approaches, model taxonomy, and evaluation metrics. Similarly, Tian et al.~\cite{tianlarge} consider \gls{vqa} as a form of scenario analysis using \glspl{vlm}, but cover a small number of resources and provide minimal discussion.

\subsection{Critical Summary}  \rev{To the best of our knowledge, existing surveys on \gls{fm}-based scenario generation and/or anaysis in autonomous driving} are limited by the following aspects (summary in Table \ref{tab:survey_comparison_restructured}):

\begin{table*}[htp]
\centering
\caption{Comparison of surveys on \glspl{fm} for scenario generation and analysis in \gls{ad}. Our survey is the first to cover all \gls{fm} types, scenario categories, input modalities, datasets, model types, techniques, and evaluation metrics.}
\label{tab:survey_comparison_restructured}
\begin{threeparttable}
\resizebox{1\textwidth}{!}{%
\begin{tabular}{
 >{\raggedright\arraybackslash}p{2cm}
 >{\centering\arraybackslash}p{0.3cm}
 >{\centering\arraybackslash}p{0.3cm}
 >{\centering\arraybackslash}p{0.6cm}
 >{\centering\arraybackslash}p{0.1cm}
 >{\centering\arraybackslash}p{0.4cm}
 | >{\centering\arraybackslash}p{0.85cm}
 >{\centering\arraybackslash}p{0.85cm}
 >{\centering\arraybackslash}p{0.6cm}
 >{\centering\arraybackslash}p{1.6cm}
 >{\centering\arraybackslash}p{0.5cm}
 >{\centering\arraybackslash}p{0.9cm}
 >{\centering\arraybackslash}p{0.5cm}
 >{\raggedright\arraybackslash}p{0.6cm}
 |>{\centering\arraybackslash}p{0.85cm}
 >{\centering\arraybackslash}p{0.85cm}
 >{\centering\arraybackslash}p{0.6cm}
 >{\centering\arraybackslash}p{0.5cm}
 >{\centering\arraybackslash}p{0.9cm}
 >{\centering\arraybackslash}p{0.5cm}
 >{\raggedright\arraybackslash}p{0.6cm}
}
\toprule
\multirow{2}{*}{\textbf{Survey}} & \multirow{2}{*}{\textbf{LLM}} & \multirow{2}{*}{\textbf{VLM}} & \multirow{2}{*}{\textbf{MLLM}} & \multirow{2}{*}{\textbf{DM}} & \multirow{2}{*}{\textbf{WM}} 
& \multicolumn{8}{c|}{\textbf{Scenario Generation}\tnote{1}} 
& \multicolumn{7}{c}{\textbf{Scenario Analysis}\tnote{2}} \\
\cmidrule(lr){7-14} \cmidrule(lr){15-21}
& & & & & 
& \textbf{\shortstack{Scenario \\ Category}} & 
\textbf{\shortstack{Input \\ Modality}} & 
\textbf{\shortstack{Dataset}} & 
\textbf{\shortstack{Scenario \\Controllability }} & 
\textbf{Model} &
\textbf{Technique} &
\textbf{Metric} & 
\textbf{[n/m]}& 
\textbf{\shortstack{Scenaio \\ Category}} &
\textbf{\shortstack{Input\\ Modality}} &
\textbf{Dataset} &
\textbf{Model} &
\textbf{Technique} &
\textbf{Metric} &
\textbf{[n/m]} \\
\midrule
2023 Huang.~\cite{huang2023applications}   & \checkmark & \checkmark & \checkmark & \checkmark & \checkmark 
&  & \checkmark &  & &\checkmark & & & 13/261 
& & \checkmark &  & \checkmark & & & 5/261 \\
%2023 Cui.~\cite{cui2024survey} &  &  & \checkmark &  &  &  & \checkmark & & & \checkmark& \checkmark & & 1/211 & &\checkmark  &  &\checkmark  & \checkmark& 6/211 \\
%2024 Gao.~\cite{gao2024survey}             & \checkmark & \checkmark & \checkmark & \checkmark & \checkmark &   & \checkmark  & & &\checkmark& & &5/63  & &\checkmark & &\checkmark & & 5/63 \\
%2024 Guan.~\cite{guan2024world}            &  &  &  &  & \checkmark & & & & &\checkmark &\checkmark & & 5/114 & & & & & & \\
%2024 Zhou.~\cite{zhou2024vision}           &  &\checkmark  &\checkmark &  &  &  & \checkmark &  & & & & & 5/182 & &\checkmark & & &  &13/182 \\
2024 Yang.~\cite{yang2023llm4drive}        & \checkmark & \checkmark &  &  &  & &\checkmark & & & &&\checkmark & 11/155 &\checkmark & & &\checkmark & & &13/155 \\
%2024 Zhu.~\cite{zhu2024will}               & \checkmark &  &  &  &  & &\checkmark & & & & & & 6/122 &\checkmark & &\checkmark & &  & 10/122 \\
%2024 Fourati.~\cite{fourati2024xlm}        & \checkmark & \checkmark & \checkmark &  &  & & &\checkmark & &\checkmark & & & 5/133 & & &\checkmark &\checkmark &  & 9/133 \\
2024 Fu.~\cite{fu2024exploring}            &  &  &  &  & \checkmark &  &\checkmark  & & &\checkmark &  & & 11/114 & & & & &  &  \\
2024 Tian.~\cite{tianlarge}        & \checkmark & \checkmark &  &  &\checkmark  &\checkmark & &\checkmark & &\checkmark & & & 15/124 &  & &\checkmark &\checkmark & && 9/124 \\
2025 Feng.~\cite{feng2025survey}           &  &  &  & & \checkmark & & & & &\checkmark &\checkmark & &13/166 & & & & &  &  &\\
%2025 Tu.~\cite{tu2025role}                 &  &  &  & & \checkmark & &\checkmark & &\checkmark & & &  & 15/63 & & & & &  &  \\
%2025 Li.~\cite{li2025applications}                & \checkmark &  &\checkmark &  &  & & & & &\checkmark & & & 6/164 & & \checkmark& &\checkmark &  & 8/164\\
\hline
\rowcolor{gray!15}\textbf{Our Work}        & \checkmark & \checkmark & \checkmark & \checkmark & \checkmark 
& \checkmark & \checkmark & \checkmark & \checkmark & \checkmark & \checkmark  & \checkmark  &  \textbf{93/345}\tnote{*} 
& \checkmark & \checkmark & \checkmark & \checkmark & \checkmark &\checkmark& \textbf{56/345}\tnote{*} \\
\bottomrule
\end{tabular}
}
\vspace{0.5em}
\begin{tablenotes}
\footnotesize
\item[1] \rev{\textbf{Scenario Category}: e.g., safety-critical scenario; 
\textbf{Input Modality}: e.g., text, image, sensor signal; 
\textbf{Dataset}: e.g., nuScenes; \\
\textbf{Scenario Controllability}: full (script), partial (trajectory);
\textbf{Model}: e.g., GPT, latent diffusion; 
\textbf{Technique}: e.g., prompting; 
\textbf{Metric}: e.g., realism.}
\item[2] \rev{\textbf{Scenario Category}: e.g., evaluation, risk assessment; 
\textbf{Input Modality}: e.g., text, image, sensor signal; 
\textbf{Dataset}: e.g., nuScenes; 
\textbf{Model}: e.g., GPT;\\
\textbf{Technique}: e.g., zero-shot,  adapter layer; \textbf{Metric}: e.g., accuracy, language generation quality.
\item \textbf{[n/m]} = number of papers using \glspl{fm} (large pre-trained models) / total papers reviewed in the survey.}
\item[*]\rev{ The 345 papers are categorized as follows: 93 on scenario generation, 56 on scenario analysis, 57 on datasets, 21 on simulators, 25 on benchmark \\ challenges, and 93 on other related topics (e.g., \glspl{fm}' implementation).}
\end{tablenotes}
\end{threeparttable}
\end{table*}

\begin{itemize}

\item \textbf{Lack of focus on scenario generation:} None of the reviewed surveys explicitly focuses on scenario generation using \glspl{fm}. When addressed, scenario generation is either mentioned briefly (e.g.,~\cite{huang2023applications, li2025applications}) or discussed without in-depth analysis of generation techniques, scenario controllability, or evaluation metrics (e.g.,\cite{yang2023llm4drive, fu2024exploring, tianlarge}).

\item \textbf{Incomplete coverage of scenario analysis:} Tasks such as scenario understanding, evaluation, and risk assessment are overlooked. When addressed (e.g.,\cite{yang2023llm4drive} and~\cite{tianlarge}), the analysis is typically reduced to question answering, with little attention paid to task-specific models, methods, or evaluation strategies.

\item \textbf{Limited connections among modalities and tasks:} While several surveys consider the input modalities of \glspl{fm}, they do not establish clear links between these modalities and the techniques, models, and datasets used for scenario generation and analysis. 

\item \textbf{Absence of a structured classification:} No prior work presents a structured classification of \glspl{fm} that spans both scenario generation and scenario analysis, considering pre-trained model types, adaptation methods (e.g., prompting, fine-tuning), input modalities, datasets, and evaluation metrics.
\end{itemize}

\subsection{Contributions}  
To address the limitations of the existing literature reviews, this survey evaluates the landscape of \glspl{fm} in the fields of scenario generation and scenario analysis (Table \ref{tab:survey_comparison_restructured}). In summary, this work provides the following key contributions:
\begin{enumerate}

    \item \rev{We present the first review on the use of \textbf{\glspl{fm} for scenario generation and analysis} in \gls{ad}.}

    \item \rev{\textbf{Structured classification of existing methods}: Our work offers a structured classification covering all \gls{fm} types (i.e., \glspl{llm}, \glspl{vlm}, \glspl{mllm}, \glspl{dm}, \glspl{wm}), scenario categories, input modalities, model types, datasets, techniques, and evaluation metrics.}

    \item \rev{\textbf{Review of datasets, simulation platforms and existing benchmarking competitions}: We review the openly-accessible datasets and simulators used for scenario generation and analysis. Meanwhile, we provide the first review on benchmarking competitions for \glspl{fm} in \gls{ad}. }

    \item \rev{\textbf{Identification of open challenges and future directions}: We identify open research challenges in applying \glspl{fm} to scenario-based testing. By leveraging our analysis, we propose future research directions to improve the adaptability and robustness of \gls{fm}-driven approaches in scenario generation and analysis.}
\end{enumerate}

\hypertarget{sec:llm}{}
\section{Large Language Models (LLMs)}
\label{sec:llm}
This section introduces the foundation and evolution of \glspl{llm},
presents key technological advancements, and reports on common adaptation techniques (e.g., prompt engineering and fine-tuning strategies). We then explore how \glspl{llm} support scenario generation, safety-critical cases, real-world scene synthesis, driving policy evaluation, closed-loop simulation, and \gls{adas} testing. The section concludes with a discussion on scenario analysis, including question answering, scenario understanding, and scenario evaluation.

\subsection{Development of LLMs}
\begin{figure*}[ht]
    \centering
    \includegraphics[width=1\linewidth]{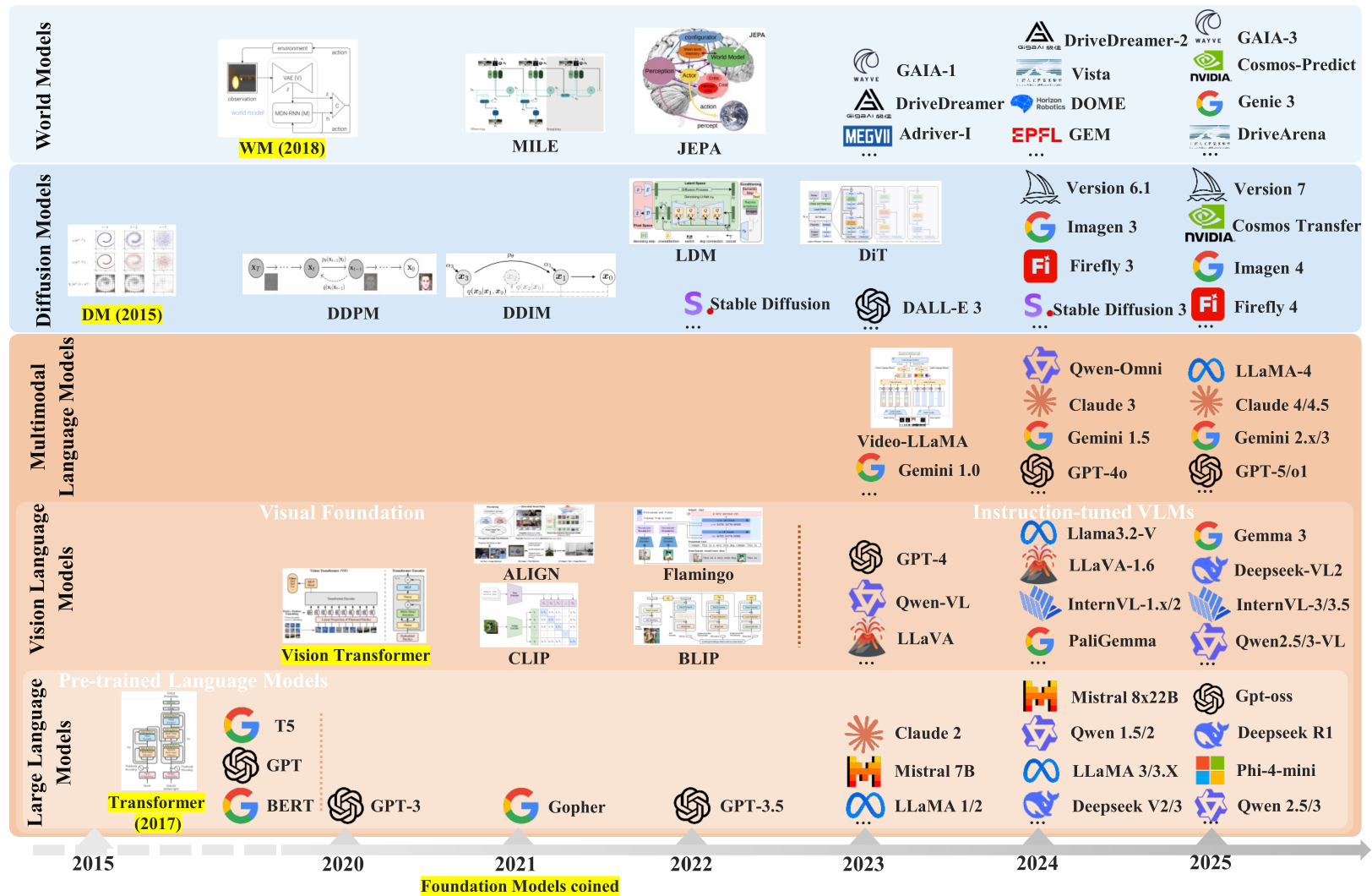}
      \caption{\revB{Timeline of the development of \glspl{fm}. 
    \glspl{llm}, including transformer-based architectures (e.g., BERT), are shown at the bottom.
    \glspl{vlm} built upon visual \glspl{fm} (e.g., \gls{vit} and \gls{clip}) are illustrated in the middle layer, together with instruction-tuned \glspl{vlm} that enable interactive, chat-style vision--language reasoning.
    \glspl{mllm} are shown above.
    In parallel, the evolution of visual \glspl{fm} is highlighted through \glspl{dm} (e.g., the first diffusion model~\cite{firstDM}) and \glspl{wm} (e.g., the world model architecture~\cite{ha2018world}).
    Highlighted entries indicate key conceptual milestones.}}
    \label{fig: LLM_timeline}
\end{figure*}

The most prominent category of \glspl{fm} are \glspl{llm}, which focus on the text modality and are built upon the transformer architecture~\cite{vaswani2017attention}.
\rev{A defining characteristic of these models is their use of self-supervised learning, where they learn language representations by predicting masked or missing parts of text from large-scale unlabeled corpora. This paradigm has enabled models to capture rich contextual and semantic information without the need for manual annotation. The foundation for this approach was laid by static word embeddings~\cite{mikolov2013efficient}, which evolved into pre-trained language models such as GPT~\cite{radford2018improving}, BERT~\cite{devlin2019bert}, and T5~\cite{raffel2020exploring}. These models replaced static embeddings with dynamic, context-aware representations learned directly from text. The release of GPT-3, with 175B parameters~\cite{brown2020language}, marked a major milestone in self-supervised language modeling by demonstrating strong generalization and few-shot learning capabilities, significantly reducing the need for task-specific fine-tuning compared to earlier generations.} \revB{The historical evolution of text-only pre-trained language models and \glspl{llm} is summarized in Figure~\ref{fig: LLM_timeline}.}

% The transformer consists of an encoder-decoder structure featuring Multi-Head Attention and Feedforward Neural Networks (FFN). Multi-head attention enables the model to capture diverse contextual dependencies by attending to different parts of the input simultaneously, while the FFN projects these representations into a shared feature space.

% A key advancement toward self-supervised learning came with the introduction of static word embeddings~\cite{mikolov2013efficient}, laying the groundwork for 
% pre-trained language models. This led to context-aware models like GPT~\cite{radford2018improving}, BERT~\cite{devlin2019bert},  and T5~\cite{raffel2020exploring}, which replaced static embeddings with dynamic representations and learned directly from unlabeled text.
% The release of GPT-3 marked a significant turning point. With 175B parameters and training on a massive corpus~\cite{brown2020language}, it demonstrated generalization and few-shot learning abilities, significantly reducing the need for task-specific fine-tuning compared to earlier models like GPT-1 (117M parameters) and GPT-2 (1.5B parameters).

% A key driver of \glspl{llm} progress has been the rise of large-scale AI accelerators (e.g., NVIDIA A100, Google TPU), enabling efficient training of massive models. 
OpenAI's 2020 scaling laws~\cite{kaplan2020scaling} showed that \glspl{llm} performance improves predictably with increased model size, data, and compute, fueling the trend toward ever-larger foundation models. However, DeepSeek~\cite{bi2024deepseek} challenged this assumption by demonstrating that data quality and alignment matter as much as scale. Through supervised fine-tuning on synthetic expert data and reinforcement learning via \gls{grpo}, they trained smaller models that achieved competitive performance.

% Recently, \glspl{llm} have been integrated throughout the \glspl{ad} pipeline~\cite{wu2025multi, zhu2024will, li2024large}, encompassing both modular and end-to-end architectures. Their reasoning and inference capabilities, along with their adaptability to task-specific objectives through text modality, contribute to enhanced robustness, interpretability, and adaptability.

Since \glspl{llm} are pre-trained on large-scale unlabeled data, various adaptation techniques such as prompt engineering and fine-tuning
have been developed to tailor LLM's behavior to specific tasks.
Sahoo et al.~\cite{sahoo2024systematic} provide an overview of these techniques in their recent survey.
Here, we focus on the adaptation techniques that were applied in the context of driving scenario generation and analysis. 

\textbf{Prompt Engineering:}
This refers to designing and structuring input prompts to guide a pre-trained language model toward producing desired outputs, without modifying its internal parameters. The different techniques are:

\emph{1) \gls{cp}:} Augments prompt with task-specific context or background information, thereby helping the model align more closely with the intended application domain.

\emph{(2) \gls{cot}:} Encourages the model to generate intermediate reasoning steps before producing a final answer. This structured reasoning enhances the logical consistency of the model's reasoning chain, which is particularly beneficial for complex, multi-step tasks.

\emph{(3) \gls{icl}:} Involves task demonstrations (e.g., \textit{one-shot}, or \textit{few-shot}) in the prompt to guide the model towards the correct task behavior.

\emph{(4) \gls{sc}:} A decoding strategy that samples multiple outputs for a given prompt and selects the most frequent or consistent one, improving answer robustness and reliability.

\emph{(5) \gls{rag}:} Enhances the performance on specific tasks by retrieving external knowledge from a database at inference time. A retrieval component identifies the relevant documents to condition the model's response, thereby improving its accuracy.

{\textbf{Fine-Tuning:}} These techniques train the model on datasets to improve its ability for specific tasks. The different fine-tuning techniques are \gls{fft} and \gls{peft}. \gls{fft} updates all model parameters using domain-specific data. While effective, it requires significant computational resources and has limited scalability. \gls{peft} updates only a small portion of the model's parameters, while keeping most of the model frozen. A specific \gls{peft} method is \textit{\gls{lora} }, which injects trainable low-rank matrices into the attention modules of the model, enabling adaptation with minimal parameter updates and reducing computational cost. For instance, full fine-tuning of GPT-3 requires updating all 175B parameters, whereas \gls{lora} can achieve comparable performance by training only around 37.7M parameters~\cite{hu2022lora}.

Additionally, more advanced techniques to adapt \glspl{llm} exist in the reviewed papers.
These include \emph{multi-stage prompting}~\cite{deng2023target, tang2024legend} and \emph{\glspl{mla}}~\cite{aasi2024generating}, which coordinate multiple interacting \gls{llm}s to solve complex tasks collaboratively. Tooling frameworks such as LangChain\footnote{\url{https://github.com/langchain-ai/langchain}} facilitate the construction of modular, agent-based architectures that extend beyond traditional single-prompt interactions. 

\subsection{LLM-Based Scenario Generation}
\label{subsec:scenario_generation}

Advances in \glspl{llm} have triggered the development of \gls{llm}-driven scenario generation to test intelligent vehicle systems.
Based on their individual objectives, we classify the existing works into six categories and list representative works within each category in \autoref{tab:llm_sum}:

\textbf{Safety-Critical Scenario Generation:} 
% \label{sub3:Safety} - label no longer usable
A key application of \gls{llm}-based scenario generation lies in the creation of safety-critical scenarios.
Often termed ``corner cases'', ``long-tail'', or ``\gls{ood}'' \revB{situations, these scenarios involve high collision risk, abnormal agents' behavior, or reduced safety margins~\cite{gambi2022generating}. Recent \gls{llm}-based approaches can synthesize rare trajectories and scene configurations beyond nominal driving conditions, to stress-test the robustness of \gls{ad} systems and uncover residual safety risks. Unlike \gls{adas} test scenarios, this category neither targets specific assistance functions nor follows predefined regulatory test protocols.}

The work LLMScenario~\cite{chang2024llmscenario} focuses on safety-critical scenario generation based on the HighD dataset~\cite{Krajewski2018highd} with GPT-4. They use \gls{icl}, incorporating critical examples, which are evaluated based on reality and rarity, to guide the generation. Using \gls{cot} and \gls{sc} prompting, the framework generates safety-critical trajectories step-by-step in MetaScenario~\cite{chang2022metascenario}.
\rev{ChatScene~\cite{zhang2024chatscene} uses GPT-4 with \gls{rag} to translate textual safety-critical descriptions into \gls{dsl} scripts such as Scenic~\cite{fremont2019scenic} for CARLA~\cite{dosovitskiy2017carla}.
Its retrieval database is built using Sentence-T5 embeddings that map behaviors and geometric patterns to code snippets. These snippets are then retrieved through \gls{rag} and assembled into complete Scenic scripts.}
Building on structured generation, Aasi et al.~\cite{aasi2024generating} propose a multi-agent pipeline that constructs a branching tree of \gls{ood} scenarios using \gls{cot} prompting. Their Augmenter-LLM, based on GPT-4o, translates descriptions into CARLA scene configurations, which contain maps, weather, objects, and behaviors via API calls. 
A \gls{vlm} then classifies the simulated scenes by \gls{ood} type to identify the safety-critical scenarios.

The methods discussed above operate open loop. In contrast, Mei et al.~\cite{mei2025seeking} focus on online interactive scenario generation using  Waymo Open Motion Dataset~\cite{ettinger2021large}. Their retrieval-augmented framework uses DeepSeek-V3 and DeepSeek-R1 to infer risky behaviors of a vehicle in real time and synthesize adversarial trajectories for it to collide with the ego vehicle. A memory module stores and retrieves intent–planner pairs, allowing continuous refinement and adaptation of the generated scenarios. 

%\vspace{-2mm}
Despite promising advances, current works often operate offline or focus on limited risk types, limiting their generalizability to complex, multi-agent contexts. Future work could integrate interactive generation, enhance safety verification in simulation, and develop evaluation pipelines by leveraging \glspl{vlm} to assess the plausibility and criticality of the generated scenarios.

\textbf{Real-World Scenario Replication:}
% \label{sub3:Real} - label no longer usable
\rev{Creating realistic driving scenarios is challenging due to the difficulty of accurately reproducing real-world conditions. A common strategy involves replaying recorded driving data in simulation environments or leveraging real crash reports to reconstruct the corresponding events. Realistic traffic scenes can also be replicated by grounding them on real-world maps, thereby preserving authentic infrastructure, road layouts, and environmental features.}

LCTGen~\cite{tan2023language} leverages GPT-4 with \gls{icl} and \gls{cot} prompting to convert crash report into structured YAML-like descriptions. Then a retriever module matches these structured descriptions with relevant maps from the Waymo Open Dataset~\cite{sun2020waymoopen}.
These ``map-grounded'' inputs are then processed by a generative model using multi-layer perceptrons and learned masks to produce realistic driving scenarios. 
In  Chat2Scenario~\cite{zhao2024chat2scenario}, recorded datasets from HighD with user-defined criticality and textual descriptions are used as input. They use a templated contextual prompting scheme with GPT-4 and retrieve relevant scenarios that match the user's input with ASAM OpenScenario~\cite{asam2024openscenario} format. %Then visualize them using Esmini\footnote{\url{https://esmini.github.io/}} and CarMaker\footnote{\url{https://www.ipg-automotive.com/}}.
For microscopic simulation, ChatSUMO~\cite{li2024chatsumo} utilizes Llama~3.1~\cite{LLaMA2023} with template-based prompts to extract user requirements for traffic volume, city, and network type. Then, ChatSUMO translates these parameters into SUMO~\cite{Lopez2018-sumo} configurations, with osm~\cite{OpenStreetMap} maps retrieved through \gls{rag}. Simulation outputs, including traffic density, travel time, and emissions, are visualized and summarized via a Streamlit\footnote{\url{https://streamlit.io/}}.
SeGPT~\cite{li2024chatgpt} synthesizes diverse and challenging test data from recorded trajectories. Their framework supports large-scale scenario synthesis and compares zero-shot prompting with \gls{cot} to evaluate LLM-guided generation performance on the dataset from~\cite{zhan2019interaction}.

%\vspace{-2mm}
The reviewed papers generate scenarios from recorded data and crash reports. One of the future directions is to first generate scenarios from recorded datasets, and then incorporate natural language descriptions as feedback. This hybrid approach could significantly enhance both the realism and diversity of the resulting scenarios by aligning data-driven generation with human-intuitive requirements.

% \textbf{Metrics:} Evaluation of real-world scenarios considers the following:
% (I) \textit{Trajectory accuracy}, which can be quantified with the mean Average Displacement Error (mADE), mean Final Displacement Error (mFDE), and Maximum Mean Discrepancy (MMD), to measure how closely simulated behaviors match the ground truth~\cite{tan2023language,li2024chatgpt};
% (II) \textit{Semantic correctness}, which uses classification metrics like accuracy, F1 score to assess alignment with crash report descriptions~\cite{zhao2024chat2scenario};
% and, (III) \textit{Traffic realism}, which analyzes the traffic density, travel time, fuel consumption, and emissions to evaluate map-based simulations~\cite{li2024chatsumo}.

\textbf{Driving Policy Test Scenario Generation:}
%\label{sub3:Policy} - label no longer usable
\revB{Driving policy test scenario generation focuses on evaluating automated driving policies, such as motion planners or controllers, under systematically constructed traffic situations. Recent \gls{llm}-based methods generate executable scenarios that are deployed in simulations to assess planning behavior, policy robustness during algorithm development.}

In LCTGen~\cite{tan2023language}, generated real-world crash scenarios are used to assess the performance of a motion planner within the MetaDrive~\cite{li2022metadrive} simulator.
In TTSG~\cite{ruan2024traffic}, GPT-4o-generated scenarios are used for multi-agent planning validation in critical scenarios. 
Specifically, they constructed a road and agent database using \gls{rag} with \glspl{llm} and proposed ranking strategies to select the best-fitting road based on the agent's behavior. 
In contrast, AutoSceneGen~\cite{aiersilan2024generating} uses a code-designed filter to select the valuable parts of the scenario description based on simulation documents and \gls{icl}, and adds scenario examples to the prompt. A code-based validator then transfers and verifies the GPT-4 output, which directly generates \gls{dsl}-style configuration code compatible with CARLA. \rev{The resulting scenarios are subsequently used to evaluate the performance of a motion planner.}

%  \textbf{Metrics:} Driving policy performance is commonly evaluated through:  (I) \textit{Trajectory accuracy}, quantified with ADE and FDE~\cite{aiersilan2024generating} capturing deviations between predicted and ground-truth trajectories; % (2)
% (II) \textit{Policy robustness}, using collision rate and overall scores, for instance, using the CARLA SafeBench to assess safety and resilience of policies under diverse conditions~\cite{ruan2024traffic}.
%\vspace{-2mm}

\textbf{Closed-loop Scenario Generation:}
% \label{sub3:Close-loop} - label no longer usable
Recent works address the limitations of static datasets by introducing closed-loop scenario generation with \glspl{llm}.
Closed-loop scenarios enable the validation of multi-agent interactions and ego-reactive behaviors.

ProSim~\cite{tan2024promptable} presents a promptable closed-loop simulation framework, where prompts such as goal points, route sketches, action tags, and natural language instructions are used to guide an agent's behavior.
Llama3.1-8B is fine-tuned with \gls{lora} to generate policy tokens, and a lightweight policy module rolls out the agent's trajectories in a closed loop within the Waymo simulator.
In \gls{llm}-attacker~\cite{mei2025llm}, an adversarial scenario generation is proposed. It employs three coordinated modules based on Llama3.1-8B, for initialization, reflection, and modification, to identify and refine adversarial vehicle behaviors using \gls{cot}.
These modules iteratively generate and adjust the attacker's trajectories to collide with the ego vehicle. Their framework is trained with reinforcement learning in a closed-loop setting using the Waymo Open Dataset.
In contrast, CRITICAL~\cite{tian2024enhancing} focuses on ego-agent policy learning without adversarial agents. It integrates Mistral-7B via LangChain into a standard reinforcement learning loop in the HighwayEnv environments~\cite{highway-env}.
Their LLM is used to generate diverse scenario configurations, e.g., vehicle density, number of cars, and to shape safety-related rewards, enabling robust policy learning under different conditions.

Together, these works demonstrate complementary strategies: ProSim enables fine-grained control and interactivity, LLM-Attacker focuses on adversarial testing, and CRITICAL supports LLM-guided training environments. Future research could benefit from unifying these paradigms into a single framework that supports diverse behavior modeling, adversarial robustness, and controllable training environments.

% \textbf{Metrics:} Evaluation in closed-loop simulation is typically done with (I) \textit{trajectory accuracy}, assessed via displacement-based metrics such as ADE and FDE\cite{tan2024promptable,mei2025llm}, and (II) \textit{controllability}, defined as the relative improvement in ADE when prompts are applied, thus measuring how well an agent's behavior follows user-specified intentions\cite{tan2024promptable}.
%\vspace{-2mm}

\textbf{Image Datasets Generation:}
% \label{sub3:Dataset} - label no longer usable
Real-world camera datasets are widely used in autonomous driving research, but often lack the diversity and editability required for generating specific test cases. To address this, recent work explores language-guided editing of recorded images.

ChatSim~\cite{wei2024editable} introduces a collaborative multi-agent framework with GPT-4, where each \gls{llm} agent handles a specialized scene editing task, such as viewpoint changes, vehicle manipulation, asset insertion, and motion planning, based on natural language instructions. ChatSim leverages neural rendering and lighting estimation to achieve photorealistic, multi-camera scene synthesis with external digital assets.

% \textbf{Metrics:} The evaluation of generated datasets focuses on (I) \textit{semantic alignment}, which assesses how closely the generated content matches the input command~\cite{wei2024editable}, and (II) \textit{visual quality}, measured by pixel-level fidelity, structural similarity, and perceptual similarity~\cite{wei2024editable}. Most metrics focus on (III) \textit{human-perceived realism}, such as pixel-level fidelity and structural similarity, but do not necessarily reflect how well scenes support perception tasks in autonomous systems. %Future work could adopt task-aware evaluation to assess how these scenes impact downstream tasks like detection or planning.

\textbf{\gls{adas} Test Scenario Generation:}
\revB{\gls{adas} test scenario generation focuses on translating functional descriptions derived from regulations, standards, and test protocols into executable and reproducible test scenarios for function-level evaluation. Recent \gls{llm}-based approaches parse regulatory text, specifications, or reports into structured representations, and generate logical or concrete scenarios in domain-specific languages for simulation-based testing of \gls{adas} stacks, such as Apollo~\cite{ap}. This category emphasizes standardization, reproducibility, and coverage, and is closely aligned with regulatory and assessment frameworks such as OpenXOntology and UN~R157.}
\begin{table*}[htp]
\centering
\caption{Summary of Scenario Generation Studies Using Large Language Models.}
\label{tab:llm_sum}
\begin{threeparttable}
\resizebox{1\textwidth}{!}{%
\begin{tabular}{>{\raggedright\arraybackslash}p{2.4cm} >{\centering\arraybackslash}p{1cm}
 >{\centering\arraybackslash}p{0.6cm}
 >{\raggedright\arraybackslash}p{2.1cm}  >{\raggedright\arraybackslash}p{2cm} >{\raggedright\arraybackslash}p{2.1cm} >{\raggedright\arraybackslash}p{1.9cm} >{\raggedright\arraybackslash}p{2.1cm} c >{\raggedright\arraybackslash}p{2cm}}
\toprule
\multirow{2}{*}{\textbf{Category}} &
\multicolumn{4}{c}{\textbf{Input}} &
\multirow{2}{*}{\textbf{Model}} &
\multirow{2}{*}{\textbf{Technique\textsuperscript{\textbf{1}}}} &
\multirow{2}{*}{\textbf{Simulator}} &
\multirow{2}{*}{\textbf{Output\textsuperscript{\textbf{2}}}} &
\multirow{2}{*}{\textbf{Paper}} \\
\cmidrule(lr){2-5}
& \textbf{Trajectory} & \textbf{DSL}  & \textbf{Dataset} & \textbf{Database} &
& & & & \\
\midrule

\multirow{8}{*}{\textbf{\makecell{Safety-critical\\ Scenario}}}
&\checkmark &   &HighD\cite{Krajewski2018highd} &&  GPT-4 &CoT, ICL, SC &Metascenario\cite{chang2022metascenario} &\trajectory  &LLMScenario~\cite{chang2024llmscenario}  \\
&\cellcolor{LightGray} 
&\cellcolor{LightGray}  \checkmark 
&\cellcolor{LightGray} 
&\cellcolor{LightGray}\makecell{Map\\ Position\\ Behaviors}
&\cellcolor{LightGray}\makecell{GPT-4} 
&\cellcolor{LightGray}CoT, ICL, RAG 
&\cellcolor{LightGray}CARLA\cite{dosovitskiy2017carla}
&\cellcolor{LightGray}\script
&\cellcolor{LightGray}ChatScene~\cite{zhang2024chatscene} \\
& &   &CARLA  & &\makecell[{{l}}]{GPT-4o \\ Claude 3.5 Sonnet \\ Gemini 1.5 Pro} &CoT, MLAs &CARLA &\script &Aasi et al.~\cite{aasi2024generating} \\
&\cellcolor{LightGray}\checkmark  &\cellcolor{LightGray}  &\cellcolor{LightGray}WOMD\cite{ettinger2021large}  &\cellcolor{LightGray}\makecell{Trajectory\\ Behaviors}  &\cellcolor{LightGray}\makecell[{{l}}]{DeepSeek V3 \\ DeepSeek R1} &\cellcolor{LightGray}\makecell{CoT, CP, ICL,\\RAG} &\cellcolor{LightGray} WOMD &\cellcolor{LightGray}\script &\cellcolor{LightGray}Mei et al.~\cite{mei2025seeking} \\
\hline

\multirow{5}{*}{\textbf{\makecell{Real-world\\ Scenario Replication}}}
& &   &Waymo Open\cite{sun2020waymoopen}  &\makecell{Map\\NHTSA\cite{nhtsa2024nmvccs}} &GPT-4 &\makecell{CoT, ICL, RAG} &MetaDrive~\cite{li2022metadrive} &\trajectory  &LCTGen~\cite{tan2023language} \\
&\cellcolor{LightGray}  &\cellcolor{LightGray}    &\cellcolor{LightGray}HighD &\cellcolor{LightGray}  &\cellcolor{LightGray} GPT-4 &\cellcolor{LightGray} CoT &\cellcolor{LightGray}\makecell{Esmini\\CarMaker} &\cellcolor{LightGray}\script &\cellcolor{LightGray}Chat2Scenario~\cite{zhao2024chat2scenario} \\
& &  &SUMO, OSM\cite{OpenStreetMap}  &Map  &Llama 3.1-8b &CoT, RAG &SUMO\cite{Lopez2018-sumo} &\script  &ChatSUMO~\cite{li2024chatsumo} \\
&\cellcolor{LightGray} \checkmark &\cellcolor{LightGray}   &\cellcolor{LightGray}Interaction~\cite{zhan2019interaction}  &\cellcolor{LightGray}   &\cellcolor{LightGray}  GPT-4 &\cellcolor{LightGray}  CoT &\cellcolor{LightGray}  Interaction &\cellcolor{LightGray}\trajectory  &\cellcolor{LightGray}SeGPT~\cite{li2024chatgpt} \\
\hline

\multirow{2}{*}{\textbf{\makecell{Driving policy\\ Testing Scenario}}}
& &   &CARLA  &\makecell{Map\\ Position\\ Behaviors} & GPT-4o & CoT, RAG & CARLA &\script  &TTSG~\cite{ruan2024traffic} \\
&\cellcolor{LightGray} &\cellcolor{LightGray}\checkmark   &\cellcolor{LightGray}CARLA  &\cellcolor{LightGray} &\cellcolor{LightGray}GPT-4 &\cellcolor{LightGray}CoT, CP &\cellcolor{LightGray}CARLA &\cellcolor{LightGray}\script &\cellcolor{LightGray}AutoSceneGe~\cite{aiersilan2024generating}\\
\hline

\multirow{4}{*}{\textbf{\makecell{Closed-loop\\ Scenario}}}
&  &    &Waymo Open  &\makecell{Map\\ States}  &Llama 3.1-8b & LoRA & Waymo Sim &\trajectory &ProSim~\cite{tan2024promptable}\\
&\cellcolor{LightGray}\checkmark &\cellcolor{LightGray}  &\cellcolor{LightGray}\makecell{Waymo Open\\MetaDrive\cite{li2022metadrive}}  &\cellcolor{LightGray} &\cellcolor{LightGray}Llama 3.1-8b &\cellcolor{LightGray}CoT, RL &\cellcolor{LightGray}MetaDrive &\cellcolor{LightGray}\trajectory &\cellcolor{LightGray}LLM-attacker~\cite{mei2025llm} \\
& \checkmark &   &HighD  &  & Mistral-7B & CoT &HighwayEnv\cite{highway-env} &\trajectory  &CRITICAL~\cite{tian2024enhancing} \\
\hline

\multirow{1}{*}{\textbf{\makecell{{Image Dataset}}}}
&\cellcolor{LightGray} &\cellcolor{LightGray}   &\cellcolor{LightGray}Waymo Open  &\cellcolor{LightGray}Images &\cellcolor{LightGray}GPT-4 &\cellcolor{LightGray}MLAs &\cellcolor{LightGray} &\cellcolor{LightGray}\video &\cellcolor{LightGray}Chatsim~\cite{wei2024editable} \\
\hline

\multirow{10}{*}{\textbf{\makecell{ADAS Testing \\ Scenario}}}
& &\checkmark  &  &Regulation &GPT-4 &ICL &SUMO &\script  &Guzay et al.~\cite{guzay2023generative} \\
&\cellcolor{LightGray}  &\cellcolor{LightGray} \checkmark   &\cellcolor{LightGray}  &\cellcolor{LightGray}Traffic rule  &\cellcolor{LightGray}GPT-4 &\cellcolor{LightGray}\makecell{ Multi-stage,\\ ICL, CoT, CP} &\cellcolor{LightGray} CARLA &\cellcolor{LightGray}\script  &\cellcolor{LightGray}TARGET~\cite{deng2023target} \\
&  &\checkmark  &  &\makecell{Standard\\ Regulation\\ Test Specification}  & \makecell[{{l}}]{GPT-4 \\ Llama 3} & \makecell{CP, Multi-stage} & CARLA &\script &Petrovic et al.~\cite{petrovic2024llm} \\
&\cellcolor{LightGray} &\cellcolor{LightGray}  &\cellcolor{LightGray}LGSVL\cite{rong2020lgsvl} &\cellcolor{LightGray}NHTSA  &\cellcolor{LightGray}GPT-4 &\cellcolor{LightGray}CP, ICL &\cellcolor{LightGray}LGSVL &\cellcolor{LightGray}\script  &\cellcolor{LightGray}SoVAR~\cite{guo2024sovar} \\
& &\checkmark  & &NHTSA  &GPT-4 &\makecell{CP, ICL,\\ Multi-stage} &LGSVL  &\script  &LeGEND~\cite{tang2024legend} \\
&\cellcolor{LightGray} &\cellcolor{LightGray}\checkmark &\cellcolor{LightGray}   &\cellcolor{LightGray}\makecell{NHTSA\\ OpenXOntology}  &\cellcolor{LightGray}GPT-4 &\cellcolor{LightGray}\makecell{CoT, ICL, SC,\\ Multi-stage} &\cellcolor{LightGray}CARLA &\cellcolor{LightGray}\script &\cellcolor{LightGray}Text2Scenario~\cite{cai2025text2scenario} \\
& &\checkmark  &OpenDRIVE  &\makecell{UNECE R157\\ OpenX Ontology\\ maps}  &Llama 3.1 &\makecell{CP} &VTD &\script &Zhou et al.~\cite{zhou2024automatic} \\
\bottomrule
\end{tabular}}
\begin{tablenotes}
\scriptsize
\item[1] Techniques: 
CoT = Chain-of-Thought prompting; 
ICL = In-Context Learning; 
SC = Self-Consistency; 
CP = Contextual Prompting;
RAG = Retrieval-Augmented Generation;\\ 
RL = Reinforcement Learning; 
LoRA = Low-Rank Adaptation; 
MLAs = Multi-LLM Agent Systems.
\item[2] Output: \video ~Video, \trajectory ~Trajectory, \script ~Scenario script.
\end{tablenotes}
\end{threeparttable}
\end{table*}

One of the pioneering papers on \gls{adas} test scenario generation using \glspl{llm} is Guzay et al.~\cite{guzay2023generative}, who use GPT-4 with \gls{icl} to convert regulatory descriptions into SUMO-compatible XML files.
Expanding on this, TARGET~\cite{deng2023target} introduces a multi-stage prompting by using GPT-4 that parses traffic rules into a \gls{dsl} of CARLA using \gls{cot} and \gls{icl} to evaluate multiple \gls{adas} software. A rule-to-script generator then produces a scenario script. 
Petrovic et al.\cite{petrovic2024llm} extend this direction by processing \gls{adas} test topologies and standardization documents from UNECE R157
\footnote{\url{https://unece.org/transport/documents/2021/03/standards/un-regulation-no-157-automated-lane-keeping-systems-alks}}. The test topology is converted into a metamodel that includes elements such as the environment, sensor, and actuator configurations. Standardization documents are parsed into Object Constraint Language (OCL) using LLMs. Based on the combined metamodel, OCL constraints, and a specific test description, an LLM (e.g., GPT-4 or Llama~3) is used to generate \gls{dsl} test scenarios, which are then simulated in CARLA. 
In a more data-driven approach, SoVAR~\cite{guo2024sovar} reconstructs crash scenarios from NHTSA~\cite{nhtsa2024nmvccs} reports by extracting structured attributes with GPT-4 and generating trajectories and simulation scripts via constraint solving, producing LGSVL~\cite{rong2020lgsvl}-compatible test scenarios via API calls focused on realism.
In contrast, LeGEND~\cite{tang2024legend} follows a top-down approach: it abstracts reports into functional scenarios, transforms them into logical \gls{dsl} representations via a two-stage GPT-4 pipeline, and applies multi-objective search to generate diverse and critical scenarios to evaluate Apollo \gls{adas} stack.

Text2Scenario~\cite{cai2025text2scenario} introduces a standardized hierarchical scenario repository based on the \glspl{sotif} framework and applies multi-stage prompting (\gls{cot}, \gls{icl}, \gls{sc}) with GPT-4 to generate logical scenarios from free-form descriptions. The resulting logical scenario is then converted into the OpenScenario format through code and simulated in CARLA to evaluate the performance of multiple \gls{adas} stacks.
Finally, Zhou et al.~\cite{zhou2024automatic} focus on lane-keeping systems by using Llama~3.1 and prompt templates to extract scene elements from UNECE R157
~aligned descriptions. These descriptions are structured and converted into OpenScenario DSL files using OpenXOntology\footnote{\url{https://www.asam.net/standards/asam-openxontology/}} and OpenDRIVE\footnote{\url{https://www.asam.net/standards/detail/opendrive/ maps}}, then simulated in the VTD\footnote{\url{https://hexagon.com/de/products/virtual-test-drive}} simulation environment. 

\vspace{0.8mm}
While LLM-based frameworks effectively generate \gls{adas} test scenarios from crash reports and regulations, they often overemphasize rare edge cases~\cite{guo2024sovar,tang2024legend}, neglecting common driving scenarios that are essential for broader testing. Currently, we are missing the incorporation of  routine test cases and utilizing real-world maps from \gls{osm} or SUMO to enhance the scenario diversity and fidelity. %Moreover, assessing what-if scenarios, which means slight variations of base scenarios, can help estimate the risk of scenarios that appear safe at first glance. Sensitivity analysis methods, such as those based on adjoint equations, could be integrated to formally evaluate how minor changes in inputs impact the system's behavior, thereby enriching the safety validation.

% \textbf{Metrics:} Scenario generation for \gls{adas} is commonly evaluated in three ways: (I) \textit{Feasibility} is measured using compile error rate, runtime error rate, and execution rate to assess robustness~\cite{cai2025text2scenario,petrovic2024llm};
% (II) \textit{Accuracy} is based on semantic alignment between generated scenarios and input descriptions, often judged by human-labeled matching~\cite{cai2025text2scenario,guo2024sovar};
% (III) \textit{Efficiency}, instead, is measured by generation time or speed-up over manual scripting~\cite{petrovic2024llm}.
\begin{table*}[htp]
\centering
\caption{Summary of Scenario-Analysis Studies Using Large Language Models.}
\label{tab:llm_Analysis}
\begin{threeparttable}
\resizebox{1\textwidth}{!}{%
\begin{tabular}{
  >{\raggedright\arraybackslash}p{1.9cm} 
  >{\raggedright\arraybackslash}p{2.8cm}  
  >{\raggedright\arraybackslash}p{1.9cm}  
  >{\raggedright\arraybackslash}p{2.3cm}   
  >{\raggedright\arraybackslash}p{1.8cm} 
  p{1.7cm} 
  p{1.2cm} 
  p{1.7cm} 
  l }
\toprule
\multirow{2}{*}{\textbf{Category}} &
\multicolumn{4}{c}{\textbf{Input}} &
\multirow{2}{*}{\textbf{Model}} &
\multirow{2}{*}{\textbf{Technique\textsuperscript{\textbf{1}}}} &
\multirow{2}{*}{\textbf{Focus}} &
\multirow{2}{*}{\textbf{Paper}} \\
\cmidrule(lr){2-5}
& \textbf{\makecell{Scenario Elements}}  
& \textbf{\makecell{Elements \\ Narrator}}  
& \textbf{Dataset} 
& \textbf{Database} 
& & & & \\
\midrule
\multirow{1}{*}{\textbf{\makecell{Question\\ Answering (QA)}}}
&\makecell{Road, Ego, NPC Vehicles,\\  Pedestrians}    
&\makecell{Language\\ Generator}  
&In-house 
& 
&GPT-3.5 
&ICL 
&Driving QA 
&Chen et al.~\cite{chen2024driving} \\
\hline
\multirow{1}{*}{\textbf{\makecell{Scenario\\ Understanding}}}
&\cellcolor{LightGray}\makecell{Road, Weather, Ego,\\ Traffic Light,\\ NPC Vehicles} 
&\cellcolor{LightGray}\makecell{LLaVA,\\Language Parsing} 
&\cellcolor{LightGray}DLR UT\cite{schicktanz2025dlr}  
&\cellcolor{LightGray}\makecell{Traffic Condition\\Structured Data}   
&\cellcolor{LightGray}\makecell{GPT-4} 
&\cellcolor{LightGray}\makecell{CoT\\ RAG} 
&\cellcolor{LightGray}\makecell{Reasoning} 
&\cellcolor{LightGray}SenseRAG~\cite{luo2025senserag} \\
\hline
\multirow{8}{*}{\textbf{\makecell{Scenario\\ Evaluation}}}
&\makecell{Road, Weather, \\ Traffic Sign,\\ NPC Vehicles} 
&\makecell{OWL-ViT,\\Language Parsing} 
&CARLA 
&\makecell{}   
&\makecell{text-davinci-003} 
&\makecell{CoT\\ ICL} 
&Anomaly Detection
&Elhafsi et al.~\cite{elhafsi2023semantic} \\
&\cellcolor{LightGray}\makecell{Road, Weather, Ego,\\ NPC Vehicles}      
&\cellcolor{LightGray}Vector Parse  
&\cellcolor{LightGray}DeepScenario\cite{lu2023deepscenario} 
&\cellcolor{LightGray} 
&\cellcolor{LightGray}\makecell{GPT-3.5 \\ LLaMA2-13B\\ Mistral-7B} 
&\cellcolor{LightGray}\makecell{CP \\ ICL} 
&\cellcolor{LightGray}Realism  
&\cellcolor{LightGray}Reality Bites~\cite{wu2024reality}  \\

&\makecell{Road, Ego,\\ NPC Vehicles}
&\makecell{Cartesian Parser,\\ Ego Parser}
&CommonRoad\cite{althoff2017commonroad}
& 
&\makecell{GPT-4o \\ Gemini-1.5Pro \\ Deepseek-V3} 
&\makecell{CP\\CoT\\ICL}
&Safety-Criticality
&Gao et al.~\cite{gao2025risk} \\

&\cellcolor{LightGray}\makecell{Road, Ego, Traffic Light, \\ NPC Vehicles} 
&\cellcolor{LightGray}Not Specified 
&\cellcolor{LightGray}CARLA  
&\cellcolor{LightGray}Interview Data  
&\cellcolor{LightGray}GPT-4o 
&\cellcolor{LightGray}\makecell{CoT\\ RAG}  
&\cellcolor{LightGray}Driving Styles 
&\cellcolor{LightGray}You et al.~\cite{you2025comprehensive}  \\
\bottomrule
\end{tabular}%
}
\begin{tablenotes}
\scriptsize
\item[1] Techniques: 
CoT = Chain-of-Thought prompting; 
ICL = In-Context Learning; 
CP = Contextual Prompting; 
RAG = Retrieval-Augmented Generation. 
\end{tablenotes}
\end{threeparttable}
\end{table*}

\subsection{LLM-based Scenario Analysis}

Recent research has explored the use of \glspl{llm} as a scenario analysis tool and method. % for autonomous driving. 
A key challenge is that \glspl{llm} are primarily designed to process natural language input, whereas driving scenarios are typically defined using structured data formats, such as scripts in \gls{dsl} or sensor outputs with predefined syntax.
This creates a mismatch between how the scenario's information is represented and how \glspl{llm} operate. Bridging this gap is critical to enable effective interpretation of driving scenarios using language models. We classify the existing works into three key areas and list representative works in \autoref{tab:llm_Analysis}.

\textbf{Question Answering (QA):}
%\label{Sec3D:Dataset} - label no longer usable
Applying \glspl{llm} to scenario analysis for \gls{ad} requires domain-specific knowledge, which general-purpose pre-trained models may lack. To bridge this gap, fine-tuning with tailored datasets is essential.
\gls{qa} datasets describing driving scenarios help \glspl{llm} interpret structured driving contexts, and support downstream tasks like trajectory planning and decision-making.

A notable example is~\cite{chen2024driving}, where the authors automate the generation of \gls{qa} datasets with driving scenarios using GPT-3.5. With a structured language generator, they convert vectorized scenario data from their in-house dataset, including agents' positions, speed, and distance, into natural language. With \gls{icl} and pre-defined driving rules, their model generates diverse, context-aware \gls{qa} pairs to reflect realistic driving situations. The \gls{qa} dataset of~\cite{chen2024driving} focuses primarily on perception and prediction. %Extending it to cover other downstream tasks, such as planning or risk assessment, remains an open direction for future work.

% \textbf{Metrics:} In~\cite{chen2024driving}, the evaluation of \gls{qa} datasets is performed with
% (I) \textit{LLM-based \gls{qa} grading}, relying on GPT-3.5 to assign a score based on the answer's quality, and with
% (II) \textit{human grading}, which relies on human annotations to assess a subset of \gls{qa} samples and evaluate the LLMs' results.
%\vspace{-2mm}

\textbf{Scenario Understanding}
%\label{Sec3D:Situation} - label no longer usable
Here, the LLM processes structured sensor or simulator data, such as agent states, road layouts, and traffic signals, to support tasks like scenario captioning (concise descriptions) and reasoning (coherent narratives capturing intent and context).
% To support this, a bridging function typically converts structured inputs into natural language descriptions, enabling LLMs to perform tasks such as perception, prediction, and contextual reasoning.

The SenseRAG~\cite{luo2025senserag} introduces a \gls{rag}-based framework from the DLR urban traffic dataset~\cite{schicktanz2025dlr} for scenario understanding. They use a  \gls{vlm} to generate traffic condition descriptions into textual descriptions, which are then mapped to a structured database, including additional structured information with weather, city, and traffic participants. Using \gls{cot} prompting and \gls{sql} query generation, GPT-4 retrieves and reasons over the data to refine perception and enhance trajectory prediction.

%Future research in \gls{llm}-based scenario understanding should explore broader downstream tasks beyond normal reasoning, such as recognizing relationships and
%interactions among vehicles, and instead tackle broader downstream tasks like behavior prediction, risk, and intent inference. This will leverage the \glspl{llm}' structured reasoning capabilities on diverse scenario representations.

% \textbf{Metrics:} The evaluation of scenario understanding approaches is primarily based on (I) \textit{trajectory prediction metrics} (e.g., ADE and FDE), which assess the accuracy of the future trajectory predictions made by the \glspl{llm}~\cite{luo2025senserag}.

\textbf{Scenario Evaluation}
% \label{Sec3D:Evaluation} - label no longer usable
Recent work demonstrates how \glspl{llm} can support the evaluation of driving scenarios, by reasoning over structured simulation data or scenario images converted into natural language. This includes the evaluation of anomaly detection, scenario realism, safety-criticality, and driving behavior.
Elhafsi et al.~\cite{elhafsi2023semantic} detect semantic scenario anomalies using \glspl{llm}. %A \glspl{vlm} is applied for object detection and generation of scene descriptions via language parsing. 
Their scenarios are evaluated using OpenAI's text-davinci-003, which is prompted with \gls{cot} and \gls{icl}. Reality Bites~\cite{wu2024reality} is one of the first works to evaluate the reasoning ability of \glspl{llm} in assessing scenario realism. It transforms XML-formatted DeepScenario~\cite{lu2023deepscenario} data into natural language and uses \gls{icl} prompting with models like GPT-3.5, Llama2-13B, and Mistral-7B to judge the alignment with realistic driving conditions. Gao et al.~\cite{gao2025risk} propose a framework to analyze safety-criticality in driving scenarios simulated with a motion planner~\cite{trauth2024frenetix} in the CommonRoad~\cite{althoff2017commonroad} environment. 
They convert structured scenario data into natural language and prompt \glspl{llm} via \gls{cp}, \gls{cot}, and \gls{icl} to evaluate the safety-criticality of the scenario and infer the risk level of the agent.
Also, they generate safety-critical scenarios by modifying the trajectories of identified adversarial vehicles. Meanwhile, You et al.~\cite{you2025comprehensive} focus on holistic driving assessment, converting interview and simulation data into a structured knowledge database for \gls{rag}. 
In their framework, GPT-4o classifies driving styles (cautious, aggressive) and performance levels based on aggregated context, including scenario-level information like weather, ego vehicle data, and surrounding traffic participants. 

Overall, LLM-based scenario evaluation still depends on token-heavy prompting and handcrafted prompts. Emerging reasoning models, such as \rev{OpenAI GPT-5 and Gemini 2.5 Pro,} may enable more efficient, zero-shot approaches. %Key future directions include reducing prompt overhead, improving multimodal grounding, and enabling real-time, scalable evaluation.

\subsection{Limitations and Future Directions}
Our review of \gls{llm}-based scenario generation \rev{and analysis} reveals that many existing approaches rely heavily on prompting strategies, as summarized in \autoref{tab:llm_sum} and  \rev{ \autoref{tab:llm_Analysis}}. The effectiveness of \rev{the corresponding frameworks} often depends on manually crafted prompts. To mitigate this dependency, recent tools such as DSP\footnote{\url{https://dspy.ai/}} provide AI-driven prompt optimization frameworks that automatically generate task-aligned prompts based on user-defined evaluation metrics. Another promising direction involves leveraging advanced reasoning models, such as OpenAI’s GPT-o1 and DeepSeek-R1, which offer stronger zero-shot reasoning capabilities and may reduce the reliance on handcrafted prompts.

Furthermore, future research should explore moving beyond single-turn prompting by adopting interactive, dialogue-based generation. Structuring \glspl{llm} as chatbot-style agents would allow users to iteratively define scenario requirements, enabling the synthesis of customized, constraint-compliant scenarios rather than relying on static outputs.

\noindent \rev{\textbf{\gls{llm}-based Scenario Generation:}}
A significant gap persists between simulation-based scenario generation and real-world validation. Bridging this gap requires the development of \gls{adas} test scenarios aligned with practical safety standards such as SOTIF. By leveraging the reasoning capabilities of \glspl{llm}, future systems could generate functional and logical scenarios directly from textual descriptions and test specifications. This would facilitate the creation of challenging, safety-critical corner cases and enhance the applicability of generated scenarios to real-world testing and system validation.

\noindent \rev{\textbf{\gls{llm}-based Scenario Analysis:}}
\glspl{llm} are also increasingly used to understand and analyze driving scenarios. While many innovative frameworks have emerged, a major limitation lies in computational inefficiency. Since most \glspl{llm} operate on textual inputs, sensor data from modalities such as LiDAR, images, and radar must first be converted into natural language descriptions using narrators or intermediate modules, as shown in \autoref{tab:llm_Analysis}. This pre-processing step adds latency and increases the input complexity. Moreover, improving the quality of the analysis often requires complex prompting strategies such as chain-of-thought reasoning, further complicating real-time deployment.
To address these challenges, one promising approach is to fine-tune \glspl{llm} for scenario understanding tasks, avoiding reliance on elaborate prompting. However, this direction is currently hindered by the lack of large-scale, high-quality scenario question-answering datasets and evaluation benchmarks: most works focus on framework validation rather than dataset creation. 
% Future research should leverage existing frameworks to generate labeled data that can support supervised training and more reliable evaluation. An alternative direction involves designing lightweight or fast-inference \glspl{llm} suitable for onboard deployment. These models must strike a balance between reasoning capability and computational efficiency to enable practical use in real-time autonomous driving systems.

\hypertarget{sec:vlm}{}
\section{Vision Language Models (VLMs)}
\label{sec:vlm}
This section introduces \glspl{vlm}, summarizes their key adaptation techniques, and reviews \gls{vlm}-based scenario generation for safety-critical, real-world, and \gls{adas} testing applications, and \rev{image} datasets generation. 
Additionally, it explores how \glspl{vlm} support scenario analysis tasks such as \glspl{vqa}, scene understanding, benchmarking, and risk assessment.

\subsection{Development of VLMs}
\label{subsec:vlm-development}
In 2020, the \gls{vit}~\cite{dosovitskiy2020image} extended the transformer architecture from \gls{nlp} to computer vision by splitting an image into fixed-size patches. This enabled embedding an image as a sequence of tokens and processing the sequence of tokens with a standard transformer encoder. This success inspired researchers to combine visual and textual modalities, leading to the development of \glspl{vlm}, which now can jointly process images and text at the same time. A milestone was the development of \gls{clip}~\cite{radford2021clip}, which was trained on hundreds of millions of image–text pairs using a contrastive learning objective,  enabling effective zero-shot performance without task-specific supervision. 
ALIGN~\cite{jia2021scaling} scaled this approach to billions of noisy web-crawled pairs.  BLIP~\cite{li2022blip} unified multiple tasks with captioning and retrieval into a single training framework. Flamingo~\cite{alayrac2022flamingo} introduced few-shot vision-language prompting with frozen backbones and cross-attention layers for rapid adaptation. 

\revB{Building upon these visual foundations, a significant shift occurred with the introduction of visual instruction tuning, which aims to align vision–language inputs with human intent through instruction-following behavior. Representative models such as MiniGPT-4~\cite{zhu2023minigpt} and LLaVA~\cite{liu2023visual} align pretrained vision encoders (e.g., \gls{clip}) with large language models such as LLaMA~\cite{LLaMA2023} via lightweight projection modules and apply instruction tuning, enabling chat-style vision–language reasoning. An overview of the evolution from visual foundation models to instruction-tuned \glspl{vlm} is illustrated in Figure~\ref{fig: LLM_timeline}.}

By leveraging the \glspl{vlm}'s ability to jointly reason over images and text, researchers have explored new concepts for autonomous driving taks. As summarized in recent surveys~\cite{yang2023llm4drive,tianlarge}, \glspl{vlm} enable interpretable and adaptable systems that support open-ended interaction, improve generalization to unseen scenarios, and facilitate multimodal reasoning. These advancements mark a shift toward more intelligent and explainable autonomous vehicles, laying the groundwork for safer and more human-aligned driving agents.

\begin{figure}[h]
    \centering
    \includegraphics[width=1\linewidth]{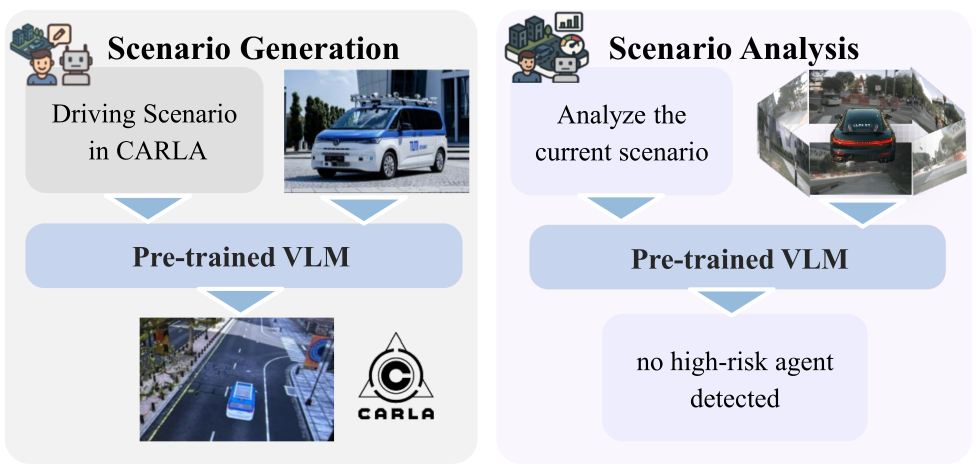}
    \caption{\rev{Pre-trained VLMs use both text descriptions and visual inputs for two tasks: (1) scenario generation using text prompts and scene images, and (2) scenario analysis using image understanding and textual reasoning for risk assessment.}}
    \label{fig: VLM_tasks}
\end{figure}

\rev{\glspl{vlm} provide three core capabilities:
Multimodal understanding jointly processes images and text, such as image captioning and \gls{vqa} (e.g., Flamingo, BLIP);
Image–text matching involves assessing semantic alignment between an image and a caption (e.g., ALIGN, \gls{clip});
Text-to-image generation involves synthesizing novel visuals from natural language prompts, pioneered by DALL-E~\cite{ramesh2021zero}. Building on these foundations, VLMs can be adapted to support individual \gls{ad} modules (perception, prediction, planning) and even end-to-end vision–language–action (VLA) frameworks that directly map visual and linguistic inputs to driving behaviors. In this survey, we focus specifically on VLM-based driving scenario generation and scenario analysis, as illustrated conceptually in Figure~\ref{fig: VLM_tasks}.}

\textbf{Adaptation Techniques for VLMs:}
%\label{sec:vlm-adapt} - label no longer usable
\rev{Current compact \glspl{vlm} use \glspl{llm} as backbones, by adding text tokenizers and vision encoders. Like \glspl{llm}, \glspl{vlm} are pre-trained and then adapted for downstream tasks.} Beyond the standard prompt engineering \rev{techniques of \glspl{llm}}, the following adaptation strategies are commonly employed in the context of scenario generation and analysis in \gls{ad}.

Modality alignment modules are additional trainable modules that transform visual inputs into formats compatible with language models. Common approaches include:

\noindent(I) \emph{\gls{qformer}:} A transformer with learnable queries that aligns image features with the language model input space via cross-attention (e.g., BLIP-2~\cite{li2023blip}).

\noindent(II) \emph{Cross-attention:} Used to resample variable-length image or video tokens into a fixed-size latent representation, enabling consistent language interaction (e.g., Flamingo~\cite{alayrac2022flamingo}).

\noindent(III) \emph{\gls{mlp} mapping:} A linear or \glspl{mlp} projects vision encoder outputs to match the dimensionality required by the language model~\cite{inoue2024nuscenes, wang2024omnidrive}.

\noindent(IV) \emph{Structure-aware encoder (Prior tokenizer):} A perception-aware module that encodes structured detection outputs, such as semantic attributes, into token embeddings for downstream reasoning. For example, Reason2Drive~\cite{nie2024reason2drive} introduces a module called Prior Tokenizer to fuse region features with object-level semantics.

%%%%%%%%%%%%%%%
Fine-tuning techniques train \glspl{vlm} on datasets of instruction–response pairs involving both visual and textual inputs to improve their ability to follow multimodal instructions. Two main strategies are used:

\noindent(I) \textit{\gls{fft}:} All model weights are updated on the target dataset. It typically yields the highest task performance but incurs high computational costs and risks of overfitting. Several reviewed works adopt full fine-tuning for smaller \glspl{vlm}, striking a practical balance between effectiveness and efficiency~\cite{inoue2024nuscenes, gopalkrishnan2024multi}.

\noindent(II) \textit{\gls{peft}:} These methods enable adaptation by updating only a small number of additional parameters.
% \emph{Adapter Layers:} Modules such as CLIP-Adapter~\cite{gao2024clip}, CoOp's learnable prompt vectors~\cite{zhou2022learning}, and Tip-Adapter's cached feature controller~\cite{zhang2021tip} that insert lightweight components into the model while keeping the backbone frozen.
One of the most common methods are 
\gls{lora} matrices, which are injected into attention or feed-forward layers, to enable efficient adaptation with minimal parameter overhead~\cite{hu2022lora}. \rev{An extension of this idea is QLoRA, which further reduces memory usage by applying quantization to the base model during adapter training.}

%Recent \glspl{vlm} increasingly incorporate a \gls{llm} as a reasoning component while using a frozen vision encoder. Models like  BLIP-2~\cite{li2023blip} and Video-LLaMA~\cite{zhang2023video} extract image features using a ViT or CLIP encoder, pass them through a trainable alignment module such as \gls{qformer}, and feed the result into a frozen or partially fine-tuned \gls{llm} like  OPT and Vicuna. Although technically part of the broader category of Multimodal Large Language Models, we refer to them as \glspl{vlm} in this survey, in line with their origin in the vision–language literature.

\begin{table*}[htp]
\centering
\caption{Summary of Scenario-Generation Studies Using Vision Language Models.}
\label{tab:vlm_sum}
\begin{threeparttable}
\resizebox{1\textwidth}{!}{%
\begin{tabular}{l c c l p{2.0cm}  p{1.4cm} p{1.5cm} p{1.4cm} p{1.3cm} c l}
\toprule
\multirow{2}{*}{\textbf{Category}} &
\multicolumn{5}{c}{\textbf{Input}} &
\multirow{2}{*}{\textbf{Model}} &
\multirow{2}{*}{\textbf{Technique\textsuperscript{\textbf{1}}}} &
\multirow{2}{*}{\textbf{Simulator}} &
\multirow{2}{*}{\textbf{Output\textsuperscript{\textbf{2}}}} &
\multirow{2}{*}{\textbf{Paper}} \\
\cmidrule(lr){2-6}
& \textbf{Text} & \textbf{Image} & \textbf{View Type}  & \textbf{Dataset} & \textbf{Database} &
& & & & \\
\midrule

\multirow{1}{*}{\textbf{\makecell{Safety-critical Scenario}}}
&\checkmark &\checkmark &BEV of Metadrive &Waymo Open & &\makecell{GPT-4o \\ LLaVA} &CoT &Metadrive &\trajectory &CurricuVLM~\cite{sheng2025curricuvlm}\\
% &\cellcolor{LightGray}\checkmark &\cellcolor{LightGray}\checkmark &\cellcolor{LightGray}&\cellcolor{LightGray}\makecell{Argoverse2\cite{wilson2023argoverse}\\SUMO}  &\cellcolor{LightGray}\makecell{Map\\ Initial States}  &\cellcolor{LightGray}\makecell[{{l}}]{Claude-3-opus \\ GPT-4-turbo } &\cellcolor{LightGray}RAG  &\cellcolor{LightGray}SMARTS\cite{zhou2021smarts} &\cellcolor{LightGray}\script  &\cellcolor{LightGray}DriveGen~\cite{zhang2025drivegen} \\
\hline
\multirow{2}{*}{\textbf{\makecell{Real-world Replication}}}
&\cellcolor{LightGray}\checkmark &\cellcolor{LightGray}\checkmark  &\cellcolor{LightGray}&\cellcolor{LightGray}SUMO  &\cellcolor{LightGray}\makecell{Road\\ Network}   &\cellcolor{LightGray} \makecell[{{l}}]{GPT-4 \\ GPT-4V } &\cellcolor{LightGray}\makecell{CoT \\ RAG}  &\cellcolor{LightGray} SUMO &\cellcolor{LightGray}\script &\cellcolor{LightGray}OmniTester~\cite{lu2024multimodal} \\
& &\checkmark  &Real FPV  &\makecell{CCD\cite{bao2020uncertainty}} & &GPT-4o &ICL &CARLA &\script &Miao et al.~\cite{miao2024dashcam} \\
\hline
\multirow{1}{*}{\makecell{\textbf{Image Dataset}}}
&\cellcolor{LightGray}\checkmark  &\cellcolor{LightGray}   &\cellcolor{LightGray} &\cellcolor{LightGray}In-house &\cellcolor{LightGray}&\cellcolor{LightGray}DALL-E2 &\cellcolor{LightGray}ICL &\cellcolor{LightGray} &\cellcolor{LightGray}\image &\cellcolor{LightGray}WEDGE~\cite{marathe2023wedge}\\
\hline
\multirow{1}{*}{\textbf{\makecell{ADAS  Testing Scenario}}}
&\checkmark  &\checkmark  &BEV of Sketch &\makecell{nuScenes\cite{caesar2020nuscenes}} &NHTSA &GPT-4o &\makecell{CoT \\ ICL} &\makecell{Metadrive \\ BeamNG} &\script &TRACE~\cite{luo2025accidents} \\
\bottomrule
\end{tabular}}
\begin{tablenotes}
\scriptsize 
\item[1] Techniques: 
CoT = Chain-of-Thought prompting; 
ICL = In-Context Learning; 
RAG = Retrieval-Augmented Generation. 
\item[2] Output: \image ~Image, \trajectory ~Trajectory, \script ~Scenario script.
\end{tablenotes}
\end{threeparttable}
\end{table*}
\subsection{VLM-based Scenario Generation}

This subsection reviews how \glspl{vlm} are used to generate driving scenarios by leveraging their understanding of visual and textual inputs. \revB{For consistency, the category definitions in this subsection follow the conceptual distinctions introduced in Section III-\ref{subsec:scenario_generation}.} We group recent works into four categories and display them in \autoref{tab:vlm_sum}: %\textit{Safety-critical Scenario}, \textit{Real-world Scenario}, \textit{Dataset}, and \textit{ADAS Test Scenario}. These applications demonstrate the versatility of \glspl{vlm} in creating diverse, interpretable, and simulation-ready scenarios for autonomous driving.

\textbf{Safety-critical Scenario Generation:}
% \label{Sec4B:Corner} - label no longer usable
Safety-critical scenario generation is a rapidly advancing application of \glspl{vlm} in autonomous driving. It enables the synthesis of rare but relevant situations that are essential for evaluating system robustness. \revB{By combining visual perception with semantic reasoning, \glspl{vlm} have the potential to identify abnormal behaviors or near-failure conditions and generate targeted, interpretable scenarios.}

Recent frameworks such as CurricuVLM\cite{sheng2025curricuvlm} illustrate the potential of \glspl{vlm}.
\rev{CurricuVLM integrates \glspl{vlm} such as LLaVA into an online curriculum-learning loop. The \gls{vlm} analyzes \gls{bev} images and task descriptions to detect safety-critical events, while GPT-4o performs batch-level pattern analysis to reveal behavioral weaknesses. These insights guide a pre-trained DenseTNT model to generate tailored agent trajectories, and reinforcement learning adaptively selects the next scenarios.}

%\vspace{1mm}
However, CurricuVLM employs pre-trained \glspl{vlm}, thus its performance in identifying safety-critical agents is limited. Future work could explore combining these frameworks with safety-aware, fine-tuned \glspl{vlm} and incorporating temporal and multi-sensor contexts to improve reliability.
% \textbf{Metrics:} The evaluation here only focuses on \textit{safety-critical metrics} such as collision rate and driving distance~\cite{sheng2025curricuvlm}, to evaluate the severity and effectiveness of the generated scenarios.

\textbf{Real-World Scenario \rev{Replication}:}
% \label{Sec4B:Realworld} - label no longer usable
\glspl{vlm} offers new opportunities for realistic driving scenario \rev{replication}, by combining language understanding with visual modalities such as scenario images, enabling the creation of realistic traffic scenes \rev{based on real-world recorded dataset or maps.}

OmniTester~\cite{lu2024multimodal} proposes a framework with \gls{llm} and \gls{vlm} to create realistic and diverse traffic scenarios in SUMO. User inputs and context from \gls{rag} with external knowledge and \gls{osm} map library are processed via GPT-4~\cite{achiam2023gpt} to generate SUMO scenario scripts. A GPT-4V analyzes the generated scenario using images and code, providing feedback in natural language. Then, the GPT-4 evaluator compares this feedback against the intended description to enhance scenario generation.
Beyond the real-world map,  the authors from~\cite{miao2024dashcam} present a fully automated pipeline that transforms sample frames of dashcam crash video from the \gls{ccd}~\cite{bao2020uncertainty} into simulation scenarios for \gls{adas} testing. Using GPT-4o with \gls{icl}, the system generates SCENIC scripts for CARLA, while a second GPT-4o compares real and simulated video frames based on predefined behavior features, enabling iterative refinement through visual feedback.

Current approaches \rev{using maps and recorded videos} lack the use of real-world log replays, \rev{which is expected to enhance realism}.

% \textbf{Metrics:} In OmniTester~\cite{lu2024multimodal}, the evaluation focuses on two key aspects. (I) \textit{Controllability} measures how well the system adheres to user-specified requirements, including scene type, the number of lanes or vehicles, and whether the generated files are valid for execution in SUMO.  (II) \textit{Diversity} captures the range and variability of generated scenarios, evaluated through statistical analysis of lane counts, edge counts, route lengths, and vehicle numbers.

\textbf{Dataset Generation:}
% \label{Sec4B:Dataset} - label no longer usable
A key application of \glspl{vlm} is text-to-image generation to build tailored driving datasets, particularly to improve perception systems under diverse conditions.

WEDGE~\cite{marathe2023wedge} showcases the use of \glspl{vlm}, specifically DALL-E 2, to synthesize images depicting $16$ diverse and extreme weather conditions relevant to autonomous driving. Their dataset includes manually annotated 2D bounding boxes and is used to fine-tune object detectors. When evaluated on the real-world dataset, object detectors trained on WEDGE exhibit improved detection performance, highlighting the potential of \gls{vlm}-generated data for enhancing perception robustness in adverse conditions.

%\vspace{1mm}
 \rev{Currently, hybrid training that combines real and synthetic data is underexplored. This approach is crucial because real-world datasets often contain very few safety-critical corner cases, whereas synthetic data enables the controlled generation of rare events, such as crashes, occlusions, and anomalies—thereby improving long-tail coverage and strengthening model robustness in high-risk scenarios.}
%
%Inspired by advances in controllable diffusion models, future work could improve \glspl{vlm}' controllability by incorporating explicit scene structure, depth, or physical constraints as additional inputs, or by conditioning signals during generation.
% \textbf{Metrics:} (I) \textit{Perception metrics} such as average precision for object classes (e.g., truck, car); and (II) \textit{Image similarity metrics} like peak signal-to-noise ratio, structural similarity index, and root mean squared error, used to assess visual fidelity between synthetic and real-world images~\cite{marathe2023wedge}.

\textbf{Generation of \gls{adas} Test Scenarios:}
%\label{Sec4B:ADAS} - label no longer usable
\glspl{vlm} extend \gls{adas} scenario generation by grounding language in visual content, enabling semantically rich and visually faithful reconstructions of complex driving events. \revB{This facilitates the transformation of regulatory descriptions, test specifications, or crash reports into executable and reproducible scenarios for simulation-based evaluation of \gls{adas} performance.}

 TRACE~\cite{luo2025accidents} reconstructs \gls{adas} test scenarios from unstructured multimodal crash reports, including textual summaries and visual sketches. It uses GPT-4o with \gls{icl} and \gls{cot} to extract road types and environmental details from sketches. An \gls{llm}, built on GPT and augmented with trajectory data from nuScenes~\cite{caesar2020nuscenes}, generates realistic vehicle paths. These components are transformed into a DSL-based scenario compatible with simulators like MetaDrive using a rule-based encoder \rev{these scenarios are further utilized to test multiple \gls{adas} algorithms.}

%\vspace{1mm}
TRACE lack online interactive scenario editing, where users could modify scenes by sketching or annotating video frames, and \glspl{vlm} could dynamically update the simulation code. This would enable human-in-the-loop control and more flexible scenario refinement.

\begin{table*}[htp]
\centering
\caption{Summary of Scenario-Analysis Studies Using Vision Language Models.}
\label{tab:vlm_Analysis}
\begin{threeparttable}
\resizebox{1\textwidth}{!}{%
\renewcommand\arraystretch{1}
\begin{tabular}{p{2.0cm} >{\centering\arraybackslash}p{0.8cm} >{\raggedright\arraybackslash}p{1.4cm} >{\raggedright\arraybackslash}p{2.3cm}  >{\raggedright\arraybackslash}p{2cm} >{\raggedright\arraybackslash}p{1.5cm} >{\raggedright\arraybackslash}p{3.4cm}  >{\raggedright\arraybackslash}p{1.4cm} >{\raggedright\arraybackslash}p{1.6cm}  l}
\toprule
\multirow{2}{*}{\textbf{Category}} &
\multicolumn{3}{c}{\textbf{Input}} &
\multicolumn{3}{c}{\textbf{Model}} &
\multirow{2}{*}{\textbf{Technique\textsuperscript{\textbf{2}}}} &
\multirow{2}{*}{\textbf{Focus}} &
\multirow{2}{*}{\textbf{Paper}} \\
\cmidrule(lr){2-4} \cmidrule(lr){5-7}
& \textbf{Context} & \textbf{Image\textsuperscript{\textbf{1}}} & \textbf{Dataset} 
& \textbf{VLM} &\textbf{LLM} &\textbf{Role} & & \\
\midrule
\multirow{10}{*}{\textbf{\makecell{Visual\\Question\\Answering\\(VQA)}}}
&  &\makecell{Multi-view}  &nuScenes &\makecell{BLIP2\\InstructBLIP2\\MiniGPT4} &GPT-4 &\makecell{VLM: BEV Feature Extraction\\LLM: QA Execution} &Zero-shot &\makecell{Perception\\Prediction} &Talk2BEV~\cite{choudhary2024talk2bev}  \\
&\cellcolor{LightGray} \checkmark     &\cellcolor{LightGray}\makecell{Multi-view}  &\cellcolor{LightGray} nuScenes &\cellcolor{LightGray}  \makecell{ViT+OPT} &\cellcolor{LightGray} GPT-4 &\cellcolor{LightGray}\makecell{VLM: VQA Execution\\LLM: QA Generation} &\cellcolor{LightGray}\makecell{MLP\\Fft} &\cellcolor{LightGray} \makecell{Perception}  &\cellcolor{LightGray}NuScenes-MQA~\cite{inoue2024nuscenes}  \\
&\checkmark    &\makecell{Multi-view} &\makecell{nuScenes} &\makecell{EVA-02-L\\+\\Llama2-7B} &GPT-4 &\makecell{VLM: VQA Execution\\LLM: QA Augumentation} &\makecell{MLP\\Q-Former\\Fft} &\makecell{Counterfactual\\ Reasoning} &OmniDrive~\cite{wang2024omnidrive}\\
&\cellcolor{LightGray}\checkmark    &\cellcolor{LightGray}FPV &\cellcolor{LightGray}\makecell{nuScenes\\Waymo Open\\Once\cite{mao2021one}} &\cellcolor{LightGray}\makecell{FlanT5-XL\\+\\Vicuna-7B} &\cellcolor{LightGray}GPT-4 &\cellcolor{LightGray}\makecell{VLM: VQA Execution\\LLM: QA Augumentation} &\cellcolor{LightGray}\makecell{Q-Former\\Tokenizer\\LoRA} &\cellcolor{LightGray}\makecell{Perception\\Prediction\\Reasoning} &\cellcolor{LightGray}Reason2Drive~\cite{nie2024reason2drive}  \\
  &\checkmark &\makecell{Multi-view}   &\makecell{nuScenes} &\makecell{InterVL2.5-8B} &GPT-4o &\makecell{VLM: VQA Execution\\LLM: QA Generation} &\makecell{LoRA} & \makecell{Perception\\Prediction\\Planning} &DriveLMM-o1~\cite{ishaq2025drivelmm} \\
&\cellcolor{LightGray}\checkmark     &\cellcolor{LightGray}FPV &\cellcolor{LightGray}\makecell{DriveLM\cite{sima2024drivelm}\\LingoQA\cite{marcu2024lingoqa}\\NuScenes-QA\cite{qian2024nuscenes}} &\cellcolor{LightGray}\makecell{Qwen2-VL-7B\\Qwen2-VL-72B\\GPT-4o} &\cellcolor{LightGray}GPT-4o &\cellcolor{LightGray}\makecell{VLM: VQA Execution\\LLM: Multi-Choice QA} &\cellcolor{LightGray}\makecell{Zero-shot} &\cellcolor{LightGray}\makecell{Perception\\Prediction\\Planning} &\cellcolor{LightGray}AutoDrive-QA~\cite{khalili2025autodrive} \\
\hline
\multirow{18}{*}{\textbf{\makecell{Scene\\ Understanding}}}
&\checkmark    &\makecell{Multi-view} &\makecell{Waymo Open} &\makecell{CLIP} &&&\makecell{Zero-shot} &\makecell{ Tagging} &Najibi et al.~\cite{najibi2023unsupervised}\\
&\cellcolor{LightGray}\checkmark    &\cellcolor{LightGray}\makecell{Multi-view}&\cellcolor{LightGray}\makecell{SemanticKITTI\cite{behley2019semantickitti}} &\cellcolor{LightGray}\makecell{Grounding DINO\\+ SAM} &\cellcolor{LightGray}GPT-3.5 &\cellcolor{LightGray}\makecell{VLM: Object Grounding \\LLM: Narrative Generation} &\cellcolor{LightGray}\makecell{Zero-shot} &\cellcolor{LightGray}\makecell{Tagging} &\cellcolor{LightGray}OpenAnnotate3D~\cite{zhou2024openannotate3d}  \\
&\checkmark    &FPV &\makecell{Cityscapes\cite{cordts2016cityscapes}\\CamVid\cite{brostow2008segmentation}\\CARLA} &\makecell{ImageGPT} &&&\makecell{Zero-shot} &\makecell{Tagging} &Kou et al.~\cite{kou2025enhancing}\\
&\cellcolor{LightGray}\checkmark    &\cellcolor{LightGray}\makecell{Multi-view} &\cellcolor{LightGray}\makecell{DriveLM} &\cellcolor{LightGray}\makecell{ViT-B/32\\+\\T5-Base/Large} &\cellcolor{LightGray}&\cellcolor{LightGray}&\cellcolor{LightGray}\makecell{MLP\\Fft/LoRA} &\cellcolor{LightGray}\makecell{Tagging} &\cellcolor{LightGray}EM-VLM4AD~\cite{gopalkrishnan2024multi} \\
&\checkmark   &FPV &FARS & \makecell{GPT-4V\\LLaVA-13B} & \makecell{Llama2-13B\\Zephyr-7b-$\alpha$}&\makecell{VLM: Scene Captioning \\LLM: Risk Assessment} &\makecell{Zero-shot} & \makecell{Captioning} &Zarz\`a et al.~\cite{de2023llm}  \\
&\checkmark    &\makecell{Roadcamera} &\makecell{RDD\cite{maeda2018road}} &\makecell{GPT4} &&&\makecell{Zero-shot} &\makecell{ Captioning} &\rev{ConnectGPT~\cite{tong2024connectgpt}}\\
&\cellcolor{LightGray}\checkmark    &\cellcolor{LightGray}\makecell{BEV} &\cellcolor{LightGray}\makecell{WOMD\cite{ettinger2021large}} &\cellcolor{LightGray}\makecell{GPT-4V} &\cellcolor{LightGray}&\cellcolor{LightGray}&\cellcolor{LightGray}\makecell{Zero-shot} &\cellcolor{LightGray}\makecell{Captioning} &\cellcolor{LightGray}Zheng et al.~\cite{zheng2024large} \\
&\checkmark     &FPV &BDD100K\cite{yu2020bdd100k} &\makecell{Multiple VLMs} & & & Zero-shot&\makecell{Tagging\\Reasoning} &Rivera et al.~\cite{rivera2025scenario}\\
&\cellcolor{LightGray}\checkmark    &\cellcolor{LightGray}FPV &\cellcolor{LightGray}\makecell{nuScenes\\BDD-X\cite{kim2018textual}\\CDD\cite{bao2020uncertainty}} &\cellcolor{LightGray}\makecell{GPT-4V} &\cellcolor{LightGray}&\cellcolor{LightGray}&\cellcolor{LightGray}\makecell{Zero-shot} &\cellcolor{LightGray}\makecell{Reasoning} &\cellcolor{LightGray}Wen et al.~\cite{wen2023road} \\
&\checkmark     &\makecell{Multi-view} &MAPLM-QA\cite{cao2024maplm} & \makecell{ViLA} & &&Zero-shot&\makecell{Reasoning} & Keskar et al.~\cite{Keskar_2025_WACV}   \\
\hline
\multirow{9}{*}{\textbf{\makecell{Benchmark\\ \& Dataset\\}}}
&\cellcolor{LightGray}\checkmark     &\cellcolor{LightGray}\makecell{FPV} &\cellcolor{LightGray}DriveLM &\cellcolor{LightGray} \makecell{Multiple VLMs} &\cellcolor{LightGray} GPT-4o &\cellcolor{LightGray}\makecell{VLM: VQA Execution\\LLM: Answer Evaluation} &\cellcolor{LightGray}\makecell{Zero-shot} &\cellcolor{LightGray}\makecell{Robustness} &\cellcolor{LightGray}DriveBench~\cite{xie2025vlms}   \\
&\checkmark   &\makecell{Multi-view}   &\makecell{nuScenes} &\makecell{ViT/V2-99\\+ LLaVA-1.5-7B} &&\makecell{} &\makecell{Tokenizer\\MLP\\LoRA} &\makecell{3D Grounding} &NuGrounding~\cite{li2025nugrounding} \\
&\cellcolor{LightGray}\checkmark    &\cellcolor{LightGray}FPV &\cellcolor{LightGray}\makecell{nuScenes\\CARLA} &\cellcolor{LightGray} \makecell{BLIP2} &\cellcolor{LightGray} &\cellcolor{LightGray} &\cellcolor{LightGray}LoRA &\cellcolor{LightGray} \makecell{GVQA} &\cellcolor{LightGray}DriveLM~\cite{sima2024drivelm}  \\
&\checkmark     &FPV &\makecell{CODA\cite{li2022coda}} &\makecell{LLaVA-llama-3-8B\\GPT-4o} &GPT-4 &\makecell{VLM: Scene Captioning\\LLM: Caption Evaluation} &\makecell{LoRA} &\makecell{Corner Cases} &CODA-LM~\cite{chen2025automated} \\
&\cellcolor{LightGray}\checkmark     &\cellcolor{LightGray}\makecell{FPV} &\cellcolor{LightGray}In-house &\cellcolor{LightGray} \makecell{GPT-4o} &\cellcolor{LightGray} &\cellcolor{LightGray}&\cellcolor{LightGray}\makecell{CP, ICL, CoT} &\cellcolor{LightGray}\makecell{ADAS-LKA} &\cellcolor{LightGray}OpenLKA~\cite{wang2025openlka}\\
\addlinespace[-3pt]
\hline
\multirow{12}{*}{\textbf{\makecell{Risk Assessment}}}
&\checkmark   &\makecell{Multi-view}   &\makecell{In-house} &\makecell{GPT-4V} &&\makecell{} &\makecell{CP\\CoT} &\makecell{Risk  Scoring} &Hwang et.al~\cite{hwang2024safe} \\
&\cellcolor{LightGray}\checkmark   &\cellcolor{LightGray}\makecell{FPV}   &\cellcolor{LightGray}\makecell{DAD\cite{chan2017anticipating}} &\cellcolor{LightGray}\makecell{Flamingo} &\cellcolor{LightGray}GPT-3.5&\cellcolor{LightGray}\makecell{} &\cellcolor{LightGray}\makecell{Zero-shot} &\cellcolor{LightGray}\makecell{Hazard \\Explanation} &\cellcolor{LightGray}Latte~\cite{zhang2025latte} \\
&\checkmark   &\makecell{FPV}   &\makecell{CARLA} &\makecell{DINOv2\\OWLV2+SAM2\\GPT-4o} &&\makecell{} &\makecell{CP} &\makecell{Anomaly\\ Detection} &Ronecker et.al~\cite{ronecker2025vision} \\
&\cellcolor{LightGray}\checkmark   &\cellcolor{LightGray} \makecell{Multi-view}  &\cellcolor{LightGray}CARLA &\cellcolor{LightGray} \makecell{InternViT} &\cellcolor{LightGray}Interlm2-chat &\cellcolor{LightGray}\makecell{VLM: Video Extraction\\LLM: Narrative Generation} &\cellcolor{LightGray}\makecell{MLP\\QLoRA\\CP, CoT} &\cellcolor{LightGray} \makecell{Situation\\ Awareness \\Reasoning} &\cellcolor{LightGray}Think-Driver~\cite{zhangthink} \\
&\checkmark   &\makecell{Partially \\occluded BEV}   &\makecell{CARLA} &\makecell{Llama3.2-11B\\LLaVA-1.6-7B\\Qwen2-VL-7B} &&\makecell{} &\makecell{LoRA\\CP} &\makecell{Uncertainty \\Scoring} &Lee et.al~\cite{lee2025sff} \\
&\cellcolor{LightGray}\checkmark   &\cellcolor{LightGray}\makecell{FPV}   &\cellcolor{LightGray}\makecell{BDD100K} &\cellcolor{LightGray}\makecell{Qwen2-VL7B} &\cellcolor{LightGray}&\cellcolor{LightGray}\makecell{} &\cellcolor{LightGray}\makecell{LoRA} &\cellcolor{LightGray}\makecell{Hazard \\Detection} &\cellcolor{LightGray}INSIGHT~\cite{chen2025insightenhancingautonomousdriving} \\
&\checkmark   &\makecell{FPV}   &\makecell{OpenLKA\cite{wang2025openlka}} &\makecell{Qwen2.5-VL-3B\\Qwen2.5-VL-7B} &&\makecell{} &\makecell{LoRA} &\makecell{LKA Failures \\Prediction} &LKAlert~\cite{wang2025bridging}\\
\addlinespace[-5pt]
\bottomrule
\end{tabular}
}
\begin{tablenotes}
\scriptsize 
\item[1] Image: FPV = First Person View; BEV = Bird Eye View.
\item[2] Techniques: Fft = Full fine-tuning; CP = Contextual Prompting; ICL = In-Context Learning;   CoT = Chain-of-Thought prompting; Tokenizer = Prior Tokenizer;\\MLP = Multi-Layer Perceptron mapping.
\end{tablenotes}
\end{threeparttable}
\end{table*}

\subsection{VLM-based Scenario Analysis}
\label{subsec:VLM-based Scenario Analysis}
The current progress in scenario generation with \glspl{vlm} is quite at the beginning, but \glspl{vlm} have already shown big promises for scenario analysis in \gls{ad}.
Examples include NuScenes-QA~\cite{qian2024nuscenes} for \glspl{vqa}, where a \gls{vlm} answers natural language questions grounded in driving scenes to support scenario analysis; NuPrompt~\cite{wu2025language} for language-guided tracking and prediction, and Refer-KITTI~\cite{wu2023referring} for multi-object referring tracking tasks. However, these models are not considered foundation models because they do not utilize fully pre-trained foundation architectures. Rather, they construct task-oriented frameworks based on \gls{llm} backbone components.

In this section, we focus on foundation \glspl{vlm} pre-trained on large-scale, diverse image–text datasets with cross-domain generalization. We examine their potential to improve transferability, explainability, and efficiency in analyzing complex \gls{ad} scenarios. We structure our discussion around four key application areas, and show their techniques and applications in \autoref{tab:vlm_Analysis}.

\textbf{Visual Question Answering (VQA):}
\gls{vqa} datasets for autonomous driving pair visual inputs with natural language queries, to evaluate scene understanding across tasks such as perception, prediction, and planning. While recent works have proposed \gls{vqa} datasets, some \gls{qa} remain conceptual or require human reasoning in their creation, whereas others make use of \glspl{llm} for automated generation. This section focuses on \gls{vqa}-based scenario analysis methods \rev{that involve \gls{vlm} execution.}

Early efforts began by enriching existing scene representations with the \textit{perception} task. Talk2BEV~\cite{choudhary2024talk2bev} uses a perception stack to generate \gls{bev} maps by fusing multi-view images and LiDAR, then applies BLIP-2 to augment these maps with object-level language descriptions. These descriptions are passed to GPT-4 with \gls{cot} prompting to answer spatial and semantic queries, enabling zero-shot \gls{vqa} with annotated \gls{qa} pairs focusing on perception and prediction.
Similarly, NuScenes-MQA~\cite{inoue2024nuscenes} uses GPT-4 to automatically generate diverse question templates within the Markup-QA scheme.  The authors fully fine-tune a \gls{vlm} that combines a \gls{clip}-pre-trained \gls{vit} as a visual encoder, and OPT as a language model, using an \gls{mlp} to align multi-camera visual features with text. This setup enables joint evaluation of caption generation and visual question answering in driving scenarios for perception.

Later works moved toward more advanced \textit{reasoning} tasks. OmniDrive~\cite{wang2024omnidrive} introduces the first 3D \gls{vqa} dataset for counterfactual reasoning in autonomous driving, evaluating \glspl{vlm} with frozen EVA-02-L and Llama2–7B backbones, and using either an \gls{mlp} projector (Omni-L) or a \gls{qformer} (Omni-Q) as trainable modality bridges.
Reason2Drive~\cite{nie2024reason2drive} presents a video–text \gls{vqa} dataset composed of sequential images from nuScenes, Waymo, and ONCE~\cite{mao2021one}, covering tasks in perception, prediction, and reasoning. The authors fine-tune a \gls{vlm} consisting of FlanT5-XL and Vicuna-7B by using \gls{lora}, leveraging a prior tokenizer and an instructed vision decoder. A \gls{qformer} module is employed to jointly predict answers and perceptual cues.

Recent works in \gls{vqa}-based scenario analysis focus on advancing multimodal reasoning and evaluation across \textit{perception, prediction}, and \textit{planning} tasks in autonomous driving. DriveLMM-o1~\cite{ishaq2025drivelmm} introduces a step-by-step reasoning dataset based on nuScenes, incorporating both images and LiDAR points into the \gls{qa} context. Their \gls{qa} pairs are initially generated using GPT-4 and subsequently refined through human annotation. The authors fine-tune InternVL2.5-8B using \gls{lora}, demonstrating improved performance on reasoning and final answer accuracy across perception, prediction, and planning.
AutoDrive-QA~\cite{khalili2025autodrive} converts open-ended \gls{qa} pairs from DriveLM~\cite{sima2024drivelm}, LingoQA~\cite{marcu2024lingoqa}, and NuScenes-QA~\cite{qian2024nuscenes} into multiple-choice questions using GPT-4o, adding distractors, which are plausible but incorrect answer choices designed to reflect realistic domain-specific errors, to simulate realistic errors. This forms a standardized benchmark to evaluate pre-trained \glspl{vlm} across key scenario analysis tasks across perception, prediction, and planning. 

Despite these advances, most of the current \gls{vqa}s overlook traffic rules and real-world driving conventions. Future work should incorporate \rev{traffic rule-aware} \gls{qa}, grounded in road semantics (e.g., right-of-way rules and road signal compliance), to enable more realistic and safety-relevant scenario reasoning.
% \textbf{Metrics:} 
% (I) \textit{Answer Accuracy}, based on binary or multiple-choice correctness in response to perception, planning, and reasoning questions~\cite{inoue2024nuscenes, choudhary2024talk2bev, ishaq2025drivelmm};
% and (II) \textit{Language Generation Quality}, measured using BLEU, CIDEr, METEOR, and ROUGE-L scores to evaluate caption and reasoning fluency~\cite{inoue2024nuscenes, nie2024reason2drive};

\textbf{Scene Understanding:}
% \label{Sec4C:Scene} - label no longer usable
\glspl{vlm} are heavily used to interpret complex driving scenarios.
Recent works have leveraged \glspl{vlm} for \textit{scene tagging}, which represents the most basic level of scene understanding, involving binary or categorical assignments. Scene tagging assigns predefined labels at either the scene level (e.g., to analyze the weather conditions), or at pixel level (semantic segmentation) to characterize visual content for downstream tasks. 
Najibi et al.~\cite{najibi2023unsupervised} leverage a pre-trained \gls{clip} to perform zero-shot scene tagging on camera images, assigning semantic labels that are projected onto LiDAR points. These labels guide the generation of 3D pseudo-labels, which are then used to train a 3D object detector without human annotations.
OpenAnnotate3D~\cite{zhou2024openannotate3d} introduces an auto-labeling system for multi-modal 3D data, using GPT-3.5 for interpreting natural language scene descriptions and a \gls{vlm} with Grounding DINO and SAM for generating dense 2D masks, which are fused spatio-temporally and projected into 3D annotations.
Kou et al.~\cite{kou2025enhancing} propose a framework to enhance \gls{vlm}s for street scene semantic understanding. They use a pre-trained ImageGPT to extract semantic features from \gls{fpv} images, and  train a lightweight perception head that maps the semantic features to pixel-wise semantic segmentation masks.
EM-VLM4AD~\cite{gopalkrishnan2024multi} proposes a lightweight \gls{vlm} trained on the dataset from DriveLM~\cite{sima2024drivelm} with a primary focus on scenario tagging. It uses a \gls{vit} image encoder and explores two adaptation strategies: full fine-tuning of T5-base and \gls{lora}-based tuning of T5-large. The model is benchmarked against baselines in terms of parameter count, \gls{flops}, and memory usage, showcasing strong efficiency for deployment in resource-constrained settings.

Building on scene tagging, recent efforts have advanced toward the intermediate-level task of \textit{scene captioning}, which bridges perception and language by generating open-form descriptions. Scene captioning generates concise natural language descriptions of visible elements.  
Zarzà et al.~\cite{de2023llm} propose a framework using structured inputs with principal component analysis, and adopt Llama2-13B with \gls{cot} and \gls{cp} to assess the risks in a scenario, suggesting driving adaptations. They test their framework with the FARS dataset\footnote{\url{https://www.nhtsa.gov/research-data/fatality-analysis-reporting-system-fars}}. Additionally, they leverage a \gls{vlm}, specifically LLaVA-13B with \gls{cp}, to perform image-based scenario captioning, enhancing scene understanding through natural language descriptions.
\revB{ConnectGPT~\cite{tong2024connectgpt} leverages \glspl{vlm} to generate standardized Cooperative Intelligent Transport Systems (C-ITS) messages for Connected and Automated Vehicles. GPT-4 is used to interpret infrastructure camera images, generate C-ITS messages, with validation conducted on a small curated set of highway images, including samples originating from the Road Damage Dataset (RDD)~\cite{maeda2018road}.}

Zheng et al.~\cite{zheng2024large} introduce a context-aware motion prediction framework using \glspl{vlm}. They employ GPT-4V to extract traffic context from a transportation context map. They combine vector map data and historical trajectories, and feed the generated scenario description into a motion transformer to improve trajectory prediction.

Several studies address the most advanced form of scene understanding: \textit{scene reasoning}, which requires interpreting interactions, causality, and abstract situational context. Scene reasoning interprets relationships and interactions among agents while producing coherent narratives that capture intent, causality, and situational context.
Rivera et al.~\cite{rivera2025scenario} propose a scalable pipeline for traffic scene classification using off-the-shelf \glspl{vlm} such as GPT-4V, LLaVA, and CogAgent-\gls{vqa} \cite{Hong2024}. These models are evaluated zero-shot to reason about predefined scenario elements, such as lane markings and vehicle maneuvers, using self-developed and the BDD100K \cite{yu2020bdd100k} datasets.
Wen et al.~\cite{wen2023road} explore GPT-4V's zero-shot capability for road scene interpretation from dashcam footages, evaluating the model on object detection, scene captioning, \gls{vqa}, and causal reasoning, while highlighting its potential and limitations for autonomous driving.
Keskar et al.~\cite{Keskar_2025_WACV} evaluate NVIDIA's ViLA on the MAPLM-QA~\cite{cao2024maplm} benchmark for traffic scene understanding. Using contextual prompting, they assess ViLA on multiple-choice \gls{vqa} tasks, including lane counting, intersection detection, scene classification, and point cloud quality assessment. ViLA shows strong performance on high-level \gls{vqa} tasks but struggles with fine-grained spatial reasoning.
%%\vspace{1mm}
%Future work should enhance spatial grounding (e.g., by using BEV maps and 3D scenarios), integrate temporal reasoning, and establish systematic benchmarks that span tagging, captioning, and reasoning to enable a systematic evaluation and deeper multimodal understanding.
% \textbf{Metrics:} The evaluation of \gls{vlm}-based scenario understanding uses the following metrics. (I) \textit{Scene tagging}: common metrics are \textit{Intersection-over-Union} and mean F1 score~\cite{zhou2024openannotate3d, kou2025enhancing}, which quantify the segmentation performance, and the BLEU-4, METEOR, ROUGE-L, and CIDEr metrics to assess the textual tagging outputs~\cite{gopalkrishnan2024multi}.
% Additionally, mean average precision is used for evaluating 3D object detection from pseudo-labels~\cite{najibi2023unsupervised}.
% (II) \textit{Scene captioning:} human or LLM-based evaluation of causal coherence~\cite{de2023llm}, and functional evaluation via intention-prediction accuracy, confusion matrices, and ablation studies on prompt designs~\cite{zheng2024large}.
% (III) \textit{Scene reasoning:} the evaluation includes attribute classification in zero-shot setups~\cite{rivera2025scenario}, task-specific accuracy on structured VQA tasks such as lane counting and intersection detection~\cite{Keskar_2025_WACV}, and rubric-based assessments of causal reasoning and multimodal consistency~\cite{wen2023road}.

\textbf{Benchmarks \& Datasets:}
% \label{Sec4C:Benchmark} - label no longer usable
To support the development and evaluation of \glspl{vlm} in autonomous driving, recent efforts have introduced specialized \textit{benchmarks} and curated \textit{datasets} covering key tasks such as perception, prediction, planning, and scenario reasoning under real-world and safety-critical conditions.

Aiming for a standardized evaluation, several works present benchmarks aligned with diverse driving scenarios.
DriveBench~\cite{xie2025vlms} introduces a benchmark for evaluating scenario reasoning across multiple driving tasks. It extends the \gls{vqa} dataset from DriveLM \cite{sima2024drivelm} and adds diverse visual corruption categories to assess the model's robustness. Using this benchmark, the authors evaluate the robustness of a range of pre-trained and fine-tuned \glspl{vlm} (e.g., GPT-4o, Qwen2-VL~\cite{wang2024qwen2}) under clean, corrupted, and text-only conditions. GPT-4o is further employed as an automatic evaluator for open-ended answers.
nuGrounding~\cite{li2025nugrounding} proposes the first 3D visual grounding benchmark with human-annotated object grounding based on nuScenes. The authors fine-tune LLaVA-1.5 using \gls{lora}, with \gls{vit} or V2-99 as the visual encoder. To incorporate 3D understanding, they extract \gls{bev} features via a \gls{bev}-based detector, map them into the \gls{llm} adapter, and fuse them with \gls{vlm} outputs through a query fuser for accurate object detection and localization.

Complementing these benchmarks, other works provide high-quality datasets to train and adapt \glspl{vlm} to complex driving environments.
DriveLM~\cite{sima2024drivelm} introduces a graph-structured visual question answering (GVQA) \rev{which leverages graph-based scene representations to answer structured perception, prediction, and planning questions in autonomous driving scenarios}, using human-curated \gls{qa} graphs from nuScenes and rule-based annotations from CARLA. A BLIP-2-based \gls{vlm} is fine-tuned with \gls{lora} and guided by graph-based question prompting to enable zero-shot interpretable scenario reasoning across perception, prediction, and planning.
CODA-LM~\cite{chen2025automated} introduces a corner-case image-text dataset derived from the CODA dataset~\cite{li2022coda}. The authors use GPT-4V to generate multi-task captions spanning perception, prediction, and planning for each image. These captions are then evaluated and refined using GPT-4. After constructing the dataset, they fine-tune a LLaVA-llama-3-8B model to enhance vision language understanding in corner-case driving scenarios.
OpenLKA~\cite{wang2025openlka} introduces a large-scale, real-world dataset for Lane Keeping Assist under diverse driving conditions. GPT-4o is used in conjunction with \gls{cp}, \gls{cot}, and \gls{icl} to generate structured scene annotations that describe lane quality, weather, and traffic context.

%\vspace{1mm}
However, the existing benchmarks and datasets still lack realism and diversity. For example, DriveBench exposes the \gls{vlm}'s vulnerability to corruption, suggesting the need for more realistic disturbances (e.g., occlusions, night-time). CODA-LM relies on filtered GPT captions, underscoring the gap in real-world edge-case coverage. %Moreover, future work should expand ADAS-focused datasets beyond lane keeping to cover a wider range of scenarios.
% \textbf{Metrics:}The evaluation of benchmarks focuses on  (I)~\textit{robustness}, which measures the VQA performance under clean, corrupted, and text-only inputs~\cite{xie2025vlms}. Another used metric is (II)~\emph{grounding precision}, which evaluates 3D mean Average Precision to evaluate 3D object detection performance and alignment accuracy for spatial localization and multimodal fusion~\cite{li2025nugrounding}. The evaluation of datasets focuses on (III)~\emph{answer fidelity}, to assess the QA correctness and consistency via graph-guided prompts~\cite{sima2024drivelm}; (IV)~\emph{caption quality}, which uses BLEU-4, METEOR, ROUGE-L, and CIDEr~\cite{chen2025automated}; (V)~\emph{annotation coverage}, which measures attribute diversity and alignment with telemetry data~\cite{wang2025openlka}.

\textbf{Risk Assessment:}
% \label{Sec4C:Risk} - label no longer usable
\glspl{vlm} are increasingly applied to autonomous driving risk assessment, addressing tasks like hazard detection, uncertainty estimation, and failure prediction. Recent approaches leverage both prompting and fine-tuning and use diverse visual inputs, including \gls{bev} maps, multi-view images, and segmentation masks. These methods aim to improve safety through interpretable reasoning and context-aware decision support.

Recent advances have explored prompting techniques for risk analysis.
Hwang et al.~\cite{hwang2024safe} utilize GPT-4V in a zero-shot setting for risk scoring in street-crossing scenarios. The model receives structured visual inputs, including bounding boxes, segmentation masks, and optical flow, alongside contextual prompts formulated using \gls{cot}. Instead of directly processing raw images, GPT-4V reasons over augmented visual features to assess safety levels and provide natural language justifications. 
Similarly, LATTE~\cite{zhang2025latte} introduces a real-time hazard detection framework that utilizes off-the-shelf computer vision modules and three lightweight attention modules for spatial reasoning, temporal modeling, and risk prediction. Upon hazard detection, Flamingo and GPT-3.5 are triggered to generate scene captions and verbal explanations. The system operates in a zero-shot manner by leveraging contextual prompting for situational reasoning. 
For anomaly object detection, Ronecker et al.~\cite{ronecker2025vision} proposed both patch-based and instance-based embedding methods using vision foundation models, evaluated on a CARLA-based dataset. They leverage the zero-shot capabilities of DINOv2 for visual embeddings and combine OWLv2 with SAM2 for object-level instance segmentation. Their instance-based approach achieves slightly better results than GPT-4o using contextual prompting.

%In contrast, other works adopt fine-tuning strategies to enhance risk assessment.
Think-Driver~\cite{zhangthink} proposes a \gls{vlm} that uses multi-view images to assess perceived traffic conditions and evaluate the risks of current driving maneuvers. It employs multi-view RGB inputs and ego state data, processed by InternViT and InterLM2-chat, respectively. The model is fine-tuned using \gls{qlora} and trained on \gls{cot}-style \gls{qa} data that cover scene understanding, hazard reasoning, and action prediction. 
In consideration of occlusion-aware \gls{bev} representations, Lee et al.~\cite{lee2025sff} first investigate the use of \gls{vlm} for uncertainty prediction in autonomous driving. They construct a dataset from CARLA using \gls{bev} images that contain occlusion masks, paired with driving actions and uncertainty scores. Three \glspl{vlm} are fine-tuned using \gls{lora} to compare their performance under occluded conditions.
For hazard detection and explanation, INSIGHT~\cite{chen2025insightenhancingautonomousdriving} fine-tunes Qwen2-VL-7B via \gls{lora}. Using annotated hazard locations in BDD100K images, the model is trained to localize high-risk regions and generate natural language descriptions. It outperforms several pre-trained \glspl{vlm} in both spatial localization and interpretability tasks. 
Finally, LKAlert~\cite{wang2025bridging} develops a \gls{vlm}-based framework for predicting lane-keeping assist failures. It integrates RGB dashcam images, CAN bus signals, and lane segmentation masks from LaneNet. A Qwen2.5-VL model is fine-tuned via \gls{lora}, with lane masks serving as spatial guidance. The model outputs binary alerts and interpretable explanations to enhance safety transparency.

%\vspace{1mm}
To enable real-world deployment, the inference latency and resource demands need to be further  reduced through model compression, efficient prompting, and lightweight \gls{vlm} architectures optimized for onboard execution in autonomous vehicles.

\subsection{Limitations and Future Directions}

\textbf{\gls{vlm}-based Scenario Generation:}
Compared to \gls{llm}-based scenario generation (Section III~\ref{subsec:scenario_generation}), \glspl{vlm} remain underexplored in areas such as scenario synthesis for training driving policies and closed-loop scenario generation. With their ability to process both visual and textual inputs, \glspl{vlm} offer a powerful extension to existing frameworks. A promising direction is to use them as auxiliary analysis modules to improve the interpretability and fidelity of the generated scenarios, while also providing feedback signals to iteratively enhance the scenario quality.

Moreover, there is strong potential to develop more sophisticated and interdisciplinary pipelines that fully leverage the multimodal reasoning capabilities of \glspl{vlm}. For instance, in scenario-based testing, real-world traffic videos could be interpreted by \glspl{vlm} to produce detailed scene captions. These captions could serve as structured conditions for \glspl{dm} to regenerate photorealistic driving scenes or videos. Such a multi-stage pipeline, linking perception, semantic understanding, and simulation, represents a promising direction for building holistic and scalable scenario generation systems.

\textbf{\gls{vlm}-based Scenario Analysis:}
In the domain of scenario analysis, \glspl{vlm} show advantages over text-only \gls{llm}-based frameworks. Current research follows two main trends.

The first trend centers on developing task-specific frameworks, often augmented with external computer vision modules (e.g., for 3D grounding or hazard detection). Meanwhile, the rapid progress of general-purpose pre-trained \glspl{vlm} raises a key research question: to what extent can these models handle scenario analysis effectively without relying on external tools like object detectors, depth estimators, or 3D grounders? Investigating the capabilities and limitations of such end-to-end \glspl{vlm} could enable more streamlined, scalable solutions that reduce the system's complexity while preserving, or even enhancing, their analytical performance.

The other trend emphasizes \gls{vqa}, designing tailored \gls{vqa} tasks that fine-tune \glspl{vlm} for improved task-oriented performance. Despite recent advances, several challenges persist. While large-scale pre-trained \glspl{vlm} exhibit strong potential, the scenario analysis pipeline in autonomous driving remains highly complex and poorly standardized. Specifically, there is a lack of benchmark datasets, consistent annotation frameworks for \gls{vqa} tasks, and unified evaluation metrics tailored to scenario analysis. Addressing these gaps is essential for developing more robust and task-specific \glspl{vlm} capable of handling real-world autonomous driving scenarios.

\hypertarget{sec:mllm}{}
\section{Multimodal Large Language Models (MLLMs)}
\label{sec:mllm}
This section begins with the development of \glspl{mllm}, highlighting their architectural evolution and adaptation techniques, such as modality bridging and instruction tuning. Then, it covers scenario generation from multimodal input and scenario analysis tasks, including \gls{vqa}, scene understanding, and risk assessment in \gls{ad} contexts.

\subsection{Development of MLLMs}
\glspl{mllm} extend pre-trained \glspl{llm} by integrating three or more modalities, such as vision, audio, and video, enabling the system to reason over richer and more diverse sensory inputs beyond image–text pairs.
Early \glspl{vlm} such as BLIP-2 \cite{li2023blip}
use frozen vision backbones connected to \glspl{llm} via adapters such as \gls{qformer}. These models primarily extend \glspl{vlm} by connecting frozen perception encoders to \glspl{llm}, but do not yet constitute full \glspl{mllm}.
In parallel, early multimodal extensions such as Video-LLaMA~\cite{zhang2023video} incorporated additional modalities, including video and audio, enabling joint reasoning over text, visual frames, and acoustic signals. Although these models marked an initial step toward \glspl{mllm}, they typically relied on frozen backbones and lacked unified multimodal training, resulting in limited temporal and cross-modal reasoning capabilities.

More recent models, including GPT-4o, represent a further step toward fully unified \glspl{mllm} by integrating vision and audio within a single model. 
Similarly, Google Gemini~\cite{team2023gemini} and Qwen-Omni~\cite{xu2025qwen2} natively support multiple modalities, enabling open-ended reasoning over images, videos, audio, and structured visual content, such as charts and diagrams. While effective for general visual-language tasks, these models fall short in autonomous driving, which requires reasoning over structured inputs like temporal object tracks, \gls{bev} layouts, and interaction-aware motion patterns. \revB{The progression from \glspl{llm} to \glspl{mllm} is depicted in Figure~\ref{fig: LLM_timeline}.}

To address the unique demands of autonomous driving, recent \gls{mllm} architectures have begun incorporating structured, domain-specific modalities such as multi-view video, LiDAR point clouds, and \gls{bev} layouts. These additions enable spatial and temporal grounding, allowing \glspl{llm} to reason more effectively over complex driving scenes and multi-agent dynamics~\cite{park2025nuplanqa}.
\begin{figure}[h]
    \centering
    \includegraphics[width=1\linewidth]{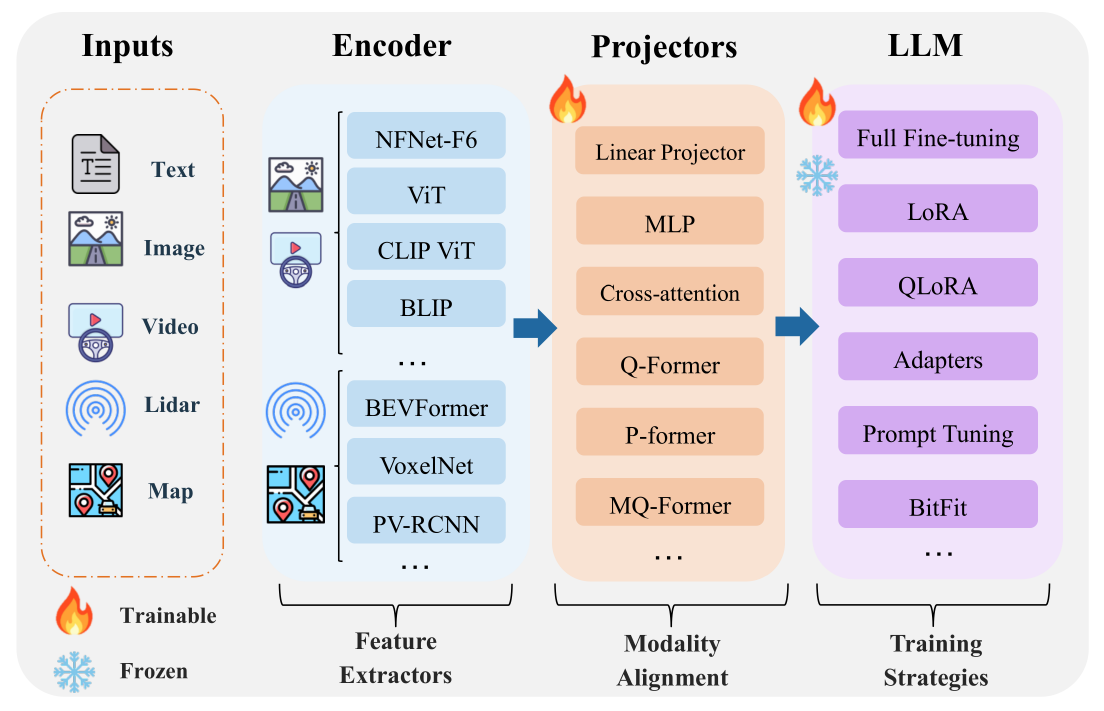}
    \caption{Overview of adaptation techniques for \glspl{mllm} in autonomous driving. \textbf{Encoders} extract features from modality-specific inputs. \textbf{Projectors} are trainable modules that map features into the \gls{llm}’s embedding space to enable cross-modal alignment. The \textbf{\gls{llm}} serves as the reasoning core and can be frozen or trainable, depending on the available resources and the task, using fine-tuning techniques.}
    \label{fig: MLLM_Tech}
\end{figure}

%\textbf{Adaptation Techniques for \glspl{mllm}:}
%\label{sec:mllm-adapt} - label no longer usable
While building on techniques from \glspl{vlm}, \glspl{mllm} are adapted to support a broader range of modalities essential for autonomous driving, such as video, LiDAR point clouds, \gls{bev} maps, and \gls{hd} semantic features. As illustrated in \autoref{fig: MLLM_Tech}, these systems typically consist of specialized modality encoders (e.g., BEVFormer~\cite{ding2024holistic}, CLIP-ViT~\cite{ishaq2025tracking}, VoxelNet~\cite{yang2025lidar}), projection modules to align multi-modal features (e.g., \glspl{mlp}, \gls{qformer}, cross-attention), and task-specific training strategies.
\glspl{mllm} often keep both the perception and language backbones frozen, with adaptation focusing on lightweight bridging and instruction tuning for downstream driving-related tasks. The main adaptation strategies are discussed in the following.

\textbf{Modality alignment modules:} These modules serve as a bridge between non-text modalities and the \gls{llm}'s token space. The main modality alignment modules are:

\emph{(1) Linear projector:} a single linear (i.e., fully connected) neural layer used to project modality-specific features into the \gls{llm}'s embedding space. It offers a lightweight mapping strategy and is often used in early-stage \glspl{vlm} or in combination with pre-trained encoders~\cite{xu2024drivegpt4, park2024vlaad}.

\emph{(2) \gls{mlp} projection:} Projects high-dimensional features from vision or spatial encoders (e.g., \gls{vit}, BEVFormer) into the \gls{llm}'s token space. Used in models such as BLIP-2 \cite{li2023blip} and driving-centric adapters like P-Adapter~\cite{ding2023hilm}, which align \gls{bev} or LiDAR features for language-based reasoning.

\emph{(3) \gls{st}-Adapter:} A lightweight temporal adapter module is used to extend image-based \glspl{mllm} to process sequential video inputs~\cite{ding2024holistic, ding2023hilm}. It enables spatiotemporal modeling without modifying the core \gls{llm} weights.

\emph{(4) Cross-attention:} Uses learnable queries to attend over image or point cloud tokens, enabling multimodal fusion for tasks such as spatial/temporal grounding, semantic alignment, and instruction following~\cite{ma2024dolphins, fan2024mllm}.

\emph{(5) \gls{qformer}:} Transformer-based query modules that distill task-relevant embeddings from multi-modal inputs using cross-attention. These modules are applied in BLIP-2 \cite{li2023blip}, InternDrive \cite{zhang2024interndrive}, and NuInstruct \cite{ding2024holistic} for structured fusion across video, LiDAR, and \gls{bev} inputs.

\emph{(6) Fusion transformers:} Specialized attention blocks designed to integrate features across multiple streams such as \gls{bev} maps, multi-view video, or LiDAR point clouds. Modules like BEV-Injection~\cite{ding2024holistic} serve as fusion transformers by aligning and injecting multi-modal features (e.g., from images or LiDAR) into a unified \gls{bev} representation. These are commonly used in driving-centric \glspl{mllm}.

\emph{(7) Structure-aware encoder:} A module that converts structured perception inputs, such as 3D bounding boxes~\cite{yang2025lidar}, scene graphs, or motion trajectories, into token embeddings suitable for language-based reasoning. 

\textbf{Multimodal fine-tuning:} Once modality alignment is achieved, an \gls{mllm} can be trained to follow task-specific prompts using paired instruction data like \gls{vqa}. This stage teaches the model to reason over multimodal contexts and produce grounded outputs. Similarly to \glspl{vlm}, two main strategies are commonly employed to achieve this adaptation: 

\textit{(1) \gls{peft}:} 
\gls{peft} strategies adapt \glspl{mllm} by updating only a small subset of the model's parameters, typically keeping the \gls{llm} frozen. While classical \gls{peft} methods such as adapter layers, \gls{lora}, and prompt tuning (discussed below) operate inside the \gls{llm}, recent works in autonomous driving often apply \gls{peft} to modality alignment modules~\cite{ding2024holistic}. For example, components like \gls{st}-Adapters and \glspl{qformer} are trained to bridge visual or spatial inputs to the \gls{llm}, enabling task adaptation without modifying the core language model.

\emph{Adapter layers:} Lightweight trainable modules inserted between the layers of an \gls{llm}, typically using a down-projection and up-projection structure. They are used in LLaMA Adapter V2 \cite{Gao2023_llama} and InternDrive \cite{zhang2024interndrive}.

\emph{\gls{lora}:} Applies low-rank updates to attention and feed-forward modules. Frequently used in driving models like DriveGPT4~\cite{xu2024drivegpt4}.
% \emph{Prompt tuning:} Uses learnable soft tokens to inject modality-specific context, which is especially helpful for rapid domain adaptation with limited resources.

\emph{\gls{peft}-\gls{ma}:} Fine-tuning only the \gls{ma} modules (e.g., \gls{qformer}, \gls{st}-Adapter), while keeping the \gls{llm}'s weights frozen.

\textit{(2) \gls{fft}:} 
Full fine-tuning updates all model parameters, including vision encoders, spatial encoders, and the \gls{llm}. While this approach typically yields the highest task-specific performance, it is computationally intensive. To reduce the computational cost, some works apply \gls{fft} to smaller models, for example using Qwen2-0.5B~\cite {zhou2025tumtraffic}.

\begin{table*}[htp]
\centering
\caption{Summary of Scenario-Generation Studies Using Multimodal Large Language Models.}
\label{tab:mllm_sum}
\begin{threeparttable}
\resizebox{1\textwidth}{!}{%
\begin{tabular}{l c c l p{1.7cm}  p{1.7cm} p{1.5cm} p{1.5cm} p{1.5cm} c l}
\toprule
\multirow{2}{*}{\textbf{Category}} &
\multicolumn{5}{c}{\textbf{Input}} &
\multirow{2}{*}{\textbf{Model}} &
\multirow{2}{*}{\textbf{Technique}} &
\multirow{2}{*}{\textbf{Simulator}} &
\multirow{2}{*}{\textbf{Output\textsuperscript{\textbf{1}}}} &
\multirow{2}{*}{\textbf{Paper}} \\
\cmidrule(lr){2-6}
& \textbf{Text} &  \textbf{Image} & \textbf{Video}  & \textbf{Dataset} & \textbf{Database} &
& & & & \\
\midrule

\multirow{1}{*}{\textbf{\makecell{Safety-critical Scenario}}}
&   \checkmark   
&   \checkmark 
& Real FPV
& \makecell{SUMO \\ CARLA\\OSM}
&  \makecell{NHTSA\\GPS}
& \makecell{GPT-4o} 
& CoT 
& CARLA 
& \script
& AutoScenario~\cite{lu2024realistic}\\
\hline
\multirow{1}{*}{\textbf{\makecell{ADAS Testing Scenario}}}
&\cellcolor{LightGray}\checkmark &\cellcolor{LightGray}  &\cellcolor{LightGray}Real FPV  &\cellcolor{LightGray} HDD\cite{ramanishka2018toward} &\cellcolor{LightGray}   &\cellcolor{LightGray}  \makecell{GPT-4V} &\cellcolor{LightGray}\makecell{CoT \\ ICL} &\cellcolor{LightGray} LGSVL &\cellcolor{LightGray}\script &\cellcolor{LightGray}LEADE~\cite{tian2024llm} \\
\bottomrule
\end{tabular}}
\begin{tablenotes}
\scriptsize
\item[1] Output icons: \image ~Image, \trajectory ~Trajectory, \script ~Scenario script.
\end{tablenotes}
\end{threeparttable}
\end{table*}
\subsection{MLLM-based Scenario Generation}
\glspl{mllm} can jointly process diverse visual inputs from vehicle sensors or human sources, enabling a comprehensive understanding of complex driving environments. Their ability to integrate multiple modalities also supports the generation of more realistic, context-aware scenarios. In the following, we categorize the use of \glspl{mllm} into two groups, summarized in Table~\ref{tab:mllm_sum}. \revB{For consistency, the scenario categories follow the definitions introduced in Section III~\ref{subsec:scenario_generation}.}

\textbf{Safety-critical Scenario Generation:}
% \label{Sec5B:Corner} - label no longer usable
\revB{\gls{mllm}-based safety-critical scenario generation focuses on synthesizing rare and high-risk driving situations by leveraging heterogeneous modalities such as videos, GPS traces, and crash reports. By reasoning jointly over spatial, temporal, and semantic cues, these methods reconstruct or generate corner cases that closely reflect real-world hazardous events.}

AutoScenario~\cite{lu2024realistic} presents a pipeline to generate realistic corner cases using multimodal crash data from NHTSA, including text, images, videos, and semi-structured reports. 
% Videos are downsampled and enhanced with depth maps to extract motion and layout using a memory-based interpreter. 
They use GPT-4o with \gls{cot} to generate structured scenario descriptions, which are then used to produce road networks in SUMO and agent behaviors in CARLA. Their scenario refinement is guided by GPS traces and frame-level similarity between simulated and real scenes, to ensure good matching with the original crash event. A promising future direction is to incorporate spatial modalities, such as LiDAR point clouds or \gls{bev} maps, to achieve more accurate scene geometries and agents' localization, enhancing realism beyond what 2D video and depth sensing alone can offer.
% \textbf{Metrics:} (I)~\textit{Realism:} accuracy of the generated scene type, vehicles count, and object attributes compared to crash report descriptions;
% (II)~\textit{Diversity:} variation in structural elements such as lane count, agent density, and route length;
% (III)~\textit{Controllability:} success rate of generating executable and constraint-compliant scenarios, which was evaluated for instance in SUMO and CARLA.

\textbf{\gls{adas} Testing Scenario Generation:}
%\label{Sec5B:ADAS} - label no longer usable
\revB{Scenario generation for \gls{adas} testing with \glspl{mllm} aims to derive executable and reproducible test cases from real-world multimodal observations, primarily targeting function-level validation of \gls{adas} stack under typical but diverse driving conditions. }
LEADE~\cite{tian2024llm} generates \gls{adas} test scenarios from real-traffic videos in the HDD dataset~\cite{ramanishka2018toward}. Key frames are used in multimodal \gls{icl} and \gls{cot} prompting with GPT-4V to create abstract scenarios, which are converted into executable programs for the LGSVL simulator \cite{rong2020lgsvl}. The Apollo \gls{adas} stack~\cite{ap} runs an ego vehicle, and a dual-layer search identifies semantic-equivalent scenarios that expose behavioral differences between Apollo and human drivers. Future work could align scenario generation with \gls{adas} test standards, enabling the synthesis of regulation-compliant scenarios. Incorporating traffic rules and structured priors would also improve controllability and test coverage.

% in~\cite{tian2024llm}
% \textbf{Metrics:} \quad 
% The evaluation of \gls{mllm}-generated ADAS test scenarios in LEADE~\cite{tian2024llm} focuses on the following. 
% (I)~\textit{Accuracy:} element-wise extraction accuracy across categories like  behavior, location, and object types from the video input;
% (II)~\textit{Correctness:} success rate of concrete scenario execution matching abstract semantics across different road types;
% (III)~\textit{Violation discovery:} quantifies the ability of the generated scenarios to expose safety-critical differences between the ADAS behavior and human driver behavior, including the number and diversity of detected violations.
\begin{table*}[htp]
\centering
\caption{Summary of Scenario-Analysis Studies Using Multimodal Large Language Models.}
\label{tab:mllm_Analysis}
\begin{threeparttable}
\resizebox{0.98\textwidth}{!}{%
\renewcommand\arraystretch{1}
\begin{tabular}{p{1.9cm} >{\centering\arraybackslash}p{0.5cm} c c  >{\centering\arraybackslash}p{0.5cm}  >{\centering\arraybackslash}p{0.5cm}  >{\raggedright\arraybackslash}p{1.4cm}  >{\raggedright\arraybackslash}p{1.8cm} >{\raggedright\arraybackslash}p{1.5cm} >{\raggedright\arraybackslash}p{2.9cm}  >{\raggedright\arraybackslash}p{1.8cm} >{\raggedright\arraybackslash}p{1.6cm}> {\raggedright\arraybackslash}p{2cm}}
\toprule
\multirow{2}{*}{\textbf{Category}} &
\multicolumn{6}{c}{\textbf{Input}} &
\multicolumn{3}{c}{\textbf{Model}} &
\multirow{2}{*}{\textbf{Technique\textsuperscript{\textbf{3}}}} &
\multirow{2}{*}{\textbf{Focus\textsuperscript{\textbf{4}}}} &
\multirow{2}{*}{\textbf{Paper}} \\
\cmidrule(lr){2-7} \cmidrule(lr){8-10}
& \textbf{Image} &\textbf{Context} &\textbf{Lidar} & \textbf{Video\textsuperscript{\textbf{1}}}&\textbf{Map} & \textbf{Dataset} 
& \textbf{MLLM} &\textbf{LLM} &\textbf{Role\textsuperscript{\textbf{2}}} & & \\
\midrule

\multirow{28}{*}{\textbf{\makecell{Visual\\Question\\Answering\\(VQA)}}}
& & \checkmark &  &\makecell{FPV} &  & BDD-X &  \makecell{CLIP\\+Llama2} &GPT-4  &\makecell{MLLM: VideoQA Exec.\\LLM: QA Gen.} &\makecell{Projector\\LoRA} & \makecell{Perception\\Reasoning\\Control} & DriveGPT4~\cite{xu2024drivegpt4} \\
&\cellcolor{LightGray}\checkmark  &\cellcolor{LightGray}\checkmark  &\cellcolor{LightGray} &\cellcolor{LightGray}FPV &\cellcolor{LightGray} &\cellcolor{LightGray}\makecell{BDD-X\\HDD} &\cellcolor{LightGray} \makecell{BLIP2\\+\\Llama2-7B} &\cellcolor{LightGray}GPT-4 &\cellcolor{LightGray}\makecell{MLLM: VideoQA Exec.\\LLM: QA Aug.} &\cellcolor{LightGray}\makecell{QueryTrans\\Projector\\PEFT-MA} &\cellcolor{LightGray} \makecell{Prediction\\Reasoning} &\cellcolor{LightGray}VLAAD~\cite{park2024vlaad}\\
&\checkmark  &\checkmark &   &FPV &\checkmark &\makecell{In-house} & \makecell{CLIP\\+\\Vicuna1.5-7B} &GPT-4 &\makecell{MLLM: VideoQA Exec.\\LLM: QA Gen.} &\makecell{QueryTrans\\ Projector} & \makecell{Prediction\\Reasoning\\Control} & LingoQA~\cite{marcu2024lingoqa} \\
&\cellcolor{LightGray}\checkmark  &\cellcolor{LightGray}\checkmark &\cellcolor{LightGray}   &\cellcolor{LightGray}Roadside &\cellcolor{LightGray} &\cellcolor{LightGray}\makecell{TUMTraffic\\
VideoQA} &\cellcolor{LightGray} \makecell{SigLIP\\+\\Qwen2-0.5B/7B} &\cellcolor{LightGray}GPT-4omini &\cellcolor{LightGray}\makecell{MLLM: VideoQA Exec.\\LLM: QA Gen.} &\cellcolor{LightGray}\makecell{MLP\\PEFT-MA\\Fft} &\cellcolor{LightGray} \makecell{ ST Reasoning} &\cellcolor{LightGray} \makecell{TUMTraffic\\VideoQA~\cite{zhou2025tumtraffic}} \\
&\checkmark  &\checkmark &   &\makecell{Multi\\View} &\checkmark &\makecell{NuPlan} & \makecell{BEV Encoder\\Llama3.2V-11B} &GPT-4o &\makecell{MLLM: VideoQA Exec.\\LLM: MC-QA Gen.} &\makecell{MLP\\FusionTrans\\PEFT-MA} & \makecell{Perception\\ST Reasoning} & NuPlanQA~\cite{park2025nuplanqa}\\
&\cellcolor{LightGray} &\cellcolor{LightGray} \checkmark &\cellcolor{LightGray}  &\cellcolor{LightGray}\makecell{Multi\\View} &\cellcolor{LightGray}  &\cellcolor{LightGray} nuScenes &\cellcolor{LightGray}  \makecell{Video-Llama} &\cellcolor{LightGray}  &\cellcolor{LightGray}\makecell{MLLM: VideoQA Exec.} &\cellcolor{LightGray}\makecell{Cross-attention\\ ST-Adapter\\QueryTrans\\FusionTrans \\PEFT-MA} &\cellcolor{LightGray} \makecell{Perception\\Prediction\\Reasoning\\Risk} &\cellcolor{LightGray} NuInstruct~\cite{ding2024holistic} \\
&\checkmark  & \checkmark &   &\makecell{Multi\\View} &  & DRAMA\cite{malla2023drama} &  \makecell{ViT\\+\\MiniGPT-4} &GPT-4o  &\makecell{MLLM: VideoQA Exec.\\LLM: VQA Aug.} &\makecell{ST-Adapter\\MLP\\ Cross-Attention\\PEFT-MA} & \makecell{Perception\\Prediction\\Reasoning\\Risk} & HiLM-D~\cite{ding2023hilm} \\
&\cellcolor{LightGray}\checkmark  &\cellcolor{LightGray}\checkmark &\cellcolor{LightGray}   &\cellcolor{LightGray}FPV &\cellcolor{LightGray} &\cellcolor{LightGray}\makecell{SHRP2\cite{hankey2016description}} &\cellcolor{LightGray} \makecell{LLaMA-VID-7B\\14 MLLMs} &\cellcolor{LightGray}\makecell{GPT-o1\\Qwen2.5-72B} &\cellcolor{LightGray}\makecell{MLLM: VideoQA Exec.\\LLM: MC-QA Gen.} &\cellcolor{LightGray}\makecell{Fft\\ICL} &\cellcolor{LightGray} \makecell{Perception\\Reasoning\\Risk} &\cellcolor{LightGray} DVBench~\cite{zeng2025vision}\\
&  & \checkmark &  \checkmark & &  & nuScenes &  \makecell{Voxel\\+\\Llama2-7B} &  &\makecell{MLLM: VQA Exec.} &\makecell{Encoder\\QueryTrans\\MLP\\PEFT-Adapter} & \makecell{Grouding\\Captioning} & Lidar-llm~\cite{yang2025lidar} \\
&\cellcolor{LightGray}\checkmark  &\cellcolor{LightGray}\checkmark &\cellcolor{LightGray}\checkmark  &\cellcolor{LightGray} &\cellcolor{LightGray}\checkmark &\cellcolor{LightGray}In-house &\cellcolor{LightGray} \makecell{CLIP\\+\\ Llama2-7B} &\cellcolor{LightGray} &\cellcolor{LightGray}\makecell{MLLM: VQA Exec.} &\cellcolor{LightGray}\makecell{Projector\\LoRA} &\cellcolor{LightGray} \makecell{Perception} &\cellcolor{LightGray} MAPLM~\cite{cao2024maplm}  \\
&\checkmark  &\checkmark &\checkmark  &\makecell{Roads\\FPV}  
&&\makecell{V2X-Real\\V2V4Real}
& \makecell{PointPilars\\+\\ LLaVA-v1.5-7b} & &\makecell{MLLM: VQA Exec.} &\makecell{Projector\\LoRA} & \makecell{Perception\\Planning} & \rev{V2V-LLM~\cite{chiu2025v2v}} \\
\hline
\multirow{15}{*}{\textbf{\makecell{Scene/Scenario\\ Understanding}}}
&\cellcolor{LightGray}\checkmark  &\cellcolor{LightGray} \checkmark &\cellcolor{LightGray}   &\cellcolor{LightGray} &\cellcolor{LightGray}  &\cellcolor{LightGray} nuScenes &\cellcolor{LightGray}  \makecell{InternVl-1.5} &\cellcolor{LightGray} GPT-4o &\cellcolor{LightGray}\makecell{MLLM: Scene Under. \\LLM: QA Gen.} &\cellcolor{LightGray}\makecell{LoRA} &\cellcolor{LightGray} \makecell{\rev{Scene}\\perception \\prediction\\Reasoning\\}  &\cellcolor{LightGray} InterDrive~\cite{zhang2024interndrive} \\
&\checkmark  &\checkmark &\checkmark  & & &\makecell{KITTI\cite{geiger2012kitti}\\nuScenes} & \makecell{Video-LLaVA\\GPT-4o} & & &\makecell{CoT} & \makecell{\rev{Scene} \\Reasoning} & Jain et al.~\cite{jain2024semantic} \\
&\cellcolor{LightGray}  &\cellcolor{LightGray}\checkmark &\cellcolor{LightGray}  &\cellcolor{LightGray}FPV &\cellcolor{LightGray} &\cellcolor{LightGray}BDD-X  &\cellcolor{LightGray} \makecell{VideoMA\\+Ada-002\\+OpenFlamingo} &\cellcolor{LightGray} &\cellcolor{LightGray}\makecell{} &\cellcolor{LightGray}\makecell{Cross-Attention\\LoRA} &\cellcolor{LightGray} \makecell{ \rev{Scenario} \\Reasoning} &\cellcolor{LightGray} Dolphins~\cite{ma2024dolphins}  \\
&\checkmark  &\checkmark &\checkmark  &\makecell{Multi\\View} & &\makecell{DriveLM} & \makecell{ViT-L/14\\Llama-Adapter V2} & & &\makecell{QueryTrans\\PEFT-Adapter\\PEFT-MA} & \makecell{ \rev{Scenario}\\Reasoning} & Ishaq et al.~\cite{ishaq2025tracking} \\
&\cellcolor{LightGray}  &\cellcolor{LightGray}\checkmark &\cellcolor{LightGray}  &\cellcolor{LightGray}FPV &\cellcolor{LightGray} &\cellcolor{LightGray}BDD100K  &\cellcolor{LightGray} \makecell{ViT-L/14 \\+Vicuna-7B} &\cellcolor{LightGray}GPT-3.5 &\cellcolor{LightGray}\makecell{MLLM: Video Capt. \\LLM: Caption Eval.} &\cellcolor{LightGray}\makecell{Projector\\QueryTrans\\PEFT-MA} &\cellcolor{LightGray} \makecell{\rev{Scenario}\\Captioning} &\cellcolor{LightGray} WTS~\cite{kong2024wts}  \\
&  &\checkmark & &FPV & &\makecell{LingoQA} & \makecell{LLaVA-VL-7B\\Qwen-VL-7B} & \makecell{Qwen2.5-1.5B\\Qwen2.5-7B}&\makecell{MLLM: Scene Extract.\\LLM: Scene Under.}  &\makecell{Zero-shot} & \makecell{ \rev{Scenario}\\Captioning} & V3LMA~\cite{lubberstedt2025v3lma} \\
\hline
\multirow{10}{*}{\textbf{\makecell{Risk Assessment}}}
&\cellcolor{LightGray}\checkmark  &\cellcolor{LightGray}\checkmark &\cellcolor{LightGray}  &\cellcolor{LightGray} &\cellcolor{LightGray}\checkmark &\cellcolor{LightGray}DeepAccident\cite{wang2024deepaccident}  &\cellcolor{LightGray} \makecell{GPT-4V} &\cellcolor{LightGray}GPT-4 &\cellcolor{LightGray}\makecell{MLLM: Image Extract.\\LLM: Narrative Gen.} &\cellcolor{LightGray}\makecell{Zero-shot} &\cellcolor{LightGray} \makecell{Accident\\ Prevention} &\cellcolor{LightGray} AccidentGPT~\cite{wang2023accidentgpt}  \\
&\checkmark  & \checkmark &   & &  & \makecell{DRAMA\\-ROLISP,\\DRAMA\\-SRIS\cite{ding2023hilm}} &  \makecell{ResNet-101\\+ Swin-L\\+Llama2-7B} & &\makecell{} &\makecell{QueryTrans\\Cross-Attention\\MLP\\PEFT-Adapter} & \makecell{Safety\\Interaction}  & MLLM-SUL~\cite{fan2024mllm} \\
&\cellcolor{LightGray}\checkmark  &\cellcolor{LightGray}\checkmark &\cellcolor{LightGray}  &\cellcolor{LightGray}FPV &\cellcolor{LightGray} &\cellcolor{LightGray}nuScenes &\cellcolor{LightGray} \makecell{VideoLlama2 } &\cellcolor{LightGray}Llama3.1-8B &\cellcolor{LightGray}\makecell{MLLM: Video Extract.\\LLM: Narrative Gen.} &\cellcolor{LightGray}Zero-shot &\cellcolor{LightGray} \makecell{Safety\\ Interaction} &\cellcolor{LightGray} ScVLM~\cite{shi2024scvlm}  \\
&\checkmark  &\checkmark &   &FPV & &\makecell{DRAMA} & \makecell{Gemini1.5V-Pro} & &\makecell{} &\makecell{ICL} & \makecell{Risk event\\Detection} & Abu et al.~\cite{abu2024using} \\

\bottomrule
\end{tabular}}
\begin{tablenotes}
\scriptsize 
\item[1] Video: FPV = First Person View; BEV = Bird Eye View;
\item[2] Role: Exec.= Execution; Aug. = Augmentation; Gen. = Generation; Under. = Understanding; Capt. = Captioning; Eval. = Evaluation; Extract. = Extraction;
\item[3] Techniques: Only focus on the techniques for MLLMs.
Projector = Linear projector; MLP = MLP projection;
QueryTrans = Query Transformer;\\ FusionTrans = Fusion Transformer; PEFT-Adapter = Adapter layers; Fft = Full fine-tuning;  PEFT-MA: Only trains modality alignment modules and LLMs are frozen;\\ Encoder = Structure-aware encoder; CP = Contextual Prompting; ICL = In-Context Learning; CoT = Chain-of-Thought prompting;
\item[4] Focus: ST Reasoning = Spatio-temporal reasoning.
\end{tablenotes}
\end{threeparttable}
\end{table*}
\subsection{MLLM-based Scenario Analysis}
\label{subsec:mllm-scenarioanalysis}
This section discusses the papers using \glspl{mllm} for scenario analysis in \gls{ad}. We categorize the existing works into three key tasks, as reported in Table \ref{tab:mllm_Analysis}.

\textbf{Visual Question Answering (VQA):}
%\label{Sec5C:VQA} - label no longer usable
In comparison with \gls{vqa}-based scenario analysis with \glspl{vlm}, \glspl{mllm} have extended capabilities to deal with multi-modal sensor data such as videos, LiDAR point clouds, and \gls{hd} maps \cite{Elghazaly2023}, besides images and text. 
Based on their task and data modality, existing \gls{vqa} datasets can be grouped into four categories:

%\vspace{1mm}
(I) \textit{General AD Tasks – Perception, Reasoning, and Control:}
Several datasets target core autonomous driving tasks, including visual perception, reasoning, and decision-making.
DriveGPT4~\cite{xu2024drivegpt4} introduces the first driving-specific, video \gls{qa}-style instruction-following dataset, generated using GPT-4 with structured inputs including object detection bounding boxes, captions, and control signals formatted as text. It fine-tunes a \gls{mllm} combining \gls{clip},   and LLaMA2 with \gls{lora} adapters to produce both textual explanations and control outputs. Meanwhile, a mix-finetuning strategy merges general visual instruction data with driving-specific samples to improve reasoning and performance.
VLAAD~\cite{park2024vlaad} introduces a multi-modal assistant for autonomous driving, trained on an instruction-following dataset derived from BDD-X and HDD videos, with \gls{qa} pairs augmented using GPT-4. The model is built on Video-LLaMA, which combines a BLIP-2-based visual encoder, a Video \gls{qformer} for temporal modeling, and a frozen LLaMA-2-7B language model. \gls{peft}-\gls{ma} is applied only to the \gls{qformer} and projection layers, enabling the model to efficiently perform tasks such as \gls{vqa}, free-form \gls{qa}, ego-intention prediction, and scenario-level reasoning.
LingoQA~\cite{marcu2024lingoqa} presents a \gls{vqa} dataset for autonomous driving, covering perception, reasoning and action. It includes an action set annotated with GPT-3.5 and a scenery set generated by GPT-4 using \gls{cot}. The baseline model processes video frames using \gls{clip} and a \gls{qformer}, with a linear projector to align features to Vicuna-1.5-7B's token space. For the fine-tuning,  \gls{peft}-\gls{ma} is applied to the \gls{qformer} and projector and the \gls{llm} remains frozen. Evaluation is conducted using the novel Lingo-Judge classifier, which is trained with \gls{lora}.

%\vspace{1mm}
(II) \textit{Spatio-Temporal Reasoning:}
Datasets in this group emphasize reasoning over agent motion, temporal dependencies, and event semantics in driving scenarios.
\revB{TUMTraffic-VideoQA~\cite{zhou2025tumtraffic} introduces a multiple-choice video \gls{qa} dataset for roadside traffic scenes, covering object captioning and spatio-temporal grounding, and facilitating fine-grained spatio-temporal reasoning in traffic scenarios.} Visual metadata is extracted using standard detectors and captioned by off-the-shelf \glspl{vlm}, while GPT-4o-mini generates \gls{qa} pairs via template-augmented prompting. The baseline model (TUMTraffic-Qwen) uses SigLIP for visual encoding, an \gls{mlp} projector for modality alignment, and Qwen2 (0.5B/7B) as the \gls{llm}, which is fully fine-tuned for instruction-following \gls{qa}.
NuPlanQA~\cite{park2025nuplanqa} introduces a video \gls{qa} dataset built on nuPlan, using GPT-4o to generate free-form \gls{qa} pairs for training and multiple-choice \gls{qa} for evaluation. To leverage this data, the authors propose BEV-LLM, an \gls{mllm} that integrates multi-view images and \gls{bev} features through a \gls{bev} encoder, a BEV-Fusion module, and an \gls{mlp} projector. The model uses LLaMA-3.2-Vision as a frozen backbone, while training is applied only to the BEV-Fusion module and projection layers, following a \gls{peft}-\gls{ma} strategy.

%\vspace{1mm}
(III) \textit{Risk-Aware Reasoning:}
To address safety-critical understanding, several datasets focus on risk recognition, intention estimation, and planning-related queries.
NuInstruct~\cite{ding2024holistic} introduces multi-view video \gls{qa} datasets covering perception, prediction, risk, and planning tasks. \glspl{qa} are generated via a structured \gls{sql} pipeline. The authors propose BEV-InMLLM, which extends \glspl{mllm} (e.g., Video-LLaMA) by utilizing \gls{st}-Adapters and a BEV-Injection module or a Fusion transformer that integrates spatial features from multi-view videos, resulting in improved performance on holistic autonomous driving tasks.
HiLM-D~\cite{ding2023hilm} introduces DRAMA-ROLISP, a risk-aware \gls{vqa} dataset \rev{for risk assessment} that is enhanced using GPT-4. The model fine-tunes MiniGPT-4 with a \gls{vit} and a \gls{st}-Adapter for video input, a ResNet-based encoder, and a P-Adapter for spatial fusion. A Query-Aware Detector integrates outputs for risk object localization and intention reasoning. The \gls{llm} itself remains frozen, with only the adapters, fusion, and projector layers fine-tuned. 
DVBench~\cite{zeng2025vision} introduces a comprehensive video-based \gls{vqa} benchmark for safety-critical autonomous driving, built on SHRP2~\cite{hankey2016description} dashcam data. Multiple-choice \gls{qa} pairs are generated and refined using GPT-4o and Qwen2.5-72B, covering perception and reasoning tasks, and classifying into 11 subcategories. The benchmark evaluates 14 \glspl{mllm} using the self proposed metric, which rotates answer positions to assess robustness. The authors also compare the performance of Qwen2-VL-2B/7B with and without full fine-tuning on the DVBench dataset.

%\vspace{1mm}
(IV) \textit{Multi-Modal Extensions with LiDAR (with/without \gls{hd} Maps:}
To extend reasoning beyond RGB data, some datasets incorporate 3D point clouds or \gls{hd} maps.
LiDAR-LLM~\cite{yang2025lidar} first tackles 3D captioning, grounding, and \gls{vqa} from LiDAR point clouds. It extracts \gls{bev} features via a voxel encoder, embeds them using a View-Aware Transformer with learnable queries, which acts as a prior tokenizer, and projects them into the language space through an \gls{mlp}. Adapter layers are fine-tuned within the \gls{llm} to support 3D scene understanding.
MAPLM~\cite{cao2024maplm} introduces a large-scale multimodal benchmark and \gls{vqa} dataset and focuses on perception and \gls{hd} map understanding in autonomous driving. It includes panoramic 2D images, \gls{bev} projections from LiDAR point clouds, and text descriptions extracted from \gls{hd} maps. The baseline model aligns visual features using pre-trained \gls{clip} encoders and lightweight projection adapters, mapping them into the \gls{llm}'s embedding space. Instruction tuning is performed via \gls{lora} on Vicuna or LLaMA-2, enabling the model to perform effective scene-level reasoning across modalities. V2V-LLM~\cite{chiu2025v2v} further extends LiDAR-based multimodal reasoning to cooperative driving by fusing point-cloud features from multiple connected vehicles. It constructs a Vehicle-to-Vehicle \gls{vqa} dataset based on dataset V2X-Real~\cite{xiang2024v2x, Xu2023-v2v4real} for perception and planning. The model adapts LLaVA by replacing RGB encoders with a LiDAR detector, aligning scene-level and object-level features through an \gls{mlp} projector and fine-tuned \gls{lora} layers.

%\vspace{1mm}
A key next step for \gls{vqa} in autonomous driving is to evaluate model robustness under out-of-distribution conditions. Current datasets mostly feature common driving scenarios and well-structured questions, leaving models largely untested on rare events, unfamiliar objects, or challenging conditions such as night, snow, or construction zones. Developing benchmarks that explicitly include these edge cases, and assessing how well models generalize to them, is essential for deploying \gls{vqa} systems in safety-critical, real-world environments.
% \textbf{Metrics:} (I) \textit{Answer Accuracy}, based on binary or multiple-choice correctness in response to task questions~\cite{ding2024holistic, cao2024maplm, park2024vlaad, zhou2025tumtraffic, park2025nuplanqa, zeng2025vision};
% (II) \textit{Language Generation Quality}, measured using BLEU, CIDEr, METEOR, SPICE, and ROUGE-L scores to assess the fluency, relevance, and completeness of generated captions or free-form responses\cite{yang2025lidar, marcu2024lingoqa, ding2023hilm, zhou2025tumtraffic};
% (III) \textit{Semantic and Spatial Grounding}, evaluated via metrics like mean Intersection-over-Union and L1/L2 distance errors to quantify grounding precision in risk detection and object localization tasks\cite{ding2023hilm, zhou2025tumtraffic};
% (IV) \textit{Robustness Evaluation}, performed using GroupEval\cite{zeng2025vision}, which tests model consistency by rotating the position of correct answers in multiple-choice settings.

\textbf{Scene/Scenario Understanding:}
%\label{Sec5C:Scene} - label no longer usable
% In autonomous driving, \glspl{mllm} play a crucial role in interpreting complex environments by integrating data from LiDAR, video, and \gls{hd} maps.
\rev{This subsection distinguishes between \textit{Scene Understanding}, which focuses on static, image-based perception, and \textit{Scenario Understanding}, which captures temporal dynamics, agent interactions, and evolving causal events.}

\textit{(I) Scene Understanding:} InternDrive~\cite{zhang2024interndrive} and Jain et al.~\cite{jain2024semantic} focus on static scene understanding using image-based inputs. 
InternDrive proposes a framework for driving scenario understanding, covering perception, prediction, and reasoning, using \gls{mllm}. It generates \gls{qa} pairs from nuScenes using GPT-4o, followed by human correction, and fine-tunes the \gls{mllm} InternVL-1.5 via \gls{lora} on these annotations. The resulting model analyzes driving scenes from \gls{fpv} images through visual instruction tuning.
Jain et al.~\cite{jain2024semantic} evaluate \gls{mllm} for safety-critical scene understanding using \gls{qa} pairs from KITTI and nuScenes across five categories. They benchmark Video-LLaVA and GPT-4V using merged image frames and textual LiDAR summaries, applying a \gls{cot} prompting approach to enhance multimodal reasoning without requiring true temporal modeling.

\textit{(II) Scenario Understanding:} In contrast, DOLPHINS\cite{ma2024dolphins}, Ishaq et al.\cite{ishaq2025tracking}, WTS\cite{kong2024wts}, and V3LMA\cite{lubberstedt2025v3lma} target scenario understanding, where temporal context, agent interaction, and causal reasoning are central. \rev{DOLPHINS~\cite{ma2024dolphins} presents an MLLM-based system for human-like understanding of driving scenarios and behaviors. The model is built on OpenFlamingo and first instruction-tuned on image–instruction pairs using a Grounded Chain of Thought (GCoT), where each reasoning step is explicitly linked to visual evidence to ensure visually grounded scenario reasoning. It is then adapted to driving videos using in-context examples retrieved by VideoMAE and Ada-002. During training, only the perceiver resampler, gated cross-attention, and \gls{lora} modules are updated, making the framework efficient while supporting multiple driving tasks.}
Ishaq et al.~\cite{ishaq2025tracking} propose a scenario-level spatial understanding framework that integrates short video clips, driving trajectories as text, and textual queries. They use a trajectory encoder and a Query Former to fuse the modalities, which are then passed into a frozen LLaMA-2 model with adapter layers. The model is fine-tuned by training both the Query Former and the adapters for efficient multimodal reasoning.

Specificly, WTS~\cite{kong2024wts} and V3LMA~\cite{lubberstedt2025v3lma} focus on the scenario captioning which emphasizes observable elements, reasoning targets spatial-temporal relations, intent inference, and causal analysis.
WTS~\cite{kong2024wts} uses GPT-3.5 externally to generate human-guided ground truth captions and evaluate model outputs via LLMScore, which assesses semantic and syntactic similarity. The proposed Instance-VideoLLM combines \gls{clip} ViT-L/14, a Video \gls{qformer}, and Vicuna-7B, with fine-tuning applied to the adapter and \gls{qformer}. The model is trained on enhanced video inputs incorporating bounding boxes, gaze data, and scene context, and is compared against other off-the-shelf \glspl{mllm}.
V3LMA~\cite{lubberstedt2025v3lma} proposes a fusion method that combines pre-trained \glspl{llm} and \glspl{vlm} to enhance zero-shot 3D scenario understanding. They use off-the-shelf tools for grounding, object detection, and depth estimation to generate structured scenario descriptions, which are fed into the \gls{llm}. Visual features from an \gls{mllm} are then fused at either the feature level or the classification head. Despite being zero-shot, the model achieves competitive performance, comparable to fine-tuned \glspl{mllm}.

%\vspace{1mm}
Current \glspl{mllm} for scene and scenario understanding primarily focus on short-term temporal contexts and curated question-answering tasks, \rev{which lacks validation in realistic, real-world settings. To move toward more comprehensive scenario understanding,} future work should explore long-range temporal modeling, causal inference across event sequences, and robust handling of out-of-distribution scenarios.
% \textbf{Metrics:} (I) \textit{Answer Accuracy}, which assesses correctness in structured tasks such as perception, planning, and reasoning~\cite{zhang2024interndrive, jain2024semantic, ishaq2025tracking};
% (II) \textit{Human Evaluation}, involving expert judgment of completeness, intent interpretation, and reasoning quality~\cite{ma2024dolphins, jain2024semantic, kong2024wts};
% (III) \textit{Spatial Reasoning Metrics}, which measure spatial understanding through localization accuracy~\cite{ishaq2025tracking}.

\textbf{Risk Assessment:}
%\label{Sec5C:Risk} - label no longer usable
The goals for the \glspl{mllm} include risk detection and violation inference for anticipating hazards, inference and scene-level safety scoring for analyzing incidents, and actionable advice generation. 

One approach to risk assessment emphasizes proactive hazard mitigation through interpretable scenario understanding. For example, AccidentGPT~\cite{wang2023accidentgpt} combines multi-modal perception, such as images, 3D detections, \gls{bev} features, and trajectories, with GPT-4V for zero-shot scenario captioning based on dataset DeepAccident~\cite{wang2024deepaccident} and GPT-4 for further safety evaluation using \gls{cot} and \gls{cp}. It supports real-time accident prevention, post-accident analysis, and interactive safety decision-making through interpretable reasoning. 

Other works focus on enabling interactive safety perception and feedback.
MLLM-SUL~\cite{fan2024mllm} fuses multi-scale visual inputs using ResNet-101 and Swin-L for low- and high-resolution features, combined via Query Formers and Gate-Attention based on the dataset Drama-ROLISP and Drama-SRIS from HiLM-D~\cite{ding2023hilm}. It fine-tunes LLaMA2-7B with adapters and applies an \gls{mlp} head for scene captioning and risk object localization. Similarly, 
ScVLM~\cite{shi2024scvlm} proposes a multi-stage \gls{mllm} framework for risk assessment based on the nuScenes dataset, combining event type classification, conflict type identification, and narrative generation. It uses VideoLLaMA2 for zero-shot visual context extraction and LLaMA 3.1 8B to generate detailed descriptions of safety-critical events based on \gls{fpv} driving videos.

A third direction emphasizes risk reasoning through structured question answering. 
Abu et al.~\cite{abu2024using} present a \gls{mllm}-based framework for safety-critical event detection using \gls{fpv} videos from the DRAMA dataset. They compare Gemini-Pro-V1.5, Gemini-Pro-Video, and LLaVA using \gls{qa}-based risk analysis with in-context learning, leveraging sliding window capture and textual context prompts to enhance risk event detection.

%\vspace{1mm}
However, to ensure practical impact, it is critical to establish the reliability and determinism of MLLM-based risk assessments. This remains a key challenge, as \glspl{mllm}' behavior is inherently stochastic and may produce inconsistent outputs.

\subsection{Limitations and Future Directions}
\rev{\glspl{mllm} offer a unique potential to generate and analyze scenarios by leveraging their multimodal capabilities. However, there is currently no pretrained \gls{mllm} specifically devised for \gls{ad} with complementary sensor modalities such as LiDAR, camera, and radar. As a future direction, this highlights the need for large-scale multi-modal datasets and pretrained \glspl{mllm} tailored to \gls{ad}.}

\textbf{\gls{mllm}-based Scenario Generation:}
As reported in Table \ref{tab:mllm_sum}, only two studies have explored \gls{mllm}-based scenario generation: one targeting safety-critical scenarios and the other focused on \gls{adas} testing. This highlights a significant research gap and suggests that the broader potential of \glspl{mllm} in this domain remains largely unexplored. Future work could extend to additional applications such as driving policy evaluation, closed-loop scenario generation, and the reconstruction of complex real-world driving events.

An emerging research direction is retrieval-augmented scenario generation. While existing retrieval-augmented generation frameworks are typically based on textual databases, \glspl{mllm} allow for the integration of multimodal knowledge bases containing maps, annotated traffic videos, and LiDAR point clouds. Such enriched context could support more diverse, realistic, and situation-aware scenario generation pipelines.

\textbf{\gls{mllm}-based Scenario Analysis:}
As summarized in Table~\ref{tab:mllm_Analysis}, current pre-trained \glspl{mllm} are not yet sufficient to address the complexity of driving scenario analysis. Existing models often struggle with specialized tasks that require aligning and processing diverse multimodal inputs. While fine-tuning strategies such as instruction tuning, adapter-based methods, and parameter-efficient techniques are being actively explored, these adaptations are often necessary because general-purpose pre-trained models lack sufficient domain-specific understanding. 
At the same time, several technical challenges must be tackled. Reliability remains a major concern, as \glspl{mllm} are prone to factual hallucinations and inconsistent output issues that are especially critical in safety-sensitive applications. Decreasing inference times is equally important. This may involve architectural innovations, model compression and distillation, or adaptation strategies that support interpretable, low-latency reasoning across multiple modalities.

From an application standpoint, promising directions include using \glspl{mllm} for high-fidelity sensor simulation and modeling complex interactions among diverse traffic participants, such as vehicles, pedestrians, and cyclists. Additionally, deploying \glspl{mllm} at the edge to support real-time situational awareness and collaborative human-machine interaction represents a valuable and unexplored opportunity for future research.

\hypertarget{sec:dm}{}
\section{Diffusion Models (DMs)}
\label{sec:dm}
This section provides an overview of \rev{\glspl{dm}}, explaining their underlying generative process and tracing their conceptual evolution. Given their generative nature, \glspl{dm} excel at synthesizing novel scenarios rather than analyzing existing ones. Accordingly, we survey their applications in scenario generation for \gls{ad}, encompassing traffic flow synthesis, road layout design, image generation, and video generation.

\subsection{Development of DMs}
\begin{figure}[ht]
    \centering
    \includegraphics[width=1\linewidth]{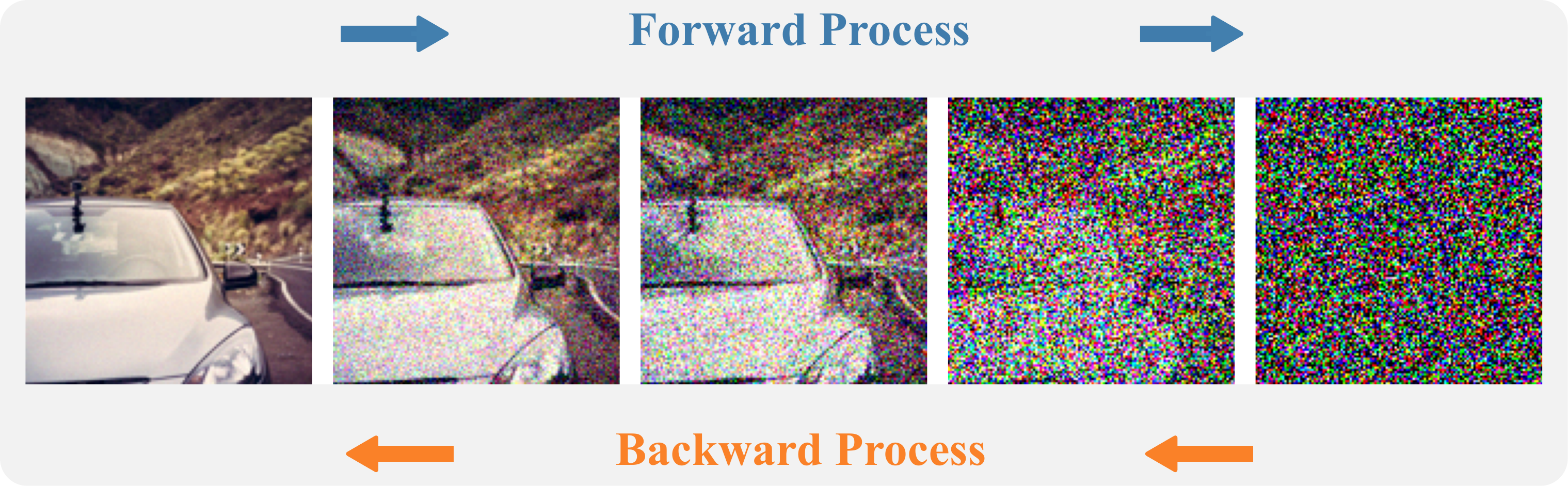}
    \caption{An illustration of how a DM transforms a clean image into noise through the forward process, and then reconstructs it in reverse during the backward process.}
    \label{fig:DM_overview}
\end{figure}

\glspl{dm} are generative models inspired by non-equilibrium thermodynamics~\cite{firstDM}, mirroring natural processes like ink diffusing through water. At their core, they follow the simple yet powerful idea of systematically and gradually destroying structure in data through iterative noise addition and learning to reverse this process in a step-wise fashion. While introduced by Sohl-Dickstein et al.~\cite{firstDM}, the approach gained widespread adoption through Ho et al.'s \gls{ddpm}~\cite{ho2020denoising}. The framework of \glspl{dm}, as illustrated in Figure \ref{fig:DM_overview}, involves two key phases: the forward process and the backward process.
%\vspace{1mm}

\noindent (1) Forward Process:
The forward process refers to the act of gradually corrupting the original data $x_0$ by adding Gaussian noise over $T$ steps, resulting in a sequence of noisy samples $x_1, x_2,\ldots, x_T$. By the final step, $x_T$, the sample is indistinguishable from pure noise.
%\vspace{1mm}

\noindent (2) Backward Process:
To generate realistic samples from pure Gaussian noise, a \gls{dm} must learn to invert its forward corrupting process. This is achieved through an iterative denoising procedure, where the model progressively refines the noisy input to recover the underlying data distribution. At each step, the model estimates and removes the noise added during the forward process, gradually reconstructing the target sample. Denoising is typically parameterized by a neural network, such as a U-Net~\cite{ho2020denoising}, which is trained to predict the noise component at each iteration.

Following the establishment of this paradigm, research advanced primarily along two key directions:

\textit{Controllability:} Unlike the original \gls{ddpm}~\cite{ho2020denoising}, which is trained unconditionally and provides little control over the generated samples, subsequent research has developed methods to guide the diffusion process toward desired outputs. Conditioning the network on auxiliary signals, such as class labels, text embeddings, layout maps, or other modalities, enables structural constraints that guide the generative process. Classifier guidance~\cite{classifierGuidance} uses gradients from a separate classifier to steer sampling towards desired outputs. Classifier-free guidance~\cite{Glide} eliminates the need for a separate classifier by jointly training the model with and without conditioning signals, allowing adjustable control at inference. ControlNet~\cite{ControlNet} further expands controllability by incorporating spatial conditions such as edges, depth, or poses, enabling fine-grained user control.

\textit{Efficiency:} The high computational cost of \gls{ddpm} stems from many iterative steps at full resolution. \rev{\glspl{ldm}} \cite{rombach2022high} address this by operating in compressed latent spaces, reducing complexity while preserving quality. \gls{dit} \cite{peebles2023dit} builds on this by replacing the U-Net with a transformer backbone, improving scalability and global context modeling.

These previous innovations have enabled the use of \glspl{dm} across a wide range of domains. 
\revB{These advances have also been adopted in large-scale commercial systems, such as  Imagen, Stable Diffusion, and Adobe Firefly, which are illustrated in Figure~\ref{fig: LLM_timeline} as part of the \glspl{dm}' development timeline.}
\gls{ad} is a particularly impactful area where \glspl{dm} are used to generate realistic scenarios efficiently and controllably.

\subsection{Scenario Generation}

\rev{This section provides an overview of \glspl{dm} for scenario generation in \gls{ad}, organized by output type: dynamic traffic flow, static traffic elements, images, and videos.}

\textbf{Traffic Flow Generation:} \label{sec:traffic_flow}
Traditional simulators~\cite{Lopez2018-sumo, dosovitskiy2017carla, rong2020lgsvl, beamng2022} typically rely on replaying driving logs or using heuristic-based controllers, which often do not accurately capture the complexity and adaptability of real human behavior. Recent advancements in generative models present an opportunity to create realistic and diverse traffic behavior of virtual agents directly. These models can generate the behavior (trajectories) of multiple agents over time. \rev{To serve as reliable simulation tools}, such models must achieve both realism and controllability, reflecting human-like driving behaviour while adhering to customizable rules. To enhance realism, these models are typically trained on large-scale real-world driving datasets to learn the underlying dynamics and diversity of traffic behavior. 
In the following, we review different techniques to achieve controllability.

(I) \textit{Gradient-Based Guidance} in \glspl{dm} works by modifying the predicted mean at each denoising step using the gradient of a control objective. 
This perturbs the generation toward samples that better fulfill the objective while still following the underlying diffusion process. \rev{Depending on how the objective is defined, such guidance can either enforce safety constraints or, conversely, induce adversarial and safety-critical scenarios.}
CTG~\cite{zhong2023guidedcond} incorporates \gls{stl} to encode traffic rules, using the robustness score of \gls{stl} as a measure of how well the rules are followed and leveraging its gradient to guide trajectory sampling. 
CCDiff~\cite{lin2024CCDiff} leverages the gradient of a constrained \gls{mdp} to guide trajectory generation for multiple agents, with the \gls{mdp} encoding specific control goals such as causing collisions. Before applying guidance, a causal reasoner ranks agents based on inter-agent influence and restricts guidance to the most impactful subset to improve efficiency and effectiveness. 
DiffScene~\cite{xu2023diffscene} defines three differentiable objectives: safety-critical (maximizing collision risk), functional (hindering ego task completion), and constraint-based (enforcing realism rules). 
Lu et al.~\cite{lu2024data} extend DiffScene by encouraging adversarial agents to exhibit aggressive maneuvers (via acceleration/yaw rate variability) and manipulate traffic density around the ego vehicle.  
AdvDiffuser~\cite{xie2024advdiffuser} trains a model to predict how likely a scenario causes failures for a given planner and uses this signal to guide the sampling process. 
SafeSim~\cite{chang2024safesim} and VBD~\cite{huang2024versatile} generate potential trajectories and identify those that would lead to collisions, then use guided diffusion to denoise them. 
A different approach is proposed by Zhong et al.~\cite{zhong2023languagediff} and \rev{LD-Scene}~\cite{peng2025ldscene} , both of which leverage an \gls{llm} to translate natural language instructions (e.g., ``aggressive lane change'') into differentiable guidance functions, bridging high-level intent with low-level control.

\begin{table*}[!htbp]
\center
\caption{Summary of Scenario Generation Studies Using \glspl{dm}.}
\begin{threeparttable}
\resizebox{1\textwidth}{!}{
\renewcommand\arraystretch{1}
\begin{tabular}{p{1.6cm}ccccccp{2cm}p{3.7cm}p{0.9cm}p{1.7cm}l}
\hline
 \multirow{2}{*}{\textbf{\makecell{Category\\(Output)}}} & \multirow{2}{*}{\textbf{\makecell{Safety\\critical\\scenario?}}} & \multicolumn{4}{c}{\textbf{Input}} & \multirow{2}{*}{\textbf{\makecell{Controll\\-ability\tnote{1}}}} & \multirow{2}{*}{\textbf{\makecell{Controllable\\Factor\tnote{2}}}} & \multirow{2}{*}{\textbf{Technique}} & \multirow{2}{*}{\textbf{\makecell{Base\\Model}}}  & \multirow{2}{*}{\textbf{Dataset}} & \multirow{2}{*}{\textbf{Paper}} \\
\cline{3-6}
 &  & \textbf{\makecell{Road\\Topology}} & \textbf{\makecell{Initial\\State}} & \textbf{\makecell{Text\\Prompt}} & \textbf{\makecell{Bounding\\Boxes}} &  &  &  & \\

% trajectory
\toprule
\multirow{19}{*} {\textbf{\makecell{Traffic Flow}}} &  \multirow{10}{*} {No} & \checkmark & \checkmark &  &   &\halfcirc  &\makecell{Speed\\Goal Waypoint} & STL as Guidance &DDPM & nuScenes & CTG \cite{zhong2023guidedcond}   \\

&    &\cellcolor{LightGray}  &\cellcolor{LightGray}   &\cellcolor{LightGray}\checkmark        &\cellcolor{LightGray}   &\cellcolor{LightGray}\fullcirc    &\cellcolor{LightGray} &\cellcolor{LightGray} LLM-Driven Scene Initialization  &\cellcolor{LightGray}DiT  &\cellcolor{LightGray}Argoverse 2 &\cellcolor{LightGray}DriveGen \cite{zhang2025drivegen}\\

&    & \checkmark &             &   &       &\halfcirc    & \makecell{\vspace{1pt}Traffic Density \\ Agents' Position \\ Agents' Speed \\ Agents' Size} & Architecture Conditioning &LDM &Argoverse 2 & Pronovost et al. \cite{pronovost2023scenariodiff} \\

&    & \cellcolor{LightGray}\checkmark & \cellcolor{LightGray}            & \cellcolor{LightGray}   & \cellcolor{LightGray}      &\cellcolor{LightGray}\halfcirc    & \cellcolor{LightGray}Traffic Density &\cellcolor{LightGray}Architecture Conditioning &\cellcolor{LightGray}DiT &\cellcolor{LightGray}{WOMD}  & \cellcolor{LightGray}SceneDiffuser~\cite{jiang2024scenediffuser} \\

&    &  &   &        &   &\emptycirc    &\makecell{} & Map-Free Scene Generation &LDM &WOMD & \makecell{DriveSceneGen \cite{sun2024drivescenegen}}\\

&    & \cellcolor{LightGray} & \cellcolor{LightGray} & \cellcolor{LightGray}   & \cellcolor{LightGray}      &\cellcolor{LightGray}\emptycirc    & \cellcolor{LightGray}\makecell{} &\cellcolor{LightGray}Raster-to-Vector Representation &\cellcolor{LightGray}DiT  & \cellcolor{LightGray}\makecell{nuPlan}  & \cellcolor{LightGray}Sledge~\cite{chitta2024sledge} \\

&    &  &  & \checkmark   &       &\halfcirc    & \makecell{Traffic Density\\Road Layout} &Vectorized Latent Diffusion &LDM   &  \makecell{WOMD\\nuPlan}  & Rowe et al.~\cite{rowe2025scenariodreamer} \\

&    & \cellcolor{LightGray}\checkmark & \cellcolor{LightGray}\checkmark  &\cellcolor{LightGray}        &\cellcolor{LightGray}   &\cellcolor{LightGray}\halfcirc    &\cellcolor{LightGray}\makecell{Speed\\Goal Waypoint} &\cellcolor{LightGray} Preference Optimization &\cellcolor{LightGray}DiT &\cellcolor{LightGray}nuScenes &\cellcolor{LightGray}Yu et al. \cite{yu2025dpo}\\
\addlinespace[-1pt]
  
\cline{2-12}

& \multirow{9}{*}{Yes}  & \checkmark & \checkmark & &   &\halfcirc & \makecell{Collision Type} & MDP as Guidance &DDPM   & \makecell{nuScenes} & CCDiff~\cite{lin2024CCDiff}\\

  &  & \cellcolor{LightGray}\checkmark & \cellcolor{LightGray}\checkmark &\cellcolor{LightGray}   &\cellcolor{LightGray} &\cellcolor{LightGray}\halfcirc & \cellcolor{LightGray}Speed &\cellcolor{LightGray}Gradient-Based Guidance &\cellcolor{LightGray}DDPM  &\cellcolor{LightGray}CARLA & \cellcolor{LightGray}DiffScene \cite{xu2023diffscene} \\

   & &\checkmark &\checkmark  & &  & \halfcirc  & \makecell{Traffic Density\\Speed} &Gradient-Based Guidance &DDPM  & nuScenes &Lu et al.~\cite{lu2024data} \\

  &  &\cellcolor{LightGray}\checkmark &\cellcolor{LightGray}\checkmark &\cellcolor{LightGray}    &\cellcolor{LightGray} &\cellcolor{LightGray}\emptycirc & \cellcolor{LightGray} &\cellcolor{LightGray}Gradient-Based Guidance &\cellcolor{LightGray}LDM   & \cellcolor{LightGray}nuScenes & \cellcolor{LightGray}AdvDiffuser~\cite{xie2024advdiffuser}\\

    &  & \checkmark & \checkmark & &   &\halfcirc & \makecell{Dirving Style\\Collision Type} & Partial Diffusion &DDPM   & \makecell{nuPlan\\nuScenes} & SafeSim~\cite{chang2024safesim}\\

   &  &\cellcolor{LightGray}\checkmark &\cellcolor{LightGray}\checkmark  &\cellcolor{LightGray}  &\cellcolor{LightGray}  & \cellcolor{LightGray}\halfcirc &\cellcolor{LightGray}Driving Style  & \cellcolor{LightGray}Gradient-Based Guidance &\cellcolor{LightGray}DiT  & \cellcolor{LightGray}WOMD  & \cellcolor{LightGray}VBD \cite{huang2024versatile}\\ 
   
  &   & \checkmark & \checkmark & \checkmark &   &\fullcirc & & LLM-Generated Loss Function &DiT  & nuScenes & Zhong et al. ~\cite{zhong2023languagediff}  \\

&    &\cellcolor{LightGray}  &\cellcolor{LightGray}   &\cellcolor{LightGray}\checkmark        &\cellcolor{LightGray}   &\cellcolor{LightGray}\fullcirc    &\cellcolor{LightGray} &\cellcolor{LightGray} LLM-Driven Scene Initialization &\cellcolor{LightGray}LDM  &\cellcolor{LightGray} nuScenes &\cellcolor{LightGray}LD-Scene \cite{peng2025ldscene}\\
\addlinespace[-1pt]

\hline
\multirow{5}{*}{\textbf{\makecell{Static Traffic \\ Element}}} &  \multirow{5}{*} {No}  & &  &\checkmark &   &\halfcirc & \makecell{Number of Lanes\\Type of Road} & Road-UNet architecture &DDPM   &\makecell{OSM} & DiffRoad~\cite{zhou2024diffroad}\\
\addlinespace[-5pt]
&   &\cellcolor{LightGray}\checkmark  &\cellcolor{LightGray}   &\cellcolor{LightGray}        &\cellcolor{LightGray}   &\cellcolor{LightGray}\emptycirc    &\cellcolor{LightGray} &\cellcolor{LightGray}End-to-End Differentiable  &\cellcolor{LightGray}LDM  &\cellcolor{LightGray} In-house &\cellcolor{LightGray}Pronovost et al. \cite{pronovost2023generating}\\

&  &\checkmark  &  & &   &\halfcirc & \makecell{Agents' Position\\Agents' Density\\Agent' Speed\\ Agents' Size} &Guided Agent Placement  &DDPM   &\makecell{Argoverse 2} & SceneControl~\cite{lu2024scenecontrol}\\

% image
\addlinespace[-1pt]
\hline
\multirow{9}{*}{\textbf{\makecell{Driving Image}}} & \multirow{9}{*}{No} &\cellcolor{LightGray}  &\cellcolor{LightGray}  &\cellcolor{LightGray}\checkmark  &\cellcolor{LightGray}   &\cellcolor{LightGray}\halfcirc  &\cellcolor{LightGray}\makecell{Road Topology\\Traffic Density\\Weather} &\cellcolor{LightGray}Structured Prompt &\cellcolor{LightGray}\makecell{LDM\\DDPM}  &\cellcolor{LightGray}nuScenes &\cellcolor{LightGray}Text2Street \cite{gu2025text2street}   \\

&  &  &  &  &\checkmark   &\halfcirc  &Camera Pose & Bounding Box Translation &LDM  &nuSences & GeoDiffusion \cite{chen2023geodiffusion}  \\
 
 &   & \multicolumn{4}{>{\columncolor{LightGray}}c}{BEV Sketch}  &\cellcolor{LightGray}\halfcirc  &\cellcolor{LightGray}\makecell{Weather\\Lighting Condition} &\cellcolor{LightGray}Controller \& Coordinator &\cellcolor{LightGray}LDM  &\cellcolor{LightGray}nuScenes  &\cellcolor{LightGray}BEVControl \cite{yang2023bevcontrol}  \\
\addlinespace[-3pt]
  &    & \checkmark &             &    & \checkmark      &\halfcirc    & \makecell{Camera Pose\\Weather\\Lighting Condition}   &Cross-View Attention &LDM   &nuScenes & MagicDrive \cite{gao2023magicdrive}\\
\addlinespace[-5pt]
 &  &\cellcolor{LightGray}\checkmark  &\cellcolor{LightGray}  &\cellcolor{LightGray}\checkmark  &\cellcolor{LightGray}\checkmark   &\cellcolor{LightGray}\halfcirc  &\cellcolor{LightGray}\makecell{Weather\\Lighting Condition\\Camera Pose} &\cellcolor{LightGray}Dual-Branch Diffusion &\cellcolor{LightGray}LDM  &\cellcolor{LightGray}nuScenes &\cellcolor{LightGray}DualDiff \cite{li2025dualdiff-}  \\
\addlinespace[-1pt]
% video
\hline
 \multirow{14}{*} {\textbf{\makecell{Driving Video}}} & \multirow{13}{*} {No}    & \multicolumn{4}{c}{BEV Sequence}  &\halfcirc  &\makecell{Weather\\Lighting Condition\\Landscape} & 4D Attention &LDM  &nuScenes  & Panacea \cite{wen2024panacea}  \\
 \addlinespace[-5pt]
&   &\multicolumn{4}{c}{\cellcolor{LightGray}3D Layout Sequence}  &\cellcolor{LightGray}\halfcirc  &\cellcolor{LightGray}\makecell{Weather\\Lighting Condition} &\cellcolor{LightGray}Cascaded Video Synthesis &\cellcolor{LightGray}LDM &\cellcolor{LightGray}nuScenes  &\cellcolor{LightGray}DrivingDiffusion \cite{li2023drivingdiffusion}  \\

& &\checkmark  &  &\checkmark  &\checkmark   &\fullcirc  &  &Multi-Control Distillation &DiT  &nuScenes &DiVE \cite{jiang2025dive}  \\

&   & \multicolumn{4}{c}{\cellcolor{LightGray}\makebox[0.1\linewidth][c]{\makecell[c]{Canny Edge Map \\ Depth Map \\ Text Prompt}}}  &\cellcolor{LightGray}\halfcirc  &\cellcolor{LightGray}\makecell{Weather\\Lighting Condition} &\cellcolor{LightGray}Dual-Branch Diffusion &\cellcolor{LightGray}LDM   &\cellcolor{LightGray}\makecell{DriveScene\\-DDM \cite{bai2025dctdm}} &\cellcolor{LightGray}DcTDM \cite{bai2025dctdm}  \\

  & &  \multicolumn{4}{c}{Initial Frames} &\emptycirc  & &Frame Sampling Scheme &DDPM  &WOMD  & DriveGenVLM \cite{fu2024drivegenvlm}  \\

& &\cellcolor{LightGray}\checkmark  &\cellcolor{LightGray}  &\cellcolor{LightGray}\checkmark  &\cellcolor{LightGray}\checkmark   &\cellcolor{LightGray}\halfcirc  &\cellcolor{LightGray}\makecell{Weather\\Lighting Condition}  &\cellcolor{LightGray}Dual-Branch Diffusion &\cellcolor{LightGray}LDM &\cellcolor{LightGray}\makecell{nuScenes\\Waymo Open} &\cellcolor{LightGray}DualDiff+ \cite{DualDiff_2024}  \\

 & &  &  &\checkmark  &   &\halfcirc  &\makecell{Weather\\Traffic Density\\Landscape}  &Adapting Existing Methods &LDM  &KITTI  &GenDDS \cite{fu2024gendds}  \\
\addlinespace[-1pt]
\cline{2-12}
\noalign{\vskip 1pt}

& \multirow{2}{*} {Yes}   &\cellcolor{LightGray}  &\cellcolor{LightGray}  &\cellcolor{LightGray}\checkmark   &\cellcolor{LightGray}  &\cellcolor{LightGray}\fullcirc  &\cellcolor{LightGray} &\cellcolor{LightGray}Temporal Shift Adapter &\cellcolor{LightGray}LDM  &\cellcolor{LightGray}DoTA \cite{yao2022dota}  &\cellcolor{LightGray}DrivingGen \cite{guo2024drivinggen}  \\

  & &   &   &\checkmark   & &\fullcirc  &  &Adapting Existing Methods &DiT  &MM-AU \cite{fang2024mmau}  &AVD2 \cite{li2025avd2} \\
\addlinespace[-2pt]

\hline
% \bottomrule
\end{tabular}}
\begin{tablenotes}
\scriptsize
\item[1] Controllability: \fullcirc~Full control (users can fully customize scenes); \halfcirc~Partial control (supports specific parameter adjustments); \emptycirc~No control.
\item[2] \begin{minipage}[t]{0.75\linewidth} Only models with partial controllability are discussed here in this column. Fully controllable models can follow any input (typically via LLMs), and models without control fall outside the scope of this discussion.
\end{minipage}\\
\end{tablenotes}
\end{threeparttable}
\label{dm_sum}
\end{table*}

(II) \textit{Architecture Conditioning} embeds the control signal directly within the network’s structure so that constraints are enforced throughout each iteration, rather than being injected afterwards as an external correction. \gls{dm} achieve this by accepting extra conditioning inputs, such as tokens that carry agent attributes, scene statistics, language descriptions, or spatial masks. These additional inputs are processed by dedicated layers, for example, cross-attention blocks or inpainting modules, and are fused with the latent scene representation at each denoising iteration.
Pronovost et al.~\cite{pronovost2023scenariodiff} encode agent attributes (speed, heading) and global scene properties (agent density) as tokens processed by cross-attention layers. 
SceneDiffuser~\cite{jiang2024scenediffuser} frames trajectory generation as an inpainting task on a 3D tensor of shape $A \times T \times D$, each representing agents, timesteps, and features. Scene editing and agent injection are made possible by adjusting the scene tensor and the associated inpainting mask. 
DriveGen~\cite{zhang2025drivegen} uses a natural language description to generate road layouts and place vehicles via an \gls{llm}. A \gls{vlm} is applied afterwards to analyze the \gls{bev} to identify potential future goals. Finally, a \gls{dm} generates realistic trajectories from each vehicle's initial state to its predicted goal.
DriveSceneGen~\cite{sun2024drivescenegen} addresses two key problems: scene initialization and rollout. It first synthesizes a \gls{bev} image of road layouts and agent positions using a \gls{dm}, then vectorizes the output for trajectory prediction with a \gls{mtr}. SLEDGE~\cite{chitta2024sledge} and ScenarioDreamer~\cite{rowe2025scenariodreamer} address the same task but optimize the generation pipeline. Specifically, SLEDGE introduces a raster-to-vector autoencoder to compress scenes into latent maps for further diffusion, whereas ScenarioDreamer further advances this by operating the DM directly in vector space. Together, these methods reflect a progression from pixel-level (DriveSceneGen) to compressed-raster (SLEDGE) to fully vectorized (ScenarioDreamer) generation.

(III) \rev{\textit{\gls{po}} moves away from gradient-guidance and architecture-conditioning. Instead of explicit control signals or hand-crafted loss functions, Yu et al.~\cite{yu2025dpo} fine-tune the \gls{dm} directly using \gls{po}. The model generates two candidate trajectories per scene, scores them via rule-based heuristics, and updates itself to favor the better one, thereby learning control preferences implicitly.}

Despite recent advances, diffusion-based traffic flow generators still rely on manually crafted control inputs. Gradient-guided models require carefully tuned objective weights, while architecture-conditioned models depend on predefined token or mask schemas to encode rules. Adapting these approaches to new constraints often requires costly retraining or extensive fine-tuning.

\textbf{Static Traffic Element:} \label{sec:traffic_element}
\rev{\glspl{dm}} have also been developed to generate various \textbf{\gls{ad}} components beyond agents' trajectories.

\rev{DiffRoad~\cite{zhou2024diffroad} synthesizes 3D road layouts from structured text inputs (e.g., “two three-way intersections”) and evaluates the outputs based on criteria such as smoothness and the presence of overlapping segments.}

Pronovost et al.~\cite{pronovost2023generating} and SceneControl~\cite{lu2024scenecontrol} focus on generating \rev{initial agent placements} for downstream traffic simulation. Pronovost et al. introduce a scene autoencoder that compresses rasterized agent layouts into latent embeddings. A \gls{dm}, conditioned on a road map, is then trained over these embeddings, and a decoder reconstructs oriented bounding boxes for the agents. SceneControl offers additional flexibility through guided sampling, allowing fine-grained user control (e.g., enforcing speed constraints) and realism guarantees (e.g., collision avoidance and lane adherence) during the generation process. To assess how well the generated scenes match real-world data, both methods compare statistical distributions between real and synthetic datasets. 

These static-scene generators still have notable gaps. When a \gls{dm} is used to synthesize road layouts, fine-grained elements such as traffic signs, signals and lane markings are often omitted. As a result, the resulting maps lack the fidelity needed for high-realism driving simulation. Moreover, initial-scene generators are also highly map-specific: they absorb the spatial priors of the training corpus and can place agents unrealistically when applied to unseen road geometries or regions with different driving conventions.

\textbf{Image Generation:} \label{sec:image_generation}
% Reliable perception in AD relies heavily on large-scale camera datasets with rich annotations. Rather than laboriously collecting and annotating street-view images from the natural environment, \glspl{dm} offer a promising solution by generating realistic images for synthesizing street-view data. 
\rev{Reliable \gls{ad} perception depends on large annotated datasets. \glspl{dm} offer an efficient alternative by generating realistic street-view images.}

Text2Street~\cite{gu2025text2street} decomposes structured prompts, such as ``a street view image with a crossing, 4 lanes, 3 cars, 2 persons, and 2 trucks on a sunny day'', into three distinct components: road topology, object layout, and weather condition. Each of these components is handled by a dedicated \gls{dm}. The first model processes the road topology to generate a \gls{bev} road layout. The second model takes this \gls{bev} layout and incorporates the object layout, producing a map that includes vehicles, pedestrians, and other foreground elements. The third model transforms this \gls{bev} representation into a realistic camera-view street scene. 
To handle geometric conditions more effectively, GeoDiffusion~\cite{chen2023geodiffusion} converts bounding boxes into textual prompts that guide a pre-trained text-to-image \gls{dm}. This involves translating continuous bounding box locations into discrete tokens and balancing the visual prominence of foreground objects with the often-dominant background regions during image generation. Baresi et al. \cite{2025-Baresi-ICSE} generate rare \gls{ood} driving scenarios (e.g., snow, desert) using three diffusion-based strategies: instruction editing, inpainting, and inpainting with refinement.
Meanwhile, other works have focused on generating multi-view images. BEVControl~\cite{yang2023bevcontrol} addresses the complexity of editing dense segmentation maps by using editable \gls{bev} sketches as input. It introduces a ``controller and coordinator'' mechanism to ensure that generated objects match the sketch accurately and maintain consistency across multiple viewpoints. 
MagicDrive~\cite{gao2023magicdrive} considers road layouts, bounding boxes, camera poses, and textual descriptions such as weather and time of day as input. It introduces a cross-view attention module that allows each camera view to access information from its immediate neighbors, ensuring visual consistency and coherence across all generated views.
\rev{DualDiff~\cite{li2025dualdiff-} adopts a dual-branch architecture that separately generates foreground and background. It projects 3D occupancy data onto camera planes to form dense feature maps, fuses them with 3D bounding boxes and road maps, and then combines the branch outputs to synthesize the final image.}

In spite of recent progress, fine details such as traffic signs, pole-mounted signals and lane markings are frequently simplified or omitted, resulting in generated images that fail to cover many visual corner cases that real perception stacks must handle. Photometric realism is also limited: simplified lighting models and the absence of camera artifacts such as rolling-shutter distortion, lens flare, and sensor noise create a noticeable domain gap when these synthetic frames are used to train or evaluate real-world detectors.

\textbf{Video Generation:} \label{sec:video_generation}
\rev{Recent work has also advanced \gls{dm}-based driving video generation, improving temporal consistency, controllability, and diversity.}

Several studies have introduced innovative architectures to ensure multi-view and temporal consistency in generated videos. 
Panacea~\cite{wen2024panacea} generates multi-view video sequences by first synthesizing images from \gls{bev} inputs and then expanding them along the temporal dimension. The method introduces a 4D attention mechanism that takes into account intra-view (within each camera), cross-view (between adjacent cameras) and cross-frame (between temporal patches). DrivingDiffusion~\cite{li2023drivingdiffusion} also employs a multi-stage approach: it first generates a consistent initial frame across all camera views from a layout, then uses a temporal model to produce short view-specific sequences, and finally refines long-term consistency via a sliding-window post-processing module.
DiVE~\cite{jiang2025dive} focuses specifically on efficient multi-view driving scene generation. It introduces Multi-Control Auxiliary Branch Distillation (MAD) to streamline multi-condition classifier-free guidance, significantly reducing inference time. DiVE also proposes view-inflated attention, a lightweight mechanism enforcing cross-view consistency without adding parameters.

Another strategy for video generation is adapting image \glspl{dm} with temporal expansion. \rev{DrivingGen~\cite{guo2024drivinggen} extends a text-to-image \gls{dm} by incorporating a temporal shift adapter that efficiently propagates information across frames using modified 2D convolutions instead of costly 3D operations.
Similarly, DcTDM~\cite{bai2025dctdm} extends image-based diffusion into the temporal domain but introduces dual conditioning with dense depth maps and Canny edge maps to preserve geometric and structural consistency across frames. 
DriveGenVLM~\cite{fu2024drivegenvlm} enhances long-term video generation through conditioning and sampling strategies, such as frame-by-frame generation and keyframe interpolation, offering trade-offs between quality and speed.}

In contrast, DualDiff+~\cite{DualDiff_2024} generates videos through a dual-branch architecture that decouples foreground and background modeling. The model first projects a 3D occupancy grid into 2D space and then fuses these features with semantic inputs, including 3D bounding boxes (foreground) and maps (background). 

\rev{Another line of research advances video generation by combining and adapting existing models.}  GenDDS~\cite{fu2024gendds} fine-tunes Stable Diffusion XL~\cite{podell2024sdxl} using \gls{lora}~\cite{hu2022lora} to produce driving images, which are then extended into videos through a temporal transformer in Hotshot-XL~\cite{hotshotxl2023}. AVD2~\cite{li2025avd2} fine-tunes the Open-Sora~1.2 model~\cite{brooks2024opensora} on the MM-AU~\cite{fang2024mmau} dataset to generate videos annotated with accident causes and avoidance strategies. 

Despite recent advances, diffusion-based generators for driving videos still face significant challenges. They often struggle to maintain consistent temporal and multi-view coherence, particularly over extended clips. Additionally, their understanding of the physical world's dynamics remains limited: for example, vehicles may behave in ways that defy inertia or violate occlusion logic.

\subsection{Limitations and Future Directions}

Although recent \glspl{dm} support conditioning through layout masks, language tokens, or attention-based inputs, these mechanisms often remain rigid and narrowly specialized. They typically depend on manual tuning, predefined conditioning schemas, or task-specific re-training, which limits their flexibility and scalability. To address this, future research should aim to develop more generalizable conditioning frameworks that can seamlessly integrate diverse or novel inputs without requiring substantial architectural modifications or re-training.

In parallel, while \glspl{dm} often achieve strong performance on statistical realism metrics, the generated trajectories and scenes frequently lack fine-grained physical plausibility. Artifacts such as implausible inertial dynamics, unnatural agent reactions, and inadequate modeling of occlusions or causal dependencies are common. One promising direction for future research is the integration of physics-informed models, which could improve the adherence to real-world physical laws and enhance the overall realism of the generated outputs.

Moreover, although \glspl{llm} are increasingly used to convert natural language inputs into guidance signals for \glspl{dm}, their potential remains underutilized. Rather than serving solely as input translators, \glspl{llm} could act as embedded knowledge sources that encode rich priors about physical dynamics, semantic scene structure, and normative driving behavior. Leveraging these capabilities may substantially improve the controllability, realism, and interpretability of diffusion-generated scenarios, particularly in complex or ambiguous environments.

\hypertarget{sec:wm}{}
\section{World Models (WMs)} \label{sec:wm}
%\vspace{1mm}
%\begin{quote}
%    ``Our results suggest that scaling video generation models is a promising path
%    towards building general purpose \textbf{simulators} of the physical \textbf{world}.''
%    \begin{flushright}
%        -- \emph{\textbf{OpenAI}'s Sora Technical Report}~\cite{brooks2024opensora}
%    \end{flushright}
%\end{quote}
%%\vspace{1mm}

%%\vspace{1mm}
%\begin{quote}
%    ``World models are \textbf{generative} AI models that understand the \textbf{dynamics} of the real world, including physics and spatial properties.''
%    \begin{flushright}
%        -- \emph{\textbf{NVIDIA}'s World Foundation Models}~\cite{nvidiaWorldModels}
%    \end{flushright}
%\end{quote}
%%\vspace{1mm}

\glspl{wm} are generative neural network models that learn compressed spatial and temporal representations of an environment~\cite{ha2018world}. They enable agents to develop an internal model of the world to make predictions about future states of the surrounding world environment, concerning both dynamic agents and static objects. In this section, we focus on their ability to generate driving scenarios, and we categorize recent works into visual, 3D occupancy, and multi-modal generation. Moreover, we discuss related architectural innovations and benchmarks.

\subsection{Development of World Models}

\begin{figure}[h]
    \centering
    \includegraphics[width=0.8\columnwidth]{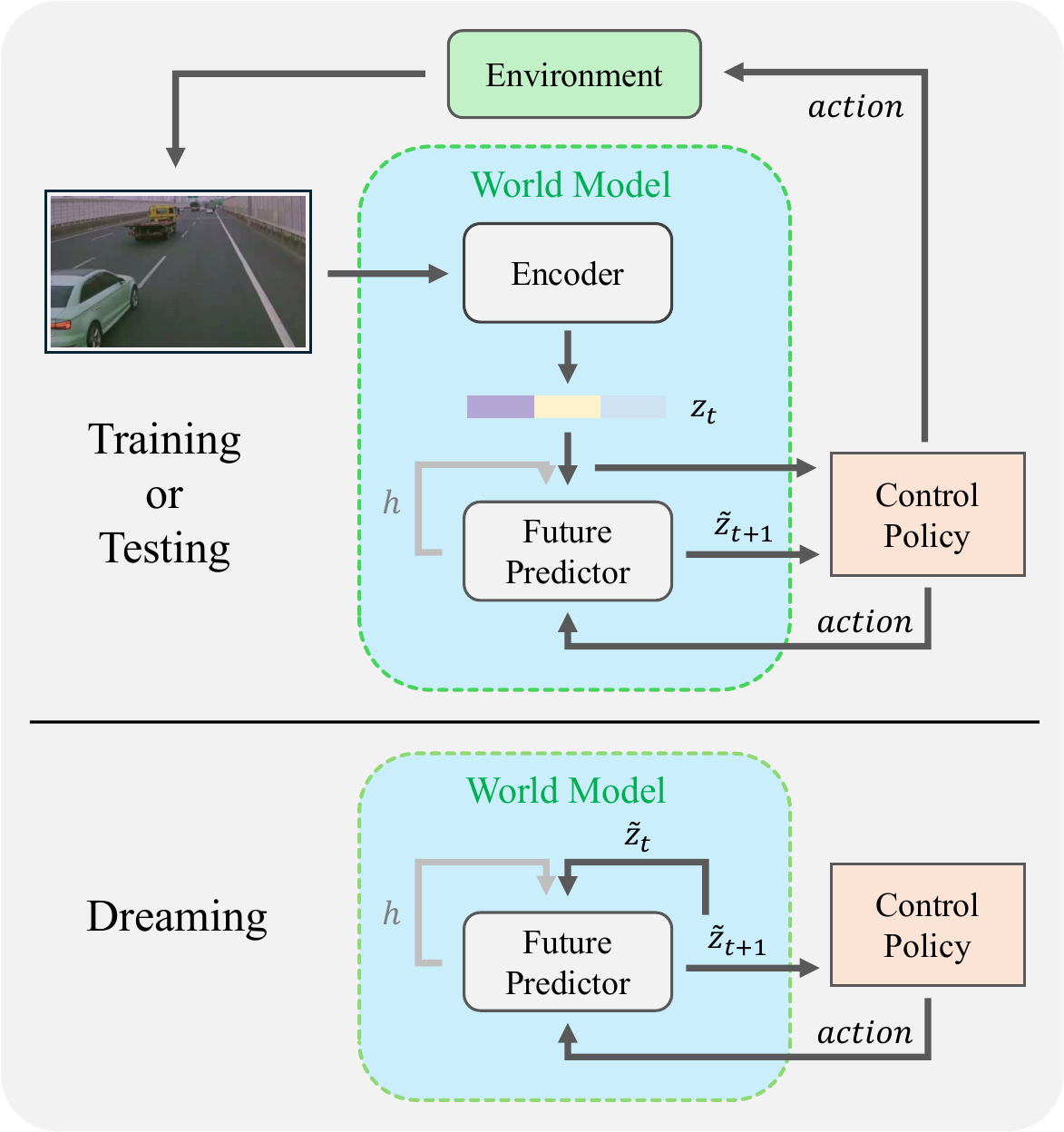}
    \caption{Overview of world model's training, testing and dreaming phases.
    In the training/testing phase (top), $z_{t}$
    is a latent representation of the input (e.g., image), $\tilde{z}_{t+1}$ is the
    prediction of the latent representation at the next time step, and $h$ is a hidden state encoding past information.
    In the dreaming phase (bottom), the model generates future latent variables $\tilde{z}_{t}$ in an auto-regressive way: $\tilde{z}_{t}$ is initialized with a $\tilde{z}_{0}$, and then recursively fed back as input of the future predictor, which computes the next value.}
    \label{fig:wm_overview}
    \vspace{-1mm}
\end{figure}

\textbf{Relations with Cognitive Science:}
The development of \glspl{wm} focuses on learning compact, predictive representations of the physical world's dynamics.
This concept draws inspiration from the human brain's ability to model and predict the physics of the real world \cite{lecun2022path}. Cognitive science has proposed predictive brain models to anticipate the evolution of real-world scenarios, such as the procedural and declarative models of Downing (2009) \cite{Downing2009} . 
Svensson et al. (2013) \cite{Svensson2013} apply brain-like “dreaming” to simulate perception-action sequences offline, for simple robotic systems. They define mental imagery as the brain's ability to generate and manipulate internal representations of the world, in a dreaming-like process without direct interaction with the environment. Similarly, as explained in the next section, \glspl{wm} can dream by generating imagined scenarios (Figure~\ref{fig:wm_overview}), without interacting with the physical world and using their learned representations to predict the future evolution of the environment.

In a related context, Plebe et al. \cite{Plebe2019imagery,Plebe2020} propose vision autoencoders that emulate neural convergence-divergence patterns of the brain \cite{Meyer2009}, to output long-term predictions of driving scenarios in the form of videos.
They suggest minimizing the free energy as a training loss function, which is inspired by the Friston's theories \cite{Friston2010} about the brain's operation. Similar minimum-free-energy principles are also proposed for the training of \glspl{wm} by LeCun \cite{lecun2022path}. For a broader review of bio-inspired cognitive agents in \gls{ad}, the reader can refer to \cite{Plebe_survey_2024}.

We argue that such cognitive theories will inspire the next generation of \glspl{wm}, which will need to learn generalizable models of the real-world dynamics from limited data. \revB{The timeline of \glspl{wm}' development is shown in Figure~\ref{fig: LLM_timeline}.}

%Moreover, our brain was shown to use convergence-divergence structures for perception \cite{Plebe2020} and mental imagery, and these structures are conceptually similar to the encoder-decoder neural architectures used in \glspl{wm}. 
%

\begin{table*}
    [htp] \center
    \caption{Comparison of key research about World Models for Scenario
    Generation in Autonomous Driving.}
    \begin{threeparttable}
        \resizebox{1.0\textwidth}{!}{
        \begin{tabular}{clccccccccllcl}
            \toprule \multirow{4}{*}{\textbf{Category}}                                                   & \multicolumn{6}{c}{\textbf{Input}} & \multirow{4}{*}{\textbf{\rotatebox{90}{Controllability}}\tnote{2}} & \multirow{4}{*}{\textbf{\makecell{Multi-view\\Generation}}} & \multirow{4}{*}{\textbf{\makecell{World Model\\Architecture}}} & \multirow{4}{*}{\textbf{\makecell{Model Types}}} & \multirow{4}{*}{\textbf{Dataset}}          & \multirow{4}{*}{\textbf{\makecell{Code}}\tnote{3}} & \multirow{4}{*}{\textbf{Paper}}  \\
            \cmidrule(lr){2-7}                                                                             %& & Image & Text & Action & trajectory & Geometry & Map &            &         \\
                                                                                                          & \rotatebox{45}{\textbf{Image}}     & \rotatebox{45}{\textbf{Text}}                                      & \rotatebox{45}{\textbf{Action}}                             & \rotatebox{45}{\textbf{Trajectory}}                            & \rotatebox{45}{\textbf{Geometry}}\tnote{1} & \rotatebox{45}{\textbf{Map}}                       &                                 &                              \\
            \midrule \multirow{18}{*}{\textbf{\shortstack{Visual\\Generation}}}         & \cmark                             & \cmark                                                             & \cmark                                                      &                                                                &                                            &                                                    & \halfcirc                       &                             & Autoregression                      & Transformer~\cite{vaswani2017attention}                      & In-house                                                                &                            & GAIA-1 \cite{hu2023gaia}                                       \\
                                                                                                          & \cellcolor{LightGray}\cmark        & \cellcolor{LightGray}\cmark                                        & \cellcolor{LightGray}\cmark                                 & \cellcolor{LightGray}                                          & \cellcolor{LightGray}\cmark                & \cellcolor{LightGray}\cmark                        & \cellcolor{LightGray}\fullcirc  & \cellcolor{LightGray}\cmark & \cellcolor{LightGray}Diffusion      & \cellcolor{LightGray}LDM~\cite{rombach2022high} & \cellcolor{LightGray}nuScenes \cite{caesar2020nuscenes}                 & \cellcolor{LightGray}\cmark & \cellcolor{LightGray}DriveDreamer \cite{wang2024drivedreamer}  \\
                                                                                                          & \cmark                             & \cmark                                                             &                                                             &                                                                &                                            &                                                    & \emptycirc                      &                             & Diffusion                           & LDM                      & nuScenes, In-house                                                      &                            & ADriver-I \cite{ADriver-I_2023}                                \\
                                                                                                          & \cellcolor{LightGray}\cmark        & \cellcolor{LightGray}\cmark                                        & \cellcolor{LightGray}\cmark                                 & \cellcolor{LightGray}\cmark                                    & \cellcolor{LightGray}\cmark                & \cellcolor{LightGray}                              & \cellcolor{LightGray}\fullcirc  & \cellcolor{LightGray}\cmark & \cellcolor{LightGray}Diffusion      & \cellcolor{LightGray}LDM & \cellcolor{LightGray}In-house                                           & \cellcolor{LightGray}      & \cellcolor{LightGray}GAIA-2 \cite{russell2025gaia}             \\
                                                                                                          & \cmark                             & \cmark                                                             & \cmark                                                      & \cmark                                                         & \cmark                                     & \cmark                                             & \fullcirc                       & \cmark                      & Diffusion                           & SVD~\cite{blattmann2023stable}                      & nuScenes                                                                & \cmark                      & DriveDreamer-2 \cite{DriveDreamer2_2024}                       \\
                                                                                                          & \cellcolor{LightGray}\cmark        & \cellcolor{LightGray}\cmark                                        & \cellcolor{LightGray}                                       & \cellcolor{LightGray}                                          & \cellcolor{LightGray}\cmark                & \cellcolor{LightGray}\cmark                        & \cellcolor{LightGray}\halfcirc  & \cellcolor{LightGray}       & \cellcolor{LightGray}Diffusion      & \cellcolor{LightGray}LDM & \cellcolor{LightGray}Waymo Open dataset \cite{sun2020waymoopen}         & \cellcolor{LightGray}\cmark & \cellcolor{LightGray}DriveDreamer4D \cite{DriveDreamer4D_2024} \\
                                                                                                          & \cmark                             & \cmark                                                             & \cmark                                                      & \cmark                                                         &                                            &                                                    & \fullcirc                       &                             & Diffusion                           & SVD                      & nuScenes, etc \cite{sun2020waymoopen, yang2024generalized, li2022coda}. & \cmark                      & Vista \cite{gao2024vista}                                      \\ % nuScenes, WOD, CODA, OpenDV
                                                                                                          & \cellcolor{LightGray}\cmark        & \cellcolor{LightGray}                                              & \cellcolor{LightGray}                                       & \cellcolor{LightGray}\cmark                                    & \cellcolor{LightGray}                      & \cellcolor{LightGray}                              & \cellcolor{LightGray}\emptycirc & \cellcolor{LightGray}       & \cellcolor{LightGray}Autoregression & \cellcolor{LightGray}Transformer & \cellcolor{LightGray}nuPlan \cite{karnchanachari2024nuplan}, In-house   & \cellcolor{LightGray}\cmark & \cellcolor{LightGray}DrivingWorld \cite{hu2024drivingworld} \\
                                                                                                          & \cmark                             & \cmark                                                             & \cmark                                                      &                                                                & \cmark                            & \cmark                                                           & \fullcirc                       & \cmark                      & Diffusion                         & VideoLDM~\cite{blattmann2023align}                      & nuScenes                                                                & \xmark                      & Drive-WM \cite{Drive-WM2024}                                   \\
                                                                                                          & \cellcolor{LightGray}\cmark        & \cellcolor{LightGray}\cmark                                        & \cellcolor{LightGray}                                       & \cellcolor{LightGray}                                          & \cellcolor{LightGray}\cmark                & \cellcolor{LightGray}\cmark                        & \cellcolor{LightGray}\halfcirc  & \cellcolor{LightGray}\cmark & \cellcolor{LightGray}Diffusion      & \cellcolor{LightGray}LDM & \cellcolor{LightGray}nuScenes                                           & \cellcolor{LightGray}\cmark & \cellcolor{LightGray}MagicDrive \cite{gao2023magicdrive}       \\
                                                                                                          &                                    & \cmark                                                             &                                                             & \cmark                                                         & \cmark                                     & \cmark                                             & \halfcirc                       & \cmark                      & Diffusion                           & LDM                      & nuScenes                                                                &                      & MagicDrive3D \cite{MagicDrive3D_2024}                          \\
                                                                                                          & \cellcolor{LightGray}\cmark        & \cellcolor{LightGray}\cmark                                        & \cellcolor{LightGray}                                       & \cellcolor{LightGray}\cmark                                    & \cellcolor{LightGray}\cmark                & \cellcolor{LightGray}\cmark                        & \cellcolor{LightGray}\fullcirc  & \cellcolor{LightGray}\cmark & \cellcolor{LightGray}Diffusion      & \cellcolor{LightGray}DiT~\cite{peebles2023dit} & \cellcolor{LightGray}nuScenes                                           & \cellcolor{LightGray} & \cellcolor{LightGray}MagicDrive-V2 \cite{gao2024magicdrivedit} \\
                                                                                                          &                              & \cmark                                                             &                                                             &                                                                & \cmark                                     & \cmark                                             & \halfcirc                       & \cmark                      & Diffusion                           & LDM                      & nuScenes, Occ3d \cite{tian2023occ3d}                                    & \cmark                      & WoVoGen \cite{lu2024wovogen}                                   \\
                                                                                                          & \cellcolor{LightGray}\cmark        & \cellcolor{LightGray}                                              & \cellcolor{LightGray}                                       & \cellcolor{LightGray}\cmark                                    & \cellcolor{LightGray}\cmark                & \cellcolor{LightGray}\cmark                        & \cellcolor{LightGray}\halfcirc  & \cellcolor{LightGray}       & \cellcolor{LightGray}Diffusion      & \cellcolor{LightGray}SVD & \cellcolor{LightGray}Waymo Open dataset                                 & \cellcolor{LightGray}\cmark & \cellcolor{LightGray}ReconDreamer \cite{ni2024recondreamer}    \\
                                                                                                          % & \cmark                             & \cmark                                                             &                                                             &                                                                & \cmark                                     & \cmark                                             & \halfcirc                       & \cmark                      & Autoregression                      & LDM                      & nuScenes                                                                & \cmark                      & DualDiff+ \cite{DualDiff_2024}                                 \\
                                                                                                          % & \cellcolor{LightGray}\cmark        & \cellcolor{LightGray}\cmark                                        & \cellcolor{LightGray}                                       & \cellcolor{LightGray}                                          & \cellcolor{LightGray}\cmark                & \cellcolor{LightGray}\cmark                        & \cellcolor{LightGray}\halfcirc  & \cellcolor{LightGray}\cmark & \cellcolor{LightGray}Diffusion      & \cellcolor{LightGray}LDM & \cellcolor{LightGray}nuScenes                                           & \cellcolor{LightGray}      & \cellcolor{LightGray}Panacea \cite{wen2024panacea}             \\
                                                                                                          & \cmark                             & \cmark                                                             &                                                             &                                                                & \cmark                                     &                                                    & \halfcirc                       &                             & Diffusion                           & DiT~\cite{peebles2023dit}                      & Cosmos \cite{agarwal2025cosmos}                                         &                            & Cosmos-Transfer1 \cite{Cosmos-Transfer1_2024}                  \\
                                                                                                          & \cellcolor{LightGray}\cmark        & \cellcolor{LightGray}                                              & \cellcolor{LightGray}                                       & \cellcolor{LightGray}\cmark                                    & \cellcolor{LightGray}\cmark                & \cellcolor{LightGray}                              & \cellcolor{LightGray}\halfcirc  & \cellcolor{LightGray}       & \cellcolor{LightGray}Diffusion      & \cellcolor{LightGray}VideoLDM~\cite{blattmann2023align} & \cellcolor{LightGray}nuScenes                                           & \cellcolor{LightGray}      & \cellcolor{LightGray}GeoDrive \cite{chen2025geodrive}          \\
            \midrule \multirow{6}{*}{\textbf{\shortstack{3D\\Occupancy\\Generation}}} &                                    &                                                                    &                                                             & \cmark                                                         &                                            &                                                    & \emptycirc                      &                            & Diffusion                           & DiT~\cite{peebles2023dit}                      & nuScenes                                                                & \cmark                      & OccSora \cite{wang2024occsora}                                 \\
                                                                                                          & \cellcolor{LightGray}\cmark        & \cellcolor{LightGray}\cmark                                        & \cellcolor{LightGray}\cmark                                 & \cellcolor{LightGray}\cmark                                    & \cellcolor{LightGray}                      & \cellcolor{LightGray}                              & \cellcolor{LightGray}\fullcirc  & \cellcolor{LightGray}      & \cellcolor{LightGray}Autoregression & \cellcolor{LightGray}Transformer & \cellcolor{LightGray}nuScenes, Lyft-Level5 \cite{houston2021one}        & \cellcolor{LightGray}\cmark & \cellcolor{LightGray}Drive-OccWorld \cite{yang2025driving}     \\
                                                                                                          &                                    &                                                                    &                                                             & \cmark                                                         & \cmark                                     &                                                    & \emptycirc                      &                            & Diffusion                           & Latent DiT~\cite{ma2024latte}                       & nuScenes                                                                & \cmark                      & DOME \cite{gu2024dome}                                         \\
                                                                                                          & \cellcolor{LightGray}              & \cellcolor{LightGray}                                              & \cellcolor{LightGray}                                       & \cellcolor{LightGray}                                          & \cellcolor{LightGray}\cmark                & \cellcolor{LightGray}                              & \cellcolor{LightGray}\emptycirc & \cellcolor{LightGray}      & \cellcolor{LightGray}Autoregression & \cellcolor{LightGray}Transformer & \cellcolor{LightGray}nuScenes                                           & \cellcolor{LightGray}      & \cellcolor{LightGray}RenderWorld \cite{yan2024renderworld}     \\
                                                                                                          &                                    & \cmark                                                             &                                                             & \cmark                                                         & \cmark                                     &                                                    & \halfcirc                       &                            & Autoregression                      & Transformer                      & nuScenes, etc \cite{tian2023occ3d, qian2024nuscenes}.                   &                            & OccLLama \cite{wei2024occllama}                                \\ %NuScenes, Occ3D, NuScenes-QA
                                                                                                          & \cellcolor{LightGray}\cmark        & \cellcolor{LightGray}\cmark                                        & \cellcolor{LightGray}\cmark                                 & \cellcolor{LightGray}                                          & \cellcolor{LightGray}                      & \cellcolor{LightGray}                              & \cellcolor{LightGray}\halfcirc  & \cellcolor{LightGray}      & \cellcolor{LightGray}Autoregression & \cellcolor{LightGray}Transformer & \cellcolor{LightGray}nuScenes, Openscene \cite{openscene2023}           & \cellcolor{LightGray}      & \cellcolor{LightGray}DriveWorld \cite{min2024driveworld}       \\
            \midrule \multirow{3}{*}{\textbf{\shortstack{Multi-modal\\Generation}}}     & \cmark                             & \cmark                                                             &                                                             &                                                                & \cmark                                     & \cmark                                             & \halfcirc                       & \cmark                      & Autoregression                      & Transformer                      & nuScenes                                                                &                            & HoloDrive \cite{wu2024holodrive}                               \\
                                                                                                          & \cellcolor{LightGray}\cmark        & \cellcolor{LightGray}                                              & \cellcolor{LightGray}\cmark                                 & \cellcolor{LightGray}                                          & \cellcolor{LightGray}\cmark                & \cellcolor{LightGray}                              & \cellcolor{LightGray}\halfcirc  & \cellcolor{LightGray}\cmark & \cellcolor{LightGray}Diffusion      & \cellcolor{LightGray}DDIM~\cite{song2020denoising} & \cellcolor{LightGray}nuScenes, Carla                                    & \cellcolor{LightGray}      & \cellcolor{LightGray}BEVWorld \cite{zhang2024bevworld}         \\
                                                                                                          & \cmark                             &                                                                    &                                                             & \cmark                                                         & \cmark                                     &                                                    & \halfcirc                       &                             & Diffusion                           & SVD~\cite{blattmann2023stable}                      & BDD \cite{xu2017end}, etc \cite{yang2024generalized, grauman2024ego}.   & \cmark                      & GEM \cite{hassan2024gem}                                       \\ %BDD, OpenDV, NuScenes,
            \midrule \multirow{2}{*}{\textbf{\shortstack{Benchmark}}}     & \cellcolor{LightGray}\cmark        & \cellcolor{LightGray}\cmark                                        & \cellcolor{LightGray}\cmark                                 & \cellcolor{LightGray}\cmark                                    & \cellcolor{LightGray}                      & \cellcolor{LightGray}                              & \cellcolor{LightGray}\fullcirc  & \cellcolor{LightGray}       & \cellcolor{LightGray}Autoregression & \cellcolor{LightGray}Transformer & \cellcolor{LightGray}nuScenes, In-house                                 & \cellcolor{LightGray}\cmark & \cellcolor{LightGray}ACT-Bench \cite{arai2024act}              \\
                                                                                                          & \cmark                             & \cmark                                                             &                                                             &                                                                & \cmark                                     & \cmark                                             & \halfcirc                       & \cmark                      & Diffusion                           & SVD                      & nuScenes                                                                & \cmark                      & DriveArena \cite{DriveArena_2023}                              \\
            \bottomrule
        \end{tabular}}
        \begin{tablenotes}
            \scriptsize \item[1] Geometry means 3D geometric representation and includes:
            3D voxel occupancy, 3D bounding box, 3D depth, 3D segmentation and 3D
            point cloud. \scriptsize \item[2]
            \begin{minipage}[t]{0.75\linewidth}
            Controllability: \fullcirc~Full control (models offer fine-grained
            scene customization with flexible control over scene elements); \halfcirc~Partial
            control (models support limited or\\ parameterized control (e.g., adjusting
            map); \emptycirc~No control.
            \end{minipage}
            \scriptsize \item[3]
            \begin{minipage}[t]{0.75\linewidth}
            Code availability: "\cmark" means code is released open-source.%;  "-" means code is either not announced to be released in the paper, or not fully released yet;"\xmark" means code is announced to\\be released in the paper, but the full code (training and inference) is still not available.
            \end{minipage}
        \end{tablenotes}
    \end{threeparttable}
    \label{tab:ad_wm_comparison}
\end{table*}

\textbf{Architecture and Evolution of World Models:}
The architecture of \glspl{wm} is typically based on an \textit{encoder-decoder} paradigm~\cite{ha2018world,
lecun2022path, brooks2024opensora, Drive-WM2024}. As illustrated in Figure~\ref{fig:wm_overview}, the \textit{encoder} (also called \textit{vision} model \cite{ha2018world}) is used to encode multimodal inputs (images, point clouds, 3D occupancy voxels, etc.) into a latent vector $z_t$. Then, the future predictor (\textit{decoder} or \textit{memory} model \cite{ha2018world}) predicts the future latent representation $\tilde{z}_{t+1}$ based on $z_t$ and an \textit{action} provided by a given control policy. When the \gls{wm}'s pre-training is finished, the future predictor can be used for both motion prediction and for ``dreaming'' (i.e., generation) of new scenarios, never seen during training. In this regard, \glspl{wm} can generate data outside the training data distribution: this is particularly valuable for \gls{ad}, where rare but critical scenarios may be underrepresented in existing datasets, yet are crucial in the validation phase.

\rev{
In addition to being designed individually, \glspl{wm} are now increasingly integrating inspirations from \glspl{llm}, \glspl{vlm}, and \glspl{mllm}, which have demonstrated a promising understanding of semantic context. More specifically, \glspl{wm} are placing emphasis on using this semantic understanding for content generation~\cite{wiedemer2025video}.
Moreover, \glspl{dm} play an important role as generative backbones of most modern \glspl{wm}, providing stable and high-fidelity generation in both images and videos.
GAIA-2~\cite{russell2025gaia}, DriveDreamer~\cite{wang2024drivedreamer}, and MagicDrive3D~\cite{MagicDrive3D_2024} are examples of this trend, which employ either latent or video diffusion to increase the temporal coherence and realism of the generated scenarios. Together, these examples show that \glspl{wm} are developing into hybrid architectures that combine \glspl{vlm}' multimodal reasoning capabilities with \glspl{dm}' generative accuracy to create coherent, controllable, and semantically grounded driving simulations.}

\rev{
As shown in Figure~\ref{fig:wm_overview}, in the early examples \cite{ha2018world}, the vision model (encoder) was implemented as a Variational Autoencoder (\gls{vae}), and compresses high-dimensional observations into a compact latent representation. This dimensionality reduction creates a manageable state space for prediction and generation. The future predictor (memory model) was implemented as a recurrent network (e.g., \gls{lstm} or \gls{gru}). The memory model captures temporal dependencies and dynamics across sequential observations, enabling the prediction of future states. Modern \glspl{wm} for \gls{ad} have improved this basic architecture to incorporate advanced techniques into the future predictor. For example, GAIA-1~\cite{hu2023gaia} uses a transformer, and the newer GAIA-2~\cite{russell2025gaia} employs a Latent Diffusion Model (\gls{ldm})~\cite{rombach2022high} for future prediction and generation. Very recently, Diffusion Transformers (\glspl{dit})~\cite{peebles2023dit}, \gls{svd}~\cite{blattmann2023stable} models and videoLDM~\cite{blattmann2023align} have gained popularity as core architectures for \glspl{wm}.}

\begin{figure}[ht!]
    \centering
    \includegraphics[width=0.88\columnwidth]{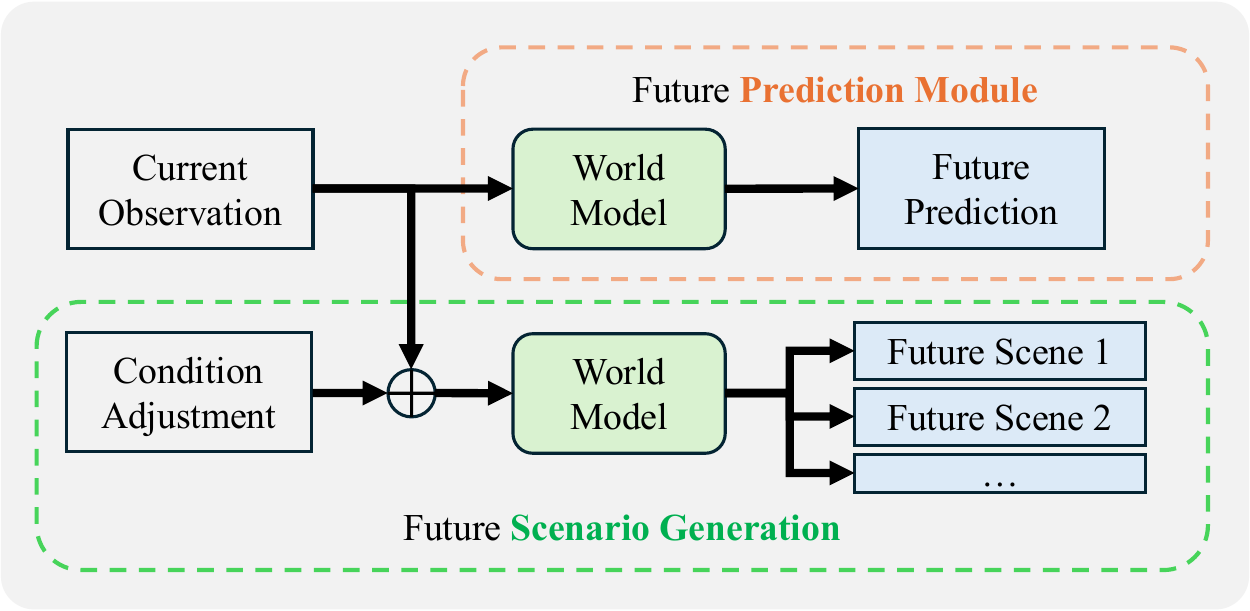}
    \caption{An illustration of how research on \glspl{wm} for \gls{ad} can be broadly categorized into two main functions: future prediction of agents' motion, and future scenario generation.}
    \label{fig:wm_functions}
    \vspace{-1mm}
\end{figure}

%Future motion prediction focuses on predicting a limited set of potential future scenarios based on the current state of the environment, including ego vehicle dynamics, surrounding vehicle behaviors, and pedestrian intentions. These predictions are typically used online to support downstream tasks such as planning and control.
%In contrast, future scenario generation, as emphasized in previous sections, aims to increase the diversity of driving data—particularly for rare or safety-critical cases—by introducing controlled variations such as altering weather conditions, adding new traffic participants, or pre-defining specific ego trajectories.

%The existing studies mostly use \glspl{wm} for scenario generation or prediction modules. 
As illustrated in Figure~\ref{fig:wm_functions}, \glspl{wm} can generally be used for two purposes in \gls{ad}: future motion prediction~\cite{hu2022model} and scenario generation~\cite{hu2023gaia, russell2025gaia}. In this section we focus primarily on the application of \glspl{wm} for scenario generation. The reviewed papers, their corresponding datasets and their code availability are summarized in Table \ref{tab:ad_wm_comparison}.

\subsection{Scenario Generation with World Model Dreaming} \label{sec:wm_dream}
\gls{wm} dreaming~\cite{ha2018world} is the use of a trained \gls{wm} to generate new scenarios by sampling from its learned latent space without additional real-world inputs. Once a \gls{wm} has captured the underlying dynamics of an environment, it can ``dream'' new scenarios that follow similar physical and logical patterns as those in the training data, but with new combinations of elements and conditions that may not have been seen during training. As shown in Table~\ref{tab:ad_wm_comparison}, recent research on \glspl{wm} for \gls{ad} can be categorized into the following four groups.

%The dreaming process involves applying recursively the prediction model (Figure~\ref{fig:wm_overview}), in an auto-regressive fashion, to generate a temporal sequence of future states. Introducing controlled perturbations or conditioning signals during this recursive generation process can guide the \gls{wm} toward generating specific types of scenarios.
 %\textit{Visual Generation}, \textit{3D Occupancy Generation}, \textit{Multi-modal Generation}, and \textit{Evaluation \& Benchmarking}.
% In the following of this section (sec.~\ref{sec:wm_dream}), we use this categorization to discuss recent representative research on scenario generation using world model "dreaming".
\rev{
\textbf{Visual Generation:} This approach focuses on creating realistic driving scenarios through the generation of images and videos. They represent the most mature category of \gls{wm} applications in \gls{ad}.
GAIA-1~\cite{hu2023gaia} pioneered the use of generative \glspl{wm} for
\gls{ad}, by demonstrating the ability to generate diverse traffic scenarios with multiple interacting agents. GAIA-1 considers world modeling as an unsupervised sequence modeling problem, mapping multimodal inputs (video, text, and action) to discrete tokens and predicting subsequent tokens. This approach enables fine-grained control over ego-vehicle behavior and scene features, showing emerging properties such as contextual awareness and 3D geometry understanding. 
GAIA-2~\cite{russell2025gaia} significantly advances the GAIA-1 paradigm through a latent diffusion \gls{wm} that supports controllable video generation conditioned on structured inputs (e.g., ego-vehicle dynamics and agent configurations). 
GAIA-2 generates high-resolution, spatio-temporally consistent multi-camera videos across diverse driving environments and countries (UK, US, Germany), making it a useful tool for complex scenario simulation with good multi-view consistency.}

\rev{To address the limitations of prior \glspl{wm}, DriveDreamer~\cite{wang2024drivedreamer} introduces a model entirely derived from real-world driving scenarios. Using its \gls{ad} Diffusion Model (Auto-\glspl{dm}) and a two-stage training pipeline, DriveDreamer first learns traffic structural constraints and then anticipates future states through video prediction. This approach excels in generating controllable driving videos and predicting driving policies, thereby enhancing perception tasks such as 3D detection.
DriveDreamer-2~\cite{DriveDreamer2_2024} extends the DriveDreamer framework \cite{wang2024drivedreamer}
by incorporating an \gls{llm} to generate user-defined driving videos.
DriveDreamer-2 converts user queries into agent trajectories and employs a unified multi-view model to ensure temporal and spatial coherence. It can also produce uncommon scenarios, such as abrupt vehicle cut-ins. DriveDreamer4D~\cite{DriveDreamer4D_2024} extends the DriveDreamer framework to 4D (spatio-temporal) scene representation.
By incorporating map, layout, and text conditioning, it enhances the realism of the generated data.}

\rev{Unlike traditional modular designs, ADriver-I~\cite{ADriver-I_2023} introduces a unified \gls{wm} using interleaved vision-action pairs to standardize visual features and control signals.} Using \glspl{mllm} and diffusion, it autoregressively predicts control signals and forecasts future frames, creating a continuous simulation loop. 
%Vista~\cite{gao2024vista} focuses on high-resolution controllable scenes generation using video diffusion techniques with versatile controllability features.It demonstrates superior performance in generating realistic driving videos with precise control over scene elements and ego-vehicle behavior, making it a robust simulation tool for scenario generation.
Also following the autoregressive style, DrivingWorld~\cite{hu2024drivingworld} introduces a GPT-style \gls{wm} for \gls{ad}, featuring spatial-temporal fusion mechanisms. It
employs next-state and next-token prediction strategies to model temporal
coherence and spatial information, implementing masking and reweighting strategies to mitigate long-term drifting and improve 3D detection and motion forecasting.

\rev{As a framework for street view generation with diverse 3D geometry controls, MagicDrive~\cite{gao2023magicdrive} includes camera poses, road maps, and 3D bounding boxes, along with textual descriptions. It addresses the challenge of 3D control in traditional \glspl{dm}, offering high-fidelity video generation with nuanced 3D geometry and multi-camera consistency.
MagicDrive3D~\cite{MagicDrive3D_2024} presents a pipeline for controllable 3D street scene generation that supports multi-condition control, including \gls{bev} maps, 3D objects, and text descriptions. Unlike methods that reconstruct scenes before training, it first trains a video generation model and then reconstructs 3D scenes from generated data, enabling high-quality scene reconstruction for any-view rendering.}

%MagicDrive-V2~\cite{gao2024magicdrivedit} builds on previous work to address scalability and geometry control in video generation. It introduces the MVDiT block and spatial-temporal conditional encoding to enable multi-view video generation with precise geometric control, generating high-resolution, long-duration videos (e.g., 848×1600, 193 frames at 10 FPS) with consistent spatial and temporal coherence.

\rev{A world volume-aware \gls{dm} is introduced by WoVoGen~\cite{lu2024wovogen} for generating controllable multi-camera driving scenes. It operates by predicting explicit 3D world volumes to guide video generation, ensuring that multi-camera perspectives align accurately with the underlying scene geometry, and maintaining high spatial and inter-sensor consistency.
While ReconDreamer~\cite{ni2024recondreamer} focuses on not only crafting \glspl{wm} for driving scene reconstruction but also online restoration. It emphasizes online learning for real-time applications, allowing continuous updates to the \gls{wm} as new data is acquired, which is critical for adaptability to changing conditions in \gls{ad}.}

\rev{With a dual-branch \gls{dm} for high-fidelity video generation, 
% DualDiff+~\cite{DualDiff_2024} incorporates Occupancy Ray Sampling for semantic-rich 3D representation and Semantic Fusion Attention to integrate multi-modal data with a foreground-aware masked loss, thereby improving the generation of small
% objects.
GeoDrive~\cite{chen2025geodrive} integrates 3D geometry conditions into driving \glspl{wm}. It enhances spatial understanding and action controllability through 3D video rendering with dynamic editing and control for spatio-temporal consistency, improving video quality with minimal training data.}

\textbf{3D Occupancy Generation:} 
3D occupancy generation predicts and generates volumetric representations of driving environments, capturing both the spatial structure and the temporal dynamics of scenes.
\rev{By treating 4D occupancy scene evolution as a video prediction task, OccSora~\cite{wang2024occsora} presents a novel a 4D scene tokenizer to obtain compact spatio-temporal representations. Then, it trains a diffusion transformer to generate 4D occupancy conditioned on trajectory prompts, enabling trajectory-aware simulation of various driving scenarios.
To consider both static and dynamic elements in complex urban environments, Drive-OccWorld~\cite{yang2025driving} combines a planner with a dynamic \gls{wm} to predict 3D occupancy and flow from multi-view images. More specifically, it uses motion-aware \gls{bev} sequences as an intermediate representation, integrating multi-view video data with motion cues to achieve robust predictions.
Also aiming at improving prediction accuracy for static and dynamic objects, RenderWorld~\cite{yan2024renderworld} further tries balancing granularity and computational efficiency. It focuses on fine-grained occupancy prediction through a novel tokenization strategy which captures spatial relationships.}

\rev{Using a continuous variational autoencoder-like tokenizer, DOME~\cite{gu2024dome} performs 3D occupancy prediction to preserve intricate spatial information. Unlike discrete tokenization methods, DOME's continuous approach captures subtle geometric details while maintaining computational efficiency, using probabilistic modeling to enhance robustness concerning sensor noise and occlusions.
OccLLama~\cite{wei2024occllama} tries to integrate a multi-modal \gls{llm} as a core component for occupancy prediction. Unlike traditional models that rely solely on geometric or visual data, OccLLama uses the reasoning capabilities of \glspl{llm} to process multi-modal inputs, understanding complex scene semantics and object interactions for enhanced prediction accuracy.
While DriveWorld~\cite{min2024driveworld} focuses on 4D scene understanding from multi-view videos. This approach is separating static spatial context from dynamic temporal changes to enable precise occupancy prediction. The model relies on self-supervised learning to reduce dependence on annotated data, thereby enhancing scalability.}

\textbf{Multi-modal Generation:}
Multi-modal generation approaches integrate multiple sensor modalities and data types as input, and output multi-modal data that can include camera images, LiDAR point clouds and depth estimation.

\rev{Aiming to address limitations of single-modality approaches, HoloDrive~\cite{wu2024holodrive} introduces a unified framework for joint 2D-3D scene generation. It employs \gls{bev}-to-Camera and Camera-to-\gls{bev} transformation modules to bridge heterogeneous generative models. Therefore, it ensures consistency between 2D and 3D representations while using both camera images and LiDAR point clouds for the generation of consistent street scenes.
Further, GEM~\cite{hassan2024gem} proposes a framework for generating realistic environments by integrating multi-modal sensor data, including camera images, and depth estimation. It employs a generative model based on a spatial-temporal transformer capable of predicting dynamic scene evolution regarding visual generation and depth estimation.
BEVWorld~\cite{zhang2024bevworld} performed world modeling through a unified \gls{bev} latent space that also integrates multi-modal sensor inputs. The framework includes a multi-modal tokenizer and a latent \gls{bev} sequence \gls{dm} that encodes multi-modal data into a unified \gls{bev} latent space. This method aims at aligning visual semantics with geometric information in a self-supervised manner.}

\textbf{Benchmarks:}
%\label{sec:wm_e} - label no longer usable
Current benchmarking frameworks provide standardized methods to assess the quality, controllability, and utility of generated scenarios, ensuring that the \glspl{wm} meet the requirements for \gls{ad} applications.
Current evaluation frameworks mainly focus on visual realism and on the performance of downstream tasks (perception, planning, etc.).
ACT-Bench~\cite{arai2024act} introduces a standardized framework to quantify action controllability, measuring how well the generated scenarios adhere to specified driving instructions. This benchmarking framework assesses the fidelity of action execution in \gls{wm}-generated scenarios. 
DriveArena~\cite{DriveArena_2023} is a closed-loop generative simulation platform that enables the evaluation of \gls{ad} systems in dynamic and realistic environments. By simulating continuous interactions between the ego-vehicle and the environment, it bridges the gap between synthetic training and real-world deployment, supporting the iterative refinement of driving policies.

\subsection{Limitations and Future Directions}
Recent research on 3D occupancy generation with \glspl{wm} has shown promising capabilities in predicting the evolution of driving environments in volumetric form. However, most models remain computationally intensive: future work should aim to develop lightweight architectures and explore finer-grained occupancy voxel representations. \rev{Recent commercial systems, such as Tesla’s \emph{Foundational Model for FSD}\footnote{\url{https://www.tesla.com/AI}}, highlight both the potential and the remaining challenges of large-scale \glspl{wm}.} \revB{Meanwhile, general-purpose generative \glspl{wm}, such as Google DeepMind’s \emph{Genie~3}\footnote{\url{https://deepmind.google/blog/genie-3-a-new-frontier-for-world-models/}}, can output interactive 3D environments from prompts, showing the potentital for diverse synthetic scenario generation.}

Moreover, current implementations struggle with physical realism~\cite{wang2025enhancing, wang2025target} when modeling complex multi-agent interactions and real-world physics,  including vehicle dynamics and kinematics laws, tire–road friction, collision forces, and weather effects. \revB{This limitation also applies to general-purpose \glspl{wm} such as Genie~3, which are not tailored to \gls{ad} and cannot guarantee physics-consistent modeling of vehicle dynamics and traffic rules.} The generated scenarios sometimes contain physically implausible elements, such as objects that appear or disappear abruptly. Hence, the surveyed \glspl{wm} can generate diverse driving scenarios but cannot accurately satisfy the laws of physics, which can lead to misleading testing results and infeasible scenarios.

\hypertarget{sec:dataset}{}
\section{Metrics Datasets, Simulators and Benchmark Challenges}
\label{sec:dataset}
In this section, we review \rev{the main evaluation metrics,}  datasets, simulation platforms, and benchmark challenges that serve as the foundation for scenario generation and analysis with \glspl{fm}. We intentionally limit our scope to the most recent and impactful resources that are relevant for \gls{fm} applications, and omit entries covered in previous work.

\rev{
\subsection{Metrics}
\autoref{tab:metrics} summarizes the main evaluation metrics from the cited papers for scenario generation and analysis with \glspl{fm}. These metrics are categorized into three main types: \emph{(1) Framework Performance Metrics}, which assess the overall performance of frameworks; \emph{(2) Content Quality Metrics}, which evalutate the quality and semantic accuracy of the generated or analyzed content; and \emph{(3) Application-Specific Metrics}, which address domain-relevant aspects. }

\rev{
\emph{(1) Framework Performance Metrics}: They evaluate the computational efficiency and operational reliability of \gls{fm}-based frameworks for scenario generation and analysis.}

\rev{
(I) \textit{Efficiency:} Measures the computational cost and time required for scenario generation or analysis. \textbf{Response time} refers to time from input submission to output generation, while \textbf{token usage} quantifies the total number of input and output tokens consumed during \gls{api} key calls. These metrics are often compared against baseline approaches such as manual scripting to assess the practical benefits of \gls{fm}-based frameworks~\cite{cai2025text2scenario, petrovic2024llm}.}

\rev{
(II) \textit{Effectiveness:} Refers to the operational robustness and reliability of the framework in producing valid outputs. This is commonly evaluated through \textbf{compile error rate}, the proportion of generated code or scenarios that fail to compile or parse correctly, and \textbf{execution success rate}, the percentage of scenarios that can be successfully instantiated and executed in a target environment~\cite{cai2025text2scenario, petrovic2024llm}.
}

\rev{
\emph{(2) Content Quality Metrics}: These metrics assess the quality and semantic accuracy of the generated or analyzed content, including trajectories, semantic understanding, and language generation outputs.}

\rev{
(I) \textit{Trajectory Accuracy:} Crucial for trajectory-centric generation and prediction tasks. Common metrics include \textbf{\gls{made}}, the average Euclidean distance between predicted and ground truth trajectories across all time steps; \textbf{\gls{mfde}}, the distance at the final prediction time step from the predicted trajectory to the ground truth trajectory; and \textbf{\gls{mmd}}, which measures the distributional similarity between generated and real trajectory sets. Additionally, \textbf{Predictive Driver Model Score} evaluates the likelihood of predicted trajectories under human driving patterns learned from real-world data, and \textbf{Arena Driving Score} assesses overall driving competence in multi-agent scenarios by evaluating collision avoidance, goal achievement, traffic rule compliance, and interaction quality with other agents~\cite{tan2023language, li2024chatgpt, aiersilan2024generating, tan2024promptable, mei2025llm, dauner2024navsim, DriveArena_2023}.}

\rev{
(II) \textit{Semantic Correctness:} Assesses how well the generated scenarios or analysis outputs reflect the intended semantics of inputs like crash reports or textual prompts. Common metrics include \textbf{accuracy} or \textbf{F1 score} for evaluating scenario categorization, semantic classification, and question-answering correctness. Additionally, \textbf{completeness} and \textbf{coherence} are evaluated through human assessment, where annotators assign scores based on how thoroughly the response covers all relevant aspects and how logically consistent and well-structured the output is~\cite{zhao2024chat2scenario, wei2024editable, cai2025text2scenario, guo2024sovar, luo2025accidents, lu2024realistic, chen2024driving, rivera2025scenario, Keskar_2025_WACV, wen2023road,inoue2024nuscenes, choudhary2024talk2bev, ishaq2025drivelmm, zheng2024large, ding2024holistic, cao2024maplm, park2024vlaad, zhou2025tumtraffic, park2025nuplanqa, zeng2025vision, zhang2024interndrive, jain2024semantic, ishaq2025tracking}}

\rev{(III) \textit{Language Quality:} Evaluates how similar generated text is to human-written reference sentences, measuring fluency, relevance, and coherence based on word overlap, structure, and meaning. Traditional metrics include \textbf{\gls{bleu}}, which measures word and phrase (n-gram) overlap focusing on precision; \textbf{\gls{cider}}, which uses weighted n-grams giving more importance to informative words; \textbf{\gls{meteor}}, which considers exact matches, stem matches, and synonyms for both precision and recall; and \textbf{\gls{rougel}}, which measures content similarity using the longest common subsequence focusing on recall. However, these word-level metrics may not capture semantic nuances. To address this, \textbf{GPT Score} leverages ChatGPT's reasoning capabilities to evaluate prediction quality and semantic meaning, assigning scores. Additionally, \textbf{human evaluation scores} provide direct assessment of output quality by human annotators who rate the generated content based on observed details~\cite{inoue2024nuscenes, nie2024reason2drive, gopalkrishnan2024multi, chen2025automated,  xie2025vlms, zeng2025vision, jain2024semantic, ma2024dolphins, kong2024wts, sima2024drivelm}.
}

\begin{table*}[htp]
    \centering
    \caption{\rev{Overview of evaluation metrics for foundation model-based scenario generation and analysis.}}
    \label{tab:metrics}
    \renewcommand{\arraystretch}{1.1}
    \small
    \resizebox{1\textwidth}{!}{%
    \rev{\begin{tabular}{lllcccccccclc}
    \toprule
    \textbf{Category} & \textbf{Sub-Category} & \textbf{Metric} & \multicolumn{2}{c}{\textbf{Task}} & \multicolumn{5}{c}{\textbf{Model}} & \textbf{Output} & \textbf{Citations} \\
    \cmidrule(lr){4-5} \cmidrule(lr){6-10}
    & & & \textbf{Gen} & \textbf{Ana} & \textbf{LLM} & \textbf{VLM} & \textbf{MLLM} & \textbf{DM} & \textbf{WM} & & \\
    \midrule
    \multirow{4}{*}{\textbf{\makecell{Framework\\Performance}}} & \multirow{2}{*}{Efficiency} & Response Time & \checkmark & & \checkmark & \checkmark & & & & S & \cite{cai2025text2scenario, petrovic2024llm} \\
    & & \cellcolor{LightGray}Token Usage &\cellcolor{LightGray}\checkmark &\cellcolor{LightGray}&\cellcolor{LightGray}\checkmark &\cellcolor{LightGray} \checkmark &\cellcolor{LightGray}&\cellcolor{LightGray}&\cellcolor{LightGray}&\cellcolor{LightGray}S &\cellcolor{LightGray}\cite{cai2025text2scenario, petrovic2024llm} \\
    \cmidrule(lr){2-12}
    & \multirow{2}{*}{Effectiveness} & Compile Error Rate & \checkmark & & \checkmark &  & \checkmark & & & S & \cite{cai2025text2scenario, petrovic2024llm,lu2024realistic} \\
    & & \cellcolor{LightGray}Execution Success Rate &\cellcolor{LightGray}\checkmark &\cellcolor{LightGray}&\cellcolor{LightGray}\checkmark &\cellcolor{LightGray}&\cellcolor{LightGray}&\cellcolor{LightGray}&\cellcolor{LightGray}&\cellcolor{LightGray}S &\cellcolor{LightGray}\cite{cai2025text2scenario, petrovic2024llm} \\
    \midrule
    \multirow{10}{*}{\textbf{\makecell{Content\\Quality}}} & \multirow{4}{*}{\makecell{Trajectory\\Accuracy}} & mADE, mFDE & \checkmark & & \checkmark & & & &\checkmark & Tr & \cite{tan2023language, li2024chatgpt, aiersilan2024generating, tan2024promptable, mei2025llm, arai2024act} \\
    & & \cellcolor{LightGray}MMD &\cellcolor{LightGray}\checkmark &\cellcolor{LightGray}&\cellcolor{LightGray}\checkmark &\cellcolor{LightGray}&\cellcolor{LightGray}&\cellcolor{LightGray}&\cellcolor{LightGray}&\cellcolor{LightGray}Tr &\cellcolor{LightGray}\cite{tan2023language, li2024chatgpt, aiersilan2024generating, tan2024promptable, mei2025llm} \\
    & & Predictive Driver Model Score & \checkmark & & & & & &\checkmark  & Tr & \cite{DriveArena_2023} \\
    & & \cellcolor{LightGray}Arena Driving Score &\cellcolor{LightGray}\checkmark &\cellcolor{LightGray}&\cellcolor{LightGray} &\cellcolor{LightGray}&\cellcolor{LightGray}&\cellcolor{LightGray}&\cellcolor{LightGray}\checkmark &\cellcolor{LightGray}Tr &\cellcolor{LightGray}\cite{DriveArena_2023} \\
    \cmidrule(lr){2-12}
    & \multirow{2}{*}{\makecell{Semantic\\Correctness}} & Accuracy / F1 Score & \checkmark &\checkmark  & \checkmark &\checkmark  &\checkmark  & & & T/S & e.g.\cite{wei2024editable, cai2025text2scenario, guo2024sovar, luo2025accidents, rivera2025scenario} \\
    & & \cellcolor{LightGray}Completeness, Coherence &\cellcolor{LightGray} &\cellcolor{LightGray}\checkmark &\cellcolor{LightGray}\checkmark &\cellcolor{LightGray}&\cellcolor{LightGray}&\cellcolor{LightGray}&\cellcolor{LightGray}&\cellcolor{LightGray}T&\cellcolor{LightGray}\cite{chen2024driving} \\
    \cmidrule(lr){2-12}
    & \multirow{3}{*}{\makecell{Language\\Quality}} & BLEU, CIDEr, METEOR, ROUGE-L &  & \checkmark& & \checkmark & \checkmark & & & T & e.g. \cite{zhou2024openannotate3d, kou2025enhancing, gu2024dome, inoue2024nuscenes, nie2024reason2drive} \\
    & & \cellcolor{LightGray}GPT Score &\cellcolor{LightGray}&\cellcolor{LightGray}\checkmark &\cellcolor{LightGray}&\cellcolor{LightGray}\checkmark &\cellcolor{LightGray}&\cellcolor{LightGray}&\cellcolor{LightGray}&\cellcolor{LightGray}T &\cellcolor{LightGray}\cite{xie2025vlms, sima2024drivelm} \\
    & & Human Evaluation &  &\checkmark & & & \checkmark & & & T & \cite{ma2024dolphins, jain2024semantic, kong2024wts} \\
    \midrule
    \multirow{25}{*}{\textbf{\makecell{Application-\\Specific}}} & \multirow{5}{*}{\makecell{Safety-\\Criticality}} & Collision Rate & \checkmark &\checkmark &\checkmark &\checkmark & & \checkmark & & S/Tr & e.g. \cite{zhang2024chatscene, mei2025seeking, sheng2025curricuvlm, xie2024advdiffuser, lu2024data} \\
    & & \cellcolor{LightGray}TTC&\cellcolor{LightGray}\checkmark &\cellcolor{LightGray}\checkmark &\cellcolor{LightGray}&\cellcolor{LightGray}\checkmark  &\cellcolor{LightGray}\checkmark &\cellcolor{LightGray}&\cellcolor{LightGray}&\cellcolor{LightGray}T/S/Tr &\cellcolor{LightGray}\cite{mei2025seeking, gao2025risk, sheng2025curricuvlm, zhang2025latte} \\
    & & Risk Score & \checkmark &\checkmark &\checkmark &  & & & & T/Tr & \cite{gao2025risk} \\
    & & \cellcolor{LightGray}Accuracy&\cellcolor{LightGray} &\cellcolor{LightGray}\checkmark &\cellcolor{LightGray}&\cellcolor{LightGray}\checkmark &\cellcolor{LightGray}&\cellcolor{LightGray}\checkmark   &\cellcolor{LightGray}&\cellcolor{LightGray}T &\cellcolor{LightGray}e.g. \cite{fan2024mllm, lee2025sff, chen2025insightenhancingautonomousdriving, wang2025bridging} \\
    & & Violation Discovery & \checkmark & & \checkmark & &\checkmark & & & S & \cite{tian2024llm} \\
    \cmidrule(lr){2-12}
    & \multirow{4}{*}{Controllability} &\cellcolor{LightGray} CLIP Alignment Score &\cellcolor{LightGray} \checkmark &\cellcolor{LightGray} &\cellcolor{LightGray} &\cellcolor{LightGray} &\cellcolor{LightGray} &\cellcolor{LightGray} \checkmark &\cellcolor{LightGray} &\cellcolor{LightGray} I/V &\cellcolor{LightGray} \cite{gu2025text2street, yang2023bevcontrol} \\
    % & & \cellcolor{LightGray}Object Detection/Segmentation &\cellcolor{LightGray}\checkmark &\cellcolor{LightGray}&\cellcolor{LightGray}&\cellcolor{LightGray}&\cellcolor{LightGray}&\cellcolor{LightGray}\checkmark &\cellcolor{LightGray}&\cellcolor{LightGray}I &\cellcolor{LightGray}\cite{chen2023geodiffusion, gao2023magicdrive} \\
    & & Accuracy & \checkmark & & \checkmark & & & &\checkmark & S/I/V & e.g. \cite{yang2023bevcontrol, chen2023geodiffusion, gao2023magicdrive, wen2024panacea, li2023drivingdiffusion, DualDiff_2024, jiang2025dive, bai2025dctdm, li2025dualdiff-, marathe2023wedge} \\
    & & \cellcolor{LightGray}Traffic Flow Compliance &\cellcolor{LightGray}\checkmark &\cellcolor{LightGray}&\cellcolor{LightGray}&\cellcolor{LightGray}&\cellcolor{LightGray}&\cellcolor{LightGray}\checkmark &\cellcolor{LightGray}&\cellcolor{LightGray}Tr &\cellcolor{LightGray}e.g. \cite{xu2023diffscene, zhong2023guidedcond, lu2024data, zhong2023languagediff, chang2024safesim} \\
    \cmidrule(lr){2-12}
    & \multirow{10}{*}{Realism} & WD, KLD & \checkmark & & & & & \checkmark & & Tr & e.g. \cite{barrow1977parametric, kim2020video, gu2024dome, luo2025accidents, miao2024dashcam} \\
    & & \cellcolor{LightGray}SSPD, Frechet Distance&\cellcolor{LightGray}\checkmark &\cellcolor{LightGray}&\cellcolor{LightGray}&\cellcolor{LightGray}&\cellcolor{LightGray}&\cellcolor{LightGray}\checkmark &\cellcolor{LightGray}&\cellcolor{LightGray}Tr &\cellcolor{LightGray}\cite{rowe2025scenariodreamer, xu2023diffscene} \\
    & & Off-Road Rate & \checkmark & & & & & \checkmark & & Tr & e.g. \cite{yu2025dpo, zhong2023guidedcond, pronovost2023scenariodiff, chang2024safesim, huang2024versatile} \\
    & & \cellcolor{LightGray}Lane Heading Distance &\cellcolor{LightGray}\checkmark &\cellcolor{LightGray}&\cellcolor{LightGray}&\cellcolor{LightGray}&\cellcolor{LightGray}&\cellcolor{LightGray}\checkmark &\cellcolor{LightGray}&\cellcolor{LightGray}Tr &\cellcolor{LightGray}\cite{pronovost2023scenariodiff} \\
    & & FID, RMSE & \checkmark & & & & & \checkmark & & V & \cite{heusel2017gans, yang2023bevcontrol, chen2023geodiffusion, gao2023magicdrive, gu2025text2street, li2025dualdiff-, marathe2023wedge} \\
    & & \cellcolor{LightGray}FVD, KVD &\cellcolor{LightGray}\checkmark &\cellcolor{LightGray}&\cellcolor{LightGray}&\cellcolor{LightGray}&\cellcolor{LightGray}&\cellcolor{LightGray}\checkmark  &\cellcolor{LightGray}\checkmark &\cellcolor{LightGray}I/V &\cellcolor{LightGray}\cite{guo2024drivinggen, fu2024drivegenvlm, li2025avd2, wen2024panacea, li2023drivingdiffusion, DualDiff_2024, jiang2025dive, bai2025dctdm, unterthiner2018towards} \\
    & & Video Panoptic Quality & \checkmark & & & & & & \checkmark & V & \cite{kim2020video, yang2025driving} \\
    & & \cellcolor{LightGray}mIoU &\cellcolor{LightGray}\checkmark &\cellcolor{LightGray}&\cellcolor{LightGray}&\cellcolor{LightGray}&\cellcolor{LightGray}&\cellcolor{LightGray}&\cellcolor{LightGray}\checkmark &\cellcolor{LightGray}O &\cellcolor{LightGray}\cite{gu2024dome} \\
    & & Chamfer Distance & \checkmark & & & & & \checkmark & & O & \cite{pronovost2023scenariodiff, chang2024safesim} \\
    & & \cellcolor{LightGray}Human Evaluation &\cellcolor{LightGray}\checkmark &\cellcolor{LightGray}&\cellcolor{LightGray} &\cellcolor{LightGray}\checkmark&\cellcolor{LightGray}&\cellcolor{LightGray} &\cellcolor{LightGray} &\cellcolor{LightGray}S &\cellcolor{LightGray}\cite{luo2025accidents, miao2024dashcam} \\
    & & Scenario Consistency & \checkmark & & &\checkmark &\checkmark &  & & S & \cite{lu2024realistic, luo2025accidents, miao2024dashcam} \\
    \cmidrule(lr){2-12}
    & Diversity &\cellcolor{LightGray}Statistical Distribution &\cellcolor{LightGray}\checkmark &\cellcolor{LightGray}\checkmark &\cellcolor{LightGray}\checkmark & \cellcolor{LightGray}\checkmark&\cellcolor{LightGray}\checkmark&\cellcolor{LightGray}&\cellcolor{LightGray}&\cellcolor{LightGray}T/S/I &\cellcolor{LightGray}\cite{chang2024llmscenario, aasi2024generating, marathe2023wedge, wang2025openlka, lu2024realistic} \\
    % \cmidrule(lr){2-12}
    % & Perception & Class-wise Average Precision & \checkmark & & & & & & \checkmark & I & \cite{jiang2025dive, bai2025dctdm} \\
    \cmidrule(lr){2-12}
    & \multirow{3}{*}{Grounding} & IoU / mIoU &  &\checkmark & & \checkmark&  & &\checkmark & T & \cite{zhou2024openannotate3d, kou2025enhancing,ding2023hilm, zhou2025tumtraffic, gu2024dome} \\
    & & \cellcolor{LightGray}3D mAP &\cellcolor{LightGray} &\cellcolor{LightGray}\checkmark &\cellcolor{LightGray}&\cellcolor{LightGray}\checkmark  &\cellcolor{LightGray} &\cellcolor{LightGray}&\cellcolor{LightGray}&\cellcolor{LightGray}T &\cellcolor{LightGray}\cite{li2025nugrounding, najibi2023unsupervised} \\
    & & L1/L2 Localization Error &  &\checkmark & & & \checkmark & & & T & \cite{ding2023hilm, zhou2025tumtraffic} \\
    \cmidrule(lr){2-12}
    &Classification & \cellcolor{LightGray}Accuracy, Precision, Recall, Confusion Matrix &\cellcolor{LightGray}&\cellcolor{LightGray}\checkmark &\cellcolor{LightGray}\checkmark &\cellcolor{LightGray}&\cellcolor{LightGray}\checkmark &\cellcolor{LightGray}&\cellcolor{LightGray}&\cellcolor{LightGray}T &\cellcolor{LightGray}\cite{zhou2024openannotate3d, kou2025enhancing, ding2023hilm, zhou2025tumtraffic, gu2024dome} \\
    \bottomrule
    \end{tabular}}
    } % close resizebox
    \rev{
    \begin{tablenotes}
        \scriptsize 
        \item[*] \textbf{Task}: Gen = Generation, Ana = Analysis. Checkmarks indicate applicable tasks.
        \item[*] \textbf{Output}: T = Text (question answering), S = Script (executable code), Tr = Trajectory (single or multi-agent paths), I = Image (2D scenes), V = Video (temporal sequences),\\O = Others (point cloud, 3D occupancy, depth map).
    \end{tablenotes}
    }
\end{table*}

\rev{
\emph{(3) Application-Specific Metrics}: They address domain-specific aspects of \gls{ad} scenarios, focusing on safety-critical properties and user-specified constraints.}

\rev{
(I) \textit{Safety-Criticality:} Evaluate the risk levels and safety-critical properties of generated scenarios. Key metrics include \textbf{collision rate}, the frequency of collisions occurring in the scenario; \textbf{\gls{ttc}}, the time remaining before a potential collision; \textbf{Risk score}, a comprehensive assessment of scenario danger level;  \textbf{Accuracy} for
evaluating safety criticality of scenarios; and \textbf{violation discovery}, the ability to identify and detect safety-critical events or rule violations in the generated scenarios~\cite{sheng2025curricuvlm, zhang2024chatscene, mei2025seeking, zhang2025latte,  chen2025insightenhancingautonomousdriving, ronecker2025vision, xie2024advdiffuser,  wang2025bridging, lu2024data, lin2024CCDiff, peng2025ldscene, xu2023diffscene, zhong2023guidedcond, pronovost2023scenariodiff, chang2024safesim, huang2024versatile, rowe2025scenariodreamer, tian2024llm}.}

\rev{
(II) \textit{Controllability:} Measures the framework's ability to follow user-specified constraints and control signals. Key metrics include \textbf{\gls{clip} Alignment Score}, which measures alignment between visual content and textual prompts via cosine similarity in \gls{clip}'s shared embedding space; \textbf{Accuracy}, which evaluate the correctness of generated content against specified control signals, such as verifying the presence, location, and class of elements via object detection; and \textbf{traffic flow compliance}, which assesses adherence to constraints such as speed, waypoints, lane assignments, vehicle counts, and scene type specifications~\cite{gu2025text2street, yang2023bevcontrol, chen2023geodiffusion, gao2023magicdrive, li2025dualdiff-, wen2024panacea, li2023drivingdiffusion, DualDiff_2024, jiang2025dive, bai2025dctdm, xu2023diffscene,lu2024data, zhong2023guidedcond,yu2025dpo, chang2024safesim, zhong2023languagediff, marathe2023wedge}.}

\rev{
(III) \textit{Realism:} Measures the realism of generated scenarios across multiple modalities. For traffic flow, metrics include \textbf{\gls{wd}} and \textbf{\gls{kld}} for statistical realism of motion dynamics (e.g., acceleration, jerk), \textbf{Frechet Distance} and \textbf{\gls{sspd}} for spatial differences between simulated and ground truth trajectories, \textbf{Off-Road Rate} for unrealistic trajectory generation, and \textbf{Lane Heading Distance} for alignment between vehicle orientation and lane direction. For image generation, \textbf{\gls{fid}} measures distributional discrepancies, and \textbf{\gls{rmse}} evaluates pixel-level accuracy. For video generation, \textbf{\gls{fvd}}, \textbf{\gls{kvd}}, and \textbf{Video Panoptic Quality} assess temporal coherence and statistical similarity. For 3D scenarios, \textbf{\gls{miou}} evaluates occupancy prediction, and \textbf{chamfer distance} measures point cloud similarity. Additionally, \textbf{scenario consistency}, and \textbf{human evaluation} assess overall scenario quality and realism~\cite{barrow1977parametric, kim2020video, gu2024dome, luo2025accidents, miao2024dashcam, lu2024realistic, marathe2023wedge, wu2024reality, elhafsi2023semantic, guo2024drivinggen, fu2024drivegenvlm, li2025avd2, wen2024panacea, li2023drivingdiffusion, DualDiff_2024, jiang2025dive, bai2025dctdm, unterthiner2018towards, yang2023bevcontrol, chen2023geodiffusion, gao2023magicdrive, gu2025text2street, li2025dualdiff-, heusel2017gans, pronovost2023scenariodiff, yu2025dpo, zhong2023guidedcond,  chang2024safesim, huang2024versatile, lin2024CCDiff, peng2025ldscene, xu2023diffscene, rowe2025scenariodreamer}.
}

\rev{
(IV) \textit{Diversity:} Captures the variability of generated scenarios by analyzing \textbf{statistical distributions} of features such as lane counts, edge counts, route lengths, and vehicle densities~\cite{wang2025openlka, chang2024llmscenario, aasi2024generating}.
}

\rev{
(VI) \textit{Grounding:} Evaluates how accurately models can ground textual descriptions to visual elements and understand spatial relationships in the driving scene. Key metrics include \textbf{\gls{iou}} and \textbf{\gls{miou}} for 2D and 3D object localization accuracy, \textbf{3D \gls{map}} for detecting and localizing objects in 3D space, and \textbf{L1/L2 localization error} for measuring spatial deviation between predicted and ground truth object positions~\cite{zhou2024openannotate3d, kou2025enhancing, ding2023hilm, zhou2025tumtraffic, gu2024dome, li2025nugrounding, najibi2023unsupervised}.
}

\rev{
(VII) \textit{Classification:} Assesses the accuracy of categorizing scenarios, behaviors, or driving conditions. Common metrics include \textbf{accuracy} for scenario type identification, \textbf{confusion matrix} for understanding misclassification patterns, and \textbf{precision/recall} for specific safety-critical event detection~\cite{you2025comprehensive, shi2024scvlm, gao2025risk}.
}

\subsection{Datasets}
\begin{table*}[htp]
\centering
\caption{Overview of impactful and recent datasets for foundation model-based scenario generation and analysis.}
\label{tab:datasets}
\renewcommand{\arraystretch}{1}
\resizebox{1\textwidth}{!}{%
\begin{tabular}{l lcccccccccccccc}
\toprule
& \multirow{2}{*}{\textbf{Dataset}} & \multirow{2}{*}{\textbf{Year}} & \multirow{2}{*}{\textbf{Real}} & \multirow{2}{*}{\textbf{View}} & \multicolumn{4}{c}{\textbf{Sensor Data}} & \multicolumn{3}{c}{\textbf{Annotation}} & \multicolumn{4}{c}{\textbf{Traffic Condition}} \\
\cmidrule(lr){6-9} \cmidrule(lr){10-12} \cmidrule(lr){13-16}
& & & & & \textbf{Image} & \textbf{LiDAR} & \textbf{RADAR} & \textbf{Traj.} & \textbf{3D} & \textbf{2D} & \textbf{Lane} & \textbf{Weather} & \textbf{Time} & \textbf{Region} & \textbf{Jam} \\
\midrule

\multirow{4}{*}{\rotatebox[origin=c]{90}{\textbf{Impactful}}}
% Name         & Year & Real   & View     & Image & LiDAR  & RADAR  & Traj.  & 3D     & 2D     & Lane   & Weathe & Tim & Region  & Jam
& HighD \cite{Krajewski2018highd}       & 2018 & \cmark & BEV      & RGB   & \xmark & \xmark & \cmark & \xmark & \cmark & \xmark & \xmark & D   & H       & \cmark \\
& \cc nuScenes \cite{caesar2020nuscenes}    & \cc 2020 & \cc \cmark & \cc FPV      & \cc RGB   & \cc \cmark & \cc \cmark & \cc \cmark & \cc \cmark & \cc \cmark & \cc \xmark & \cc \cmark & \cc D/N & \cc U & \cc \xmark \\
& Waymo Open \cite{sun2020waymoopen}  & 2020 & \cmark & FPV      & RGB   & \cmark & \xmark & \cmark & \cmark & \cmark & \cmark & \cmark & D/N & U/S     & \xmark \\
%& \cc Argoverse 2 \cite{wilson2023argoverse} & \cc 2021 & \cc \cmark & \cc FPV      & \cc RGB/S & \cc \cmark & \cc \xmark & \cc \cmark & \cc \cmark & \cc \xmark & \cc \cmark & \cc \cmark & \cc D/N & \cc U & \cc \xmark \\
& \cc DRAMA \cite{malla2023drama} & \cc 2022 & \cc \cmark & \cc FPV      & \cc RGB & \cc \xmark & \cc \xmark & \cc \cmark & \cc \xmark & \cc \cmark & \cc \xmark & \cc \xmark & \cc - & \cc U & \cc \cmark \\

\midrule
% Name         & Year & Real   & View     & Image & LiDAR  & RADAR  & Traj.  & 3D     & 2D     & Lane   & Weathe & Tim & Region  & Jam
\multirow{17}{*}{\rotatebox[origin=c]{90}{\textbf{Most Recent}}}
& Comma2k19 \cite{comma2k19}   & 2019 & \cmark & FPV      & RGB   & \xmark & \xmark & \cmark & \cmark & \xmark & \xmark & \xmark & D/N & U/S/R/H & \cmark \\
& \cc Toronto3D \cite{Tan2020toronto3d}   & \cc 2020 & \cc \cmark & \cc BEV      & \cc RGB   & \cc \cmark & \cc \xmark & \cc \cmark & \cc \cmark & \cc \xmark & \cc \cmark & \cc \xmark & \cc D/N & \cc U       & \cc \cmark \\
& A2D2 \cite{a2d2} & 2020 & \cmark & FPV      & RGB   & \cmark & \cmark & \cmark & \cmark & \xmark & \cmark & \cmark & D   & U/S/R/H & \cmark \\
& \cc WADS \cite{Kurup2022wads} & \cc 2020 & \cc \cmark & \cc FPV      & \cc RGB   & \cc \cmark & \cc \cmark & \cc \cmark & \cc \cmark & \cc \xmark & \cc \xmark & \cc \cmark & \cc D/N & \cc U/S/R & \cc \cmark \\
& SeethroughFog \cite{Bijelic2020-stf} & 2020 & \cmark & FPV    & RGB   & \cmark & \cmark & \cmark & \cmark & \xmark & \cmark & \cmark & D/N & U/S/R/H & \cmark \\
& \cc Leddar PixSet \cite{Deziel2021-pixset} & \cc 2021 & \cc \cmark & \cc FPV & \cc  RGB   & \cc \cmark & \cc \xmark & \cc \cmark & \cc \cmark & \cc \cmark & \cc \xmark & \cc \cmark & \cc D/N & \cc U/S/R   & \cc \cmark \\
& ZOD \cite{Alibeigi2023-zod}  & 2022 & \cmark & FPV      & RGB   & \cmark & \cmark & \cmark & \cmark & \cmark & \cmark & \cmark & D/N & U/S/R/H & \cmark \\
& \cc IDD-3D \cite{Dokania2023-idd3d} & \cc 2022 & \cc \cmark & \cc FPV      & \cc RGB   & \cc \cmark & \cc \xmark & \cc \xmark & \cc \cmark & \cc \cmark & \cc \xmark & \cc \xmark &  \cc -  & \cc R & \cc \cmark \\
& CODA \cite{li2022coda}  & 2022 & \cmark & FPV      & RGB   & \cmark & \cmark & \cmark & \cmark & \cmark & \cmark & \cmark & D/N & U/S/R   & \xmark \\
& \cc SHIFT \cite{Sun2022-shift} & \cc 2022 & \cc \cmark & \cc FPV & \cc RGB   & \cc \cmark & \cc \cmark & \cc \cmark & \cc \cmark & \cc \cmark & \cc \cmark & \cc \cmark & \cc D/N & \cc U/S/R/H & \cc \cmark \\
& DeepAccident \cite{wang2024deepaccident} & 2023 & \xmark & FPV/BEV  & RGB/S & \cmark & \xmark & \xmark & \cmark & \cmark & \cmark & \cmark & D/N & U/S/R/H & \cmark \\
& \cc Dual\_Radar \cite{Zhang2025-dualradar} & \cc 2023 & \cc \cmark & \cc FPV      & \cc RGB   & \cc \cmark & \cc \cmark & \cc \cmark & \cc \cmark & \cc \xmark & \cc \cmark & \cc \cmark & \cc D/N & \cc U & \cc \xmark \\
& V2V4Real \cite{Xu2023-v2v4real} & 2023 & \cmark & FPV      & RGB   & \cmark & \xmark & \cmark & \cmark & \xmark & \cmark & \xmark &  -  & U/S/H   & \xmark \\
& \cc SCaRL \cite{Ramesh2024-scarl} & \cc 2024 & \cc \xmark & \cc FPV/BEV  & \cc RGB/S & \cc \cmark & \cc \cmark & \cc \cmark & \cc \cmark & \cc \cmark & \cc \cmark & \cc \cmark & \cc D/N & \cc U/S/R/H & \cc \cmark \\
& MARS \cite{Li2024-mars} & 2024 & \cmark & FPV      & RGB   & \cmark & \cmark & \cmark & \cmark & \cmark & \cmark & \cmark & D/N & U/S/H   & \xmark \\
& \cc Scenes101 \cite{Zuern2024-wayvescenes101} & \cc 2024 & \cc \cmark & \cc FPV & \cc RGB   & \cc \xmark & \cc \xmark & \cc \cmark & \cc \xmark & \cc \xmark & \cc \cmark & \cc \cmark & \cc D/N & \cc U/S/R/H & \cc \xmark \\
& TruckScenes \cite{Fent2024-TruckScenes} & 2025 & \cmark & FPV      & RGB   & \cmark & \cmark & \cmark & \cmark & \xmark & \cmark & \cmark & D/N & H/U     & \xmark \\
\bottomrule
\end{tabular}
} % close resizebox
\begin{tablenotes}
    \scriptsize \item[*] \textbf{Impactful:} We define a dataset’s impact by the number of times it was used—not simply cited—by the papers included in our survey.
    Using this criterion, the four most\\impactful papers are associated with the following datasets: nuScenes (52 uses), Waymo Open (19), DRAMA (4), and HighD (3).
    \item[*] \textbf{View} indicates: FPV = First-person View,  BEV = Bird’s-eye View; \rev{\textbf{Image} indicates: RGB = Red, Green, Blue; S = Stereo;} \textbf{Traffic Condition} includes: D/N = Day/Night; \\ U/S/R/H = Urban/Suburban/Rural/Highway;Jam = presence of traffic congestion.
    %While CARLA is also frequently used in this context and employed as a dataset generator, we classify it as a simulator.
\end{tablenotes}
\end{table*}
A typical use of \glspl{fm} for scenario-based testing is to reproduce real-world scenarios in a simulation environment and reconstruct the corresponding events. \Glspl{llm} typically use agents' trajectory data from given datasets, while \glspl{vlm} or \glspl{mllm} can leverage additional input modalities such as LiDAR point clouds, RGB images or video streams, and rich annotations. Specifically, \glspl{dm} use inputs such as RGB images, trajectories, and potentially LiDAR data to generate realistic future scenes or motion patterns through iterative refinement. In contrast, \glspl{wm} aim to learn the underlying dynamics of driving environments by encoding multimodal sensor data (e.g., images, LiDAR, trajectories) and predicting future states or scene evolutions. Meanwhile, for scenario analysis, a common approach is to leverage \glspl{vlm} or \glspl{mllm} to analyze driving scenes, using image or video data, with or without LiDAR or HD maps, across different tasks such as perception, prediction, and reasoning.

To assess the relevance and applicability of datasets, we adopt the categorization scheme introduced by Ding et al.~\cite{ding2023survey}. This scheme enables a structured comparison across datasets, considering their sensor coverage, annotation depth, scene diversity, and potential for controllable generative tasks. In the context of \glspl{fm}, which require large, diverse, and annotated data, the choice of dataset properties is fundamental to enhance the model's generalization potential. We apply this categorization to a selection of impactful and most recent datasets in \autoref{tab:datasets}, using \cite{ding2023survey} to categorize the dataset's properties given below.

%\textbf{Fidelity:} Real-world datasets capture naturalistic driving behavior, complex human interactions, and sensor imperfections. These factors are crucial for grounding foundation models in realistic scenarios and ensuring generalization to real-world deployment.

\textbf{(1) Sensor Data:} High-quality datasets like Waymo~\cite{sun2020waymoopen} and nuScenes~\cite{caesar2020nuscenes}  offer diverse sensor modalities including RGB cameras, LiDAR, and RADAR. Such multimodal input is especially important for pre-training and aligning \glspl{llm}, \glspl{vlm}, \glspl{dm}, and \glspl{wm} across visual and spatial reasoning tasks.

\textbf{(2) Annotation:} These datasets also include detailed 2D and 3D object annotations, lane information, and agent trajectories. This level of semantic and geometric detail supports tasks such as perception, prediction, map-conditioned scenario generation, and safety analysis.

\textbf{(3) Traffic Condition:} Traffic condition describes when and where the data was collected, including time of day (day/night), environment type (urban, suburban, rural, highway), and presence of traffic congestion. These factors affect visibility, traffic flow, road layout, and driving behavior, providing diverse scenarios for evaluating autonomous driving performance.

Datasets such as Waymo Open~\cite{sun2020waymoopen} and nuScenes~\cite{caesar2020nuscenes} are particularly widespread in the literature. This is largely due to their real-world fidelity, rich multisensor coverage, and comprehensive annotations, which make them ideal for training and evaluation of \glspl{fm}. Additionally, it is worth noting that emerging (Visual) \rev{\gls{qa}} datasets relevant to scenario analysis with language \rev{\glspl{fm}} are discussed in Section \rev{IV-\ref{subsec:VLM-based Scenario Analysis}} and Section \rev{V-\ref{subsec:mllm-scenarioanalysis}}.

\subsection{Simulators}
\begin{table*}[ht]
\centering
\caption{Overview of impactful and recent simulators for foundation model-based scenario generation and analysis.}
\label{tab:simulators}
\renewcommand{\arraystretch}{1.2}
\resizebox{1\textwidth}{!}{%
\begin{tabular}{l lccccccccccc}
\toprule
& \multirow{2}{*}{\textbf{Simulator}} & \multirow{2}{*}{\textbf{Year}} & \multirow{2}{*}{\textbf{Backend}} & \multirow{2}{*}{\makecell[c]{\textbf{Open} \\ \textbf{Source}}} & \multirow{2}{*}{\makecell[c]{\textbf{Realistic} \\ \textbf{Perception}}} & \multirow{2}{*}{\makecell[c]{\textbf{Custom} \\ \textbf{Scenario}}} & \multicolumn{2}{c}{\textbf{Map Source}} & \multicolumn{3}{c}{\textbf{API Supports}} & \multirow{2}{*}{\makecell[c]{\textbf{DSL} \\ \textbf{Support}}} \\
\cmidrule(lr){8-9} \cmidrule(lr){10-12} 
& & & & & & &\textbf{Real World} & \textbf{Human Design} & \textbf{Python} & \textbf{C++} & \textbf{ROS 2} \\
\midrule
\multirow{4}{*}{\rotatebox[origin=c]{90}{\textbf{Impactful}}}
% Simulator       &Year  & Back-end          & OS     & RP     & CS     & RWM    & HDM    & Python & C++API & ROS API 
& CARLA \cite{dosovitskiy2017carla}           & 2017 & UE4               & \cmark & \cmark & \cmark & \xmark & \cmark & \cmark & \cmark & \cmark & \cmark \\
& \cc SUMO \cite{Lopez2018-sumo} & \cc 2018 & \cc None & \cc \cmark & \cc \xmark & \cc \cmark & \cc \cmark & \cc \cmark & \cc \cmark & \cc \cmark & \cc \xmark &  \cc \cmark\\
& LGSVL \cite{rong2020lgsvl} & 2020 & Unity & \cmark & \cmark & \cmark & \cmark & \cmark & \cmark & \cmark & \cmark & \cmark\\
& \cc MetaDrive \cite{li2022metadrive} & \cc 2021 & \cc Panda3D & \cc \cmark & \cc \cmark & \cc \cmark & \cc \cmark & \cc \cmark & \cc \cmark & \cc \xmark & \cc \xmark &  \cc \cmark\\

\midrule
% Simulator &Year  & Back-end & Open Source & Realistic Perception & Custom Scenario & Real World Map & Human Design Map & Python API & C++ API & ROS API
\multirow{11}{*}{\rotatebox[origin=c]{90}{\textbf{Most Recent}}}
% & TORCS           & 2000 & None              & \cmark & \cmark & \cmark & \xmark & \xmark & \xmark & \xmark & \xmark \\
% & Webots          & 2004 & ODE               & \cmark & \cmark & \cmark & \cmark & \xmark & \cmark & \cmark & \xmark \\
% & CarRacing       & 2017 & None              & \cmark & \xmark & \xmark & \xmark & \cmark & \xmark & \xmark & \xmark \\
% & SimMobilityST   & 2017 & None              & \cmark & \xmark & \xmark & \xmark & \xmark & \xmark & \xmark & \xmark \\
% & GTA-V           & 2017 & RAGE              & \xmark & \cmark & \xmark & \xmark & \xmark & \xmark & \xmark & \xmark \\
% & highway-env     & 2018 & None              & \cmark & \xmark & \cmark & \xmark & \cmark & \xmark & \xmark & \xmark \\
% & Deepdrive       & 2018 & UE4               & \cmark & \cmark & \cmark & \xmark & \cmark & \cmark & \cmark & \xmark \\
% & esmini          & 2018 & Unity             & \cmark & \xmark & \xmark & \xmark & \xmark & \cmark & \xmark & \xmark \\
% & AutonoViSim     & 2018 & PhysX             & \xmark & \cmark & \cmark & \xmark & \xmark & \cmark & \xmark & \xmark \\
% & AirSim          & 2018 & UE4               & \cmark & \cmark & \cmark & \xmark & \cmark & \cmark & \cmark & \xmark \\
% & Apollo          & 2018 & Unity             & \cmark & \cmark & \cmark & \cmark & \cmark & \cmark & \cmark & \xmark \\
% & Sim4CV          & 2018 & UE4               & \cmark & \cmark & \cmark & \xmark & \cmark & \cmark & \xmark & \xmark \\
& MATLAB AD Toolbox \cite{Matlab} & 2018 & MATLAB            & \xmark & \cmark & \cmark & \cmark & \cmark & \cmark & \cmark & \cmark & \cmark\\
& \cc Nvidia Drive Sim \cite{nvidiaDriveSim} & \cc 2019 & \cc Nvidia Omniverse       & \cc \xmark & \cc \cmark & \cc \cmark & \cc \cmark & \cc  \cmark & \cc \cmark & \cc \cmark & \cc \xmark &  \cc \xmark\\
% & \cc Scenic \cite{Elmaaroufi2024-ScenicNL} & \cc 2019 & \cc None & \cc \cmark & \cc \cmark & \cc \cmark & \cc \cmark & \cc \cmark & \cc \cmark & \cc \xmark & \cc \xmark \\
% & SUMMIT          & 2020 & UE4               & \cmark & \cmark & \cmark & \xmark & \cmark & \cmark & \cmark & \xmark \\
% & MultiCarRacing  & 2020 & None              & \cmark & \xmark & \cmark & \xmark & \cmark & \cmark & \xmark & \xmark \\
% & SMARTS          & 2020 & None              & \cmark & \cmark & \cmark & \cmark & \cmark & \cmark & \xmark & \xmark \\
% & LGSVL           & 2020 & Unity             & \cmark & \cmark & \cmark & \cmark & \cmark & \cmark & \cmark & \cmark \\
% & CausalCity      & 2020 & UE4               & \cmark & \cmark & \cmark & \cmark & \cmark & \cmark & \xmark & \xmark \\
& Vista \cite{Amini2022-Vista} & 2020 & None              & \cmark & \cmark & \cmark & \cmark & \xmark & \cmark & \xmark & \xmark & \xmark\\
% & L2R             & 2021 & UE4               & \cmark & \cmark & \cmark & \cmark & \cmark & \cmark & \cmark & \xmark \\
% & AutoDRIVE       & 2021 & Unity             & \cmark & \cmark & \cmark & \cmark & \cmark & \cmark & \cmark & \cmark \\
& \cc Nuplan \cite{karnchanachari2024nuplan} & \cc 2021 & \cc None              & \cc \cmark & \cc \cmark & \cc \cmark & \cc \cmark & \cc \cmark & \cc \cmark & \cc \xmark & \cc \xmark &  \cc \xmark \\
& AWSIM \cite{awsim2021} & 2021 & Unity             & \cmark & \cmark & \cmark & \cmark & \cmark & \xmark & \xmark & \cmark  & \cmark\\
& \cc InterSim \cite{Sun2022-intersim} & \cc 2022 & \cc None              & \cc \cmark & \cc \cmark & \cc \cmark & \cc \cmark & \cc \xmark & \cc \cmark & \cc \xmark & \cc \xmark & \cc \xmark\\
& Nocturne \cite{Vinitsky2022-nocturne} & 2022 & None              & \cmark & \cmark & \cmark & \cmark & \cmark & \cmark & \cmark & \xmark  & \xmark\\
& \cc BeamNG.tech \cite{beamng2022} & \cc 2022 & \cc Soft-body physics & \cc \xmark & \cc \cmark & \cc \cmark & \cc \xmark & \cc \cmark & \cc \cmark & \cc \xmark & \cc \cmark &  \cc \cmark\\
% & UNISim \cite{Yang2023-unisim} & 2023 & None              & \xmark & \cmark & \cmark & \cmark & \xmark & \xmark & \cmark & \xmark \\
& Waymax \cite{Gulino2023-waymax} &  2023 &  JAX &  \cmark &  \cmark &  \cmark &  \xmark &  \cmark &  \cmark &  \xmark &  \xmark & \xmark\\
& \cc TBSim \cite{Xu2023-tbsim} & \cc 2023 & \cc None              & \cc \cmark & \cc \cmark & \cc \cmark & \cc \cmark & \cc \cmark & \cc \cmark & \cc \xmark & \cc \xmark &  \cc \xmark\\

\bottomrule
\end{tabular}
} % close resizebox
\begin{tablenotes}
    \scriptsize \item[*] \textbf{Impactful:} We identify the impact of a simulator following the same criterion of Table \ref{tab:datasets}, based on the number of times the simulator was used—not simply cited—by\\the papers included in our survey. The most impactful simulators are CARLA (8 uses), MetaDrive (4), LGSVL (3), and SUMO (3).%We also note that CARLA and SUMO are cited additional times in the context of datasets,\\highlighting their dual role as simulators and datasets generators.
\end{tablenotes}
\end{table*}
Simulation platforms are essential in the development and evaluation pipeline of \gls{ad} systems. They enable safe and reproducible testing, large-scale scenario generation, and structured benchmarking. For \rev{\gls{fm}}-based scenario generation, simulators are particularly valuable for generating training data, enabling self-supervised pre-training, and facilitating the sim-to-real validation. 
\gls{fm}-based scenario generation can be performed by \glspl{llm}/\glspl{vlm}/\glspl{mllm} through either API functions or \rev{\glspl{dsl}}, allowing automatic script generation and scenario execution. \autoref{tab:simulators} summarizes the impactful and recent simulation platforms \rev{that} are relevant to scenario generation and analysis. For the classification and evaluation of the existing simulators, we extend the categorization scheme introduced by Ding et al.~\cite{ding2023survey}, focusing on features especially relevant to the development and application of \glspl{fm}. 

\textbf{(1) Backend:} The simulation backend defines the physical and rendering engine used to generate sensor data and simulate interactions. Platforms such as Unreal Engine 4 (UE4) or Unity enable high-fidelity rendering and realistic vehicle dynamics, which are valuable for training perception-driven foundation models. Lightweight or symbolic backends, like SUMO or Nocturne, are useful in large-scale planning and decision-making datasets where rendering realism is less critical.

\textbf{(2) Realistic Perception:} Simulators with realistic perception capabilities provide physics-based sensor outputs, including camera, LiDAR, or radar emulation. Such platforms are crucial for training vision-language \rev{\glspl{fm}}, sensor-fusion backbones, or multimodal \rev{\glspl{wm}}.

\textbf{(3) Custom Scenario:} The ability to define and customize traffic scenarios is a central requirement for both evaluation and data generation workflows. Particularly for \rev{\glspl{fm}}, automated and diverse scenario creation supports the pre-training of models on rare, safety-critical, or systematically varied interactions. Customization typically includes the placement and behavior of traffic participants, route definitions, or modifications of environmental conditions such as weather and lighting. Simulators like CARLA~\cite{dosovitskiy2017carla} offer rich APIs for manual customization, enabling users to script complex multi-agent interactions and adjust parameters such as vehicle behavior, density, and even scene appearance. More recently, platforms like BeamNG.tech~\cite{beamng2022} go a step further by supporting automated scenario generation at scale. This enables the procedural creation and batch testing of varied situations, making it well-suited for training and validating \rev{\glspl{fm}} in closed-loop settings.

\textbf{(4) Map Source:} We differentiate between scenarios based on real-world maps (e.g., OpenStreetMap) and those built from human design. Real-world maps ensure geographic realism and coverage, while human-designed maps enable controlled environments.

\textbf{(5) API-Supports:} API support determines how flexibly simulators can be integrated into training pipelines. Python interfaces are especially useful for data generation and model interaction. \gls{ros2} compatibility allows for testing learned policies in robotics stacks, while C++ APIs provide performance for real-time validation and closed-loop deployment.

\begin{table*}[htp]
\centering
\caption{Overview of foundation model Benchmark Challenges from 2022–2025, categorized by core capabilities.}
\label{tab:foundation-capabilities}
\renewcommand{\arraystretch}{1.2}
\resizebox{0.98\textwidth}{!}{%
%\rowcolors{4}{LightGray}{white}
\begin{tabular}{l l l c c c c c}
\toprule
& \multirow{3}{*}{\textbf{Name}} & \multirow{3}{*}{\textbf{Host}} & \multicolumn{5}{c}{\textbf{Tasks}} \\
\cmidrule(lr){4-8}
& & & \makecell[c]{\textbf{Perception} \\ \textbf{\& Interpretation}} & \makecell[c]{\textbf{Prediction} \\ \textbf{\& Planning}} & \makecell[c]{\textbf{Reasoning} \\ \textbf{\& Decision}} & \makecell[c]{\textbf{Language} \\ \textbf{Understanding}} & \makecell[c]{\textbf{Creative} \\ \textbf{Generation}} \\
\midrule
\multirow{6}{*}{\rotatebox[origin=c]{90}{\textbf{\makecell{Autonomous \\\quad Driving}}}}
& CARLA AD Challenge \cite{CARLALeaderboard} & CARLA   &  &  & &  & \cmark\\
% Name                               & Host                          & Perc   & Pred   & Reason & Lang   & Generation
& \cc DRL4Real \cite{DRLforReal2025} & \cc ICCV  & \cc  & \cc & \cc & \cc & \cc \cmark\\
& Waymo Open Dataset Challenge \cite{waymoChallenge} & Waymo / CVPR WAD              & \cmark & \cmark &        &        & \cmark      \\
%%%
& \cc Argoverse 2: Scenario Mining Competition \cite{davidson2025refav} & \cc ArgoAI & \cc& \cc& \cc\cmark & \cc\cmark & \cc\\
%%%
& Roboflow-20VL \cite{robicheaux2025roboflow100vl} & Roboflow-VL / CVPR  & \cmark & &  &\cmark & \\
%%%
& \cc AVA Challenge \cite{ava-challenge} & \cc AVA Challenge Team & \cc\cmark & \cc\cmark & \cc\cmark & \cc\cmark & \cc\\
%%%

\midrule
\multirow{18}{*}{\rotatebox[origin=c]{90}{\textbf{\makecell{\quad \quad \quad \quad \quad  Other Fields\\ Related to Generation and Analysis}}}} 
& IGLU Challenge \cite{kiseleva2023-iglu} & NeurIPS / IGLU Team           &        & \cmark & \cmark & \cmark &        \\
%%%
& \cc LLM Efficiency Challenge \cite{LLMefficiencycompetition} & \cc NeurIPS                       & \cc       & \cc       & \cc       & \cc\cmark & \cc       \\
& MMWorld \cite{he2024mmworld} & CVPR &   &     & \cmark        &  &        \\
& \cc 3D Scene Understanding \cite{3DSceneUnderstandingCVPR2025} & \cc CVPR & \cc\cmark       & \cc       & \cc       & \cc\cmark & \cc       \\
& Trojan Detection \cite{TrojanDetectionLLM} & NeurIPS / CAIS     &        &        &        & \cmark &        \\
%%%
& \cc SMART-101 \cite{Cherian2023-smart101} & \cc CVPR              & \cc\cmark & \cc       & \cc\cmark & \cc\cmark & \cc       \\
%%%
& NICE Challenge \cite{Kim2024-nice_challenge} & CVPR / LG Research         & \cmark &        & \cmark & \cmark & \\
%%%
& \cc SyntaGen \cite{Singh_2024_syntagen} & \cc CVPR  & \cc\cmark & \cc       & \cc       & \cc       & \cc\cmark \\
%%%
& Habitat Challenge \cite{habitatchallenge2023} & CVPR / FAIR & \cmark & \cmark & \cmark &        &        \\
%%%
& \cc BIG-bench \cite{srivastava2022beyond} & \cc Google Research   & \cc       & \cc       & \cc\cmark & \cc\cmark & \cc       \\
%%%
& BIG-bench Hard (BBH) \cite{BIG-bench-hard} & Google Research &        &        & \cmark & \cmark &        \\
%%%
& \cc HELM \cite{liang2022holistic} & \cc Stanford CRFM & \cc       & \cc       & \cc\cmark & \cc\cmark & \cc       \\
%%%
& MMBench \cite{Liu2024-MMBench} & OpenCompass  & \cmark &        & \cmark & \cmark &        \\
%%%
& \cc MMMU \cite{Yue2024-MMMU} & \cc CVPR / U-Waterloo / OSU  & \cc\cmark & \cc       & \cc\cmark & \cc\cmark  & \cc       \\
%%%
& Open LLM Leaderboard \cite{Myrzakhan2025-Open-LLM-Leaderboard} & VILA-Lab  &        &        &        & \cmark &        \\
%%%
& \cc Text-to-Image Leaderboard \cite{text2imageLeaderboard} & \cc Artificial Analysis  & \cc       & \cc       & \cc       & \cc \cmark & \cc\cmark \\
%%%
& Ego4D \cite{Graumann2022-EGO4D} & FAIR & \cmark & \cmark & \cmark &  \cmark &        \\
%%%
& \cc VizWiz Grand Challenge \cite{Gurari2018-VizWiz} & \cc CVPR VizWiz Workshop          & \cc\cmark & \cc       & \cc       & \cc\cmark & \cc       \\
%%%
& MedFM \cite{medfmChallenge} & NeurIPS / Shanghai AI Laboratory          & \cmark &        &        &        &        \\
\bottomrule
\end{tabular}
}
%\begin{tablenotes}
%    \scriptsize \item[*] \textbf{AD:} Autonomous Driving; \textbf{Otherfields:} Robotics, Computer Vision, Natural Language Processing, etc.
%\end{tablenotes}
\end{table*}

\textbf{(6) Domain-Specific Language (DSL) Support:} Some simulators provide \rev{\gls{dsl}} that enable structured, human-readable scenario specification through high-level functions or syntax. These interfaces are especially useful for integrating \glspl{llm}/\glspl{vlm}/\glspl{mllm} in automated scenario generation pipelines.

Based on these criteria, two simulators stand out in \autoref{tab:simulators} as particularly impactful in \rev{\gls{fm}} research: CARLA \cite{dosovitskiy2017carla} and SUMO \cite{Lopez2018-sumo}. Their complementary capabilities make them well-suited to different aspects of scenario generation and evaluation. SUMO, a microscopic traffic simulator, is designed for large-scale traffic modeling and interaction-heavy scenario simulation at the population level. It supports integration with real-world maps via OpenStreetMap, allowing for geographically accurate traffic flow simulations. These features make it a practical backend for \glspl{llm} tasked with generating or editing traffic configurations using natural language prompts or structured templates. CARLA, in contrast, is a macroscopic simulator with high-fidelity physics, sensor simulation, and photorealistic rendering. It is widely used for ego-agent policy testing in closed-loop environments. Its integration with platforms like Scenic \cite{fremont2019scenic} enables programmatic scenario definition through interpretable formal languages, while its Python API offers fine-grained control over agent behavior, environmental settings, and sensor configurations. These characteristics make CARLA particularly suitable for \glspl{llm}, \glspl{vlm}, and \glspl{mllm} in vision-language understanding, closed-loop control, and multimodal reasoning.

\subsection{Challenges and Benchmarks}

In addition to static datasets and simulation environments, open challenges and benchmarks have become useful tools to evaluate the performance of \glspl{fm}. While datasets provide the raw material for training and offline testing, challenges enable comparative analysis across models in a controlled and competitive setting. To our knowledge, this is the first survey to systematically categorize and compare challenges and benchmarks relevant to scenario generation and analysis. Although many of these challenges originate in other application domains, such as medical imaging, robotics, or general-purpose language understanding, their underlying task structures often align with those found in \gls{ad}. For example, interpreting sensor input, forecasting agent behavior, making multi-step decisions, or generating new representations (e.g., scenes, trajectories, or instructions) are all core operations in scenario understanding.
\autoref{tab:foundation-capabilities} presents a selection of challenges and benchmarks published between 2022 and 2025 while our work features a selective overview. The challenges highlight both direct contributions from autonomous driving, such as the Waymo Open Dataset Challenge~\cite{waymoChallenge}, the Argoverse 2 Scenario Mining Competition~\cite{davidson2025refav}, and the Accessibility Vision and Autonomy (AVA) Challenge~\cite{ava-challenge}, as well as structurally similar benchmarks from other fields. For example, while the Argoverse 2 challenge already touches on scenario analysis, it has not involved scenario generation yet. In contrast, tasks such as \rev{\gls{vqa}}, egocentric video understanding, or synthetic image generation often require models to interpret complex scenes and produce new, coherent outputs, an ability that is equally fundamental for scenario generation. Challenges like SyntaGen~\cite{Singh_2024_syntagen} and the Text-to-Image Leaderboard~\cite{text2imageLeaderboard} illustrate this parallel particularly well: models are asked to generate synthetic examples that exhibit structural realism and diversity. Each challenge is categorized along five core capabilities: %these categories reflect not only different input-output modalities but also increasing levels of abstraction, generalization, and task complexity. %While not all benchmarks are specific to autonomous driving, they help define a broader functional space in which scenario generation and analysis are embedded and in which foundation models are expected to operate effectively.

\textbf{(1) Perception \& Interpretation:} This category refers to the model's ability to process sensor inputs and extract meaningful semantic representations. Benchmarks such as MMBench~\cite{Liu2024-MMBench} and MMMU~\cite{Yue2024-MMMU} require fine-grained visual understanding across diagrams, images, and structured visual data. The MedFM~\cite{medfmChallenge} challenge focuses on extracting clinically relevant patterns from medical images such as X-rays and histology slides. Ego4D~\cite{Graumann2022-EGO4D} evaluates perception in the context of egocentric video, where models must interpret long, unstructured streams of first-person footage.

\textbf{(2) Prediction \& Planning:} Challenges in this category require models to forecast future events or plan a sequence of actions based on partial observations. The Waymo Open Dataset Challenge~\cite{waymoChallenge} is a prominent example, assessing motion forecasting from multi-agent sensor streams in real-world traffic scenarios. In the Habitat challenge~\cite{habitatchallenge2023}, embodied agents must navigate photo-realistic indoor environments toward semantic or visual goals.

\textbf{(3) Reasoning \& Decision Making:} This capability includes commonsense reasoning, causal inference, and multi-hop planning. The BIG-bench~\cite{srivastava2022beyond} and BIG-bench Hard (BBH)~\cite{BIG-bench-hard} benchmarks target difficult problems in logic, mathematics, and abstract reasoning, many of which remain unsolved even by large models. SMART-101~\cite{Cherian2023-smart101} evaluates reasoning in dialogue, specifically whether models can generate helpful, honest, and harmless responses.

\textbf{(4) Language Understanding \& Generation:} This encompasses tasks such as instruction following, \gls{qa}, summarization, and dialogue generation. The LLM Efficiency Challenge~\cite{LLMefficiencycompetition} evaluates how well \glspl{fm} can be fine-tuned under strict computational budgets. HELM~\cite{liang2022holistic} offers a multi-dimensional evaluation across more than a dozen application domains, measuring not only task performance but also fairness, bias, and calibration. The Open LLM Leaderboard~\cite{Myrzakhan2025-Open-LLM-Leaderboard} provides a public ranking of open-source language models based on standardized evaluations across tasks such as \gls{qa} \rev{or} summarization.

\textbf{(5) Creative Generation:} Finally, this category captures the ability of a model to generate complex artifacts such as images, captions, or synthetic data samples. The Text-to-Image Leaderboard~\cite{text2imageLeaderboard} evaluates diffusion-based generative models using human preference judgments over image outputs. SyntaGen~\cite{Singh_2024_syntagen} tests whether \rev{\glspl{dm}} can generate synthetic images that preserve sufficient structure and diversity to train robust perception models.

Overall, these benchmarks provide a structured landscape for measuring and comparing the capabilities of \rev{\glspl{fm}} beyond narrow task-specific metrics. They reflect the growing demand for models that are not only accurate but also general, adaptable, and \rev{robust across domains. For instance, the Ego4D~\cite{Graumann2022-EGO4D} benchmark requires models to understand egocentric video data across diverse daily contexts such as households, workplaces, and outdoor activities. In contrast, MedFM~\cite{medfmChallenge} evaluates the ability to analyze complex medical images, requiring high precision and domain-specific knowledge. Despite their differing domains, both tasks rely on similar underlying capabilities, illustrating the versatility required from \glspl{fm}.}

\section{Open Research Questions and Challenges}
\label{sec:discussion}

In this paper, we \rev{illustrate how} the state of the art in the emerging field of scenario generation and analysis with \glspl{fm} is quite extensive. Nevertheless, there are still some open research questions and challenges. Here, we present a list of open challenges based on additional discussions with leading researchers and experts in the field. These challenges open new research questions to use \glspl{fm} for scenario generation and analysis in \gls{ad}.

\textbf{Challenge 1 -- Balancing Plausibility and Edge Case Generation:} %Realism vs rare events
Effective scenario generation requires balancing realism with the ability to capture rare edge cases. Realistic scenarios demand that \glspl{fm} abstractly understand the real-world dynamics~\cite{gao2025nurisk}. On the other hand, edge cases essential for safety assurance~\cite{brusnicki2025savant} often approach the boundary of perceived plausibility, making them challenging for \glspl{fm} to generate without producing unrealistic outcomes. When the plausibility of the generated scenarios is compromised, the resulting scenarios cannot support safety assurance arguments~\cite{nalic2020scenario}. Thus, the key challenge is ensuring the realism of the generated scenarios, while enabling \glspl{fm} to generalize and capture critical edge-case situations.

\rev{\textbf{Challenge 2 -- Large-Scale Multimodal Data Availability:}
Many \glspl{fm} are trained on existing datasets that struggle to capture the full diversity of real-world driving scenarios. Moreover, the integration of multimodal data such as LiDAR, camera, RADAR, and text remains limited compared to single-modality \glspl{fm}. \revB{This is due to the lack of open-source LiDAR and RADAR data at publicly accessible, internet-scale volumes (comparable to those used to pretrain \glspl{fm} on web data such as news, books, and large-scale image and video collections), and to the limited size of domain-specific multimodal datasets~\cite{cui2024survey}.}
In addition, open datasets that include rare, diverse, and safety-critical events are still scarce. Thus, a major challenge is the limited availability of diverse, unbiased multimodal data needed to enable scenario generation with high realism and fidelity.}

\rev{\textbf{Challenge 3 -- Standardized Evaluation Metrics and Benchmarks for Scenario Generation:}
Currently, there is no established standard for the automated evaluation and generation of driving scenarios. Widely accepted metrics to assess realism, plausibility, dynamic feasibility, controllability, and safety-criticality are still missing, hindering fair and meaningful comparisons among different methods. To fill this gap, open-source evaluation frameworks and community challenges or leaderboards are needed, requiring participants to generate and assess autonomous driving scenarios. Such initiatives would enable consistent benchmarking, foster the development of multi-dimensional evaluation metrics, and promote reproducible research practices. This will ultimately accelerate the integration of scenario-generation methods into safety assessment pipelines.}

\textbf{Challenge 4 -- Safety, Robustness \& Verification:}
Most existing methods lack formal guarantees for safety, correctness, or scenario coverage. The stochastic nature of \glspl{fm} increases the risk of hallucinated outputs, limiting their reliability for \gls{ad} safety assurance. A key challenge is ensuring that the generated scenarios are logically grounded and validated through formal verification, constraint satisfaction, or logic-based safety rules rather than merely correlated with the intended context.

\rev{\textbf{Challenge 5 -- Computational Cost and Scalability:}}
Current \gls{fm}-based generation methods demand substantial computational resources, with training requiring massive datasets, long runtimes, and high-performance hardware. Even inference and model fine-tuning are costly without advanced infrastructure. This raises unsolved challenges in scalability, accessibility, and cost-effectiveness, particularly for smaller organizations or resource-constrained applications.

%The current \gls{fm}-based generation methods require extensive computational resources, including high-performance hardware and significant energy consumption. If we leave the realm of pre-trained models, the initial training will require massive datasets and computing clusters that are running for weeks or months. Both model updates and running inference are computationally intensive for organizations without access to advanced hardware. We see challenges in terms of scalability, accessibility, and cost-effectiveness, especially for smaller organizations or applications with limited computational budgets.

\textbf{Challenge 6 --  Industrial Transferability and Validation:}
While academia offers many methods for virtual testing and evaluation, the industry must ultimately adapt them for real-world \gls{ad} applications. Bridging this gap requires method validation, standardization~\cite{Song2024}, and seamless integration into existing workflows. Thus, a key research question lies in developing approaches that are not only theoretically sound but also practical, efficient, and accessible to diverse stakeholders, backed by robust industrial validation demonstrating clear benefits and adaptability.
%Although many methods and approaches for virtual testing and evaluation are provided by the academic community, ultimately, the industry needs to transfer these techniques for their real-world AV application. Bridging the gap between academic innovation and industrial application involves significant efforts in method validation and standardization~\cite{Song2024}, as well as integration into existing industrial processes. We see a challenge in developing methodologies that are not only theoretically sound but also practically viable, efficiently implementable, and understandable to a wide range of users and stakeholders. Moreover, it requires demonstrating tangible benefits, robustness, and adaptability through extensive industrial validation studies.

\section{Future Directions}
\label{sec:directions}

Addressing the above-mentioned challenges in scenario generation and analysis using \glspl{fm} yields several directions for future improvement and new research agendas.

\textbf{Research Direction 1 -- Improve Realism:} 
Improving the realism and plausibility of the generated scenarios will require integrating domain-specific knowledge into \glspl{fm}, enhancing their understanding of real-world dynamics and interactions. Hybrid approaches that combine physics-based models with data-driven \glspl{fm} offer promise in generating physically coherent scenarios. Also, the exploration of dreaming with \glspl{wm} \cite{ha2018world} can address gaps in sensor simulation: the data-driven nature of dreaming can capture fine-grained sensor characteristics with high fidelity.

\textbf{Research Direction 2 -- Create Rare Events:} 
Capturing rare, high-risk events requires dedicated methods to systematically identify and generate such scenarios. We recommend creating targeted datasets that focus on infrequent but critical situations to improve the accuracy of models in such cases. Additionally, incorporating reasoning techniques such as causal or counterfactual reasoning~\cite{mondorf2024beyond}, which may help \glspl{fm} deduce plausible yet uncommon scenarios.
%Representing rare events requires dedicated methods to systematically identify and generate scenarios involving rare high-risk events. We recommend to create targeted datasets that emphasize infrequent scenarios. Only those will be able to strengthen the model's ability to generate relevant, high-risk yet uncommon real-world scenarios. Another direction could be incorporating reasoning approaches like causal reasoning or counterfactual reasoning. Those methodes have the possibility to enable the \glspl{fm} to identify plausible but rare scenarios by deduction.

\textbf{Research Direction 3 -- Create Multimodal Datasets:} 
Multimodal data integration remains a major challenge, requiring large-scale datasets specifically designed for scenario generation. These should combine vehicle sensor data, such as LiDAR, RADAR, and cameras, with map data, traffic rules, control actions, human feedback, and textual annotations. We also recommend developing new model architectures and training methods specifically tailored to multimodal fusion, in order to address the current limitations in scalability and integration.

\textbf{Research Direction 4 -- Develop Metrics and KPIs for Comparison:} We heavily recommend the development of standardized evaluation methods for an objective comparison of scenarios and scenario generation approaches. This requires new benchmarks and metrics for realism, controllability, diversity, and safety-criticality, along with broad adoption by the community. Promoting these new benchmarks in competitions at the major conferences will drive progress, standardization, and community-driven innovation.

\textbf{Research Direction 5 -- Reduce Computational Demands:} Computational efficiency and scalability present major practical constraints. 
Addressing them requires further investigation of techniques such as model distillation, pruning, and quantization, specifically tailored to scenario generation and analysis tasks, to minimize computational demands without sacrificing performance. 

\rev{\textbf{Research Direction 6 -- FMs as Safe Data Flywheels:} 
A key research direction concerns the integration of \glspl{fm} into AV safety validation workflows. This includes using \glspl{fm} as safe data flywheels, where the generated scenarios continuously support testing, AV models retraining, safety assessment, and performance monitoring. Future work should ensure scenario representativeness, balance real and synthetic data, and develop robust metrics to quantify the safety impact of the generated edge cases across the AV lifecycle.}

\rev{\textbf{Research Direction 7 -- Regulatory Compliance:} 
Ethical considerations and regulatory compliance must be integral to future developments. Transparent methodologies are needed to identify, mitigate, and validate biases in the generation and analysis of \gls{ad} scenarios. Equally important are robust approaches to data privacy management, to ensure compliance with legal and ethical standards while safeguarding sensitive training data. Advancing these aspects will also support the use of generated scenarios in safety validation and certification, contributing to structured safety arguments.}

\section{Conclusion}
\label{sec:conclusion}

This survey examines the state-of-the-art in \glspl{fm} for autonomous driving applications, emphasizing their significant contributions to both scenario generation and scenario analysis. 
\glspl{fm}, including \glspl{llm}, \glspl{vlm}, \glspl{mllm}, \glspl{dm}, and \glspl{wm}, have emerged as promising tools to enhance the realism, diversity, and scalability of scenario-based testing in \gls{ad}.

The versatility of \glspl{fm} lies in their ability to learn from large-scale, heterogeneous datasets through self-supervised training. Their capability to generalize knowledge across various tasks has advanced the scenario-based testing paradigm, overcoming many limitations of traditional rule-based and data-driven methods. Particularly, the dual capability of scenario generation and scenario analysis presented by \glspl{fm} positions them as crucial enablers for robust and efficient validation frameworks in \gls{ad} systems.

Despite these advances, notable challenges persist. \rev{Achieving fine-grained controllability in safety-critical scenarios and ensuring robust realism in generated scenes are ongoing research hurdles. Computational efficiency remains a significant challenge, as many foundation models demand high memory bandwidth, high inference times, and costly GPU resources, limiting their practicality for large-scale scenario generation and real-time testing. } Additionally, while the surveyed models demonstrate promising results, further research is needed to enhance the interpretability of their outputs, improve alignment with real-world traffic conditions, and systematically address out-of-distribution scenarios. \rev{Future work should also investigate if and how improvements in \gls{fm} model designs and size may result in better generalization for scenario generation and analysis.}

%Moving forward, research should prioritize the integration of multiple modalities and the development of hybrid approaches that combine human domain expertise with the adaptive intelligence of \glspl{fm}. 
%Furthermore, efforts must focus on creating extensive benchmarks and datasets tailored explicitly to scenario generation and analysis, facilitating fair, comprehensive, and reproducible evaluations.
Ultimately, as autonomous vehicles approach broader operational domains and higher levels of automation, the role of advanced scenario generation and analysis methods will be paramount. \glspl{fm} present a powerful framework for this evolution, promising to revolutionize both the safety and efficiency of \gls{ad} development. The future trajectory of this research is expected to bring further transformative advancements, fostering safer, more reliable, and broadly accessible autonomous mobility.
\section*{ACKNOWLEDGMENT}
\rev{
\textbf{Leadership \& Structure:} Y. Gao (lead), M. Piccinini, J. Betz. 
\textbf{Technical Sections:} Y. Gao (LLMs/VLMs/MLLMs), Y. Zhang (DMs), D. Wang and M. Piccinini (WMs), K. Moller (Datasets, Simulators, and Benchmarks). 
\textbf{Research Curation:} R. Brusnicki (collection of papers, GitHub repository). 
\textbf{Oversight \& Strategy:} J. Betz, M. Piccinini, A. Gambi, K. Storms, J. F. Totz, B. Zarrouki, S. Peters, A. Stocco, B. Alrifaee, and M. Pavone contributed to critical revisions and formulating research directions.}

\bibliographystyle{IEEEtran_fixed6}
\bibliography{bibliography.bib}
%\bibliography{staging_bibliography.bib}

% Yuan
\begin{IEEEbiography}[{\includegraphics[width=1in,height=1.25in, clip,keepaspectratio]{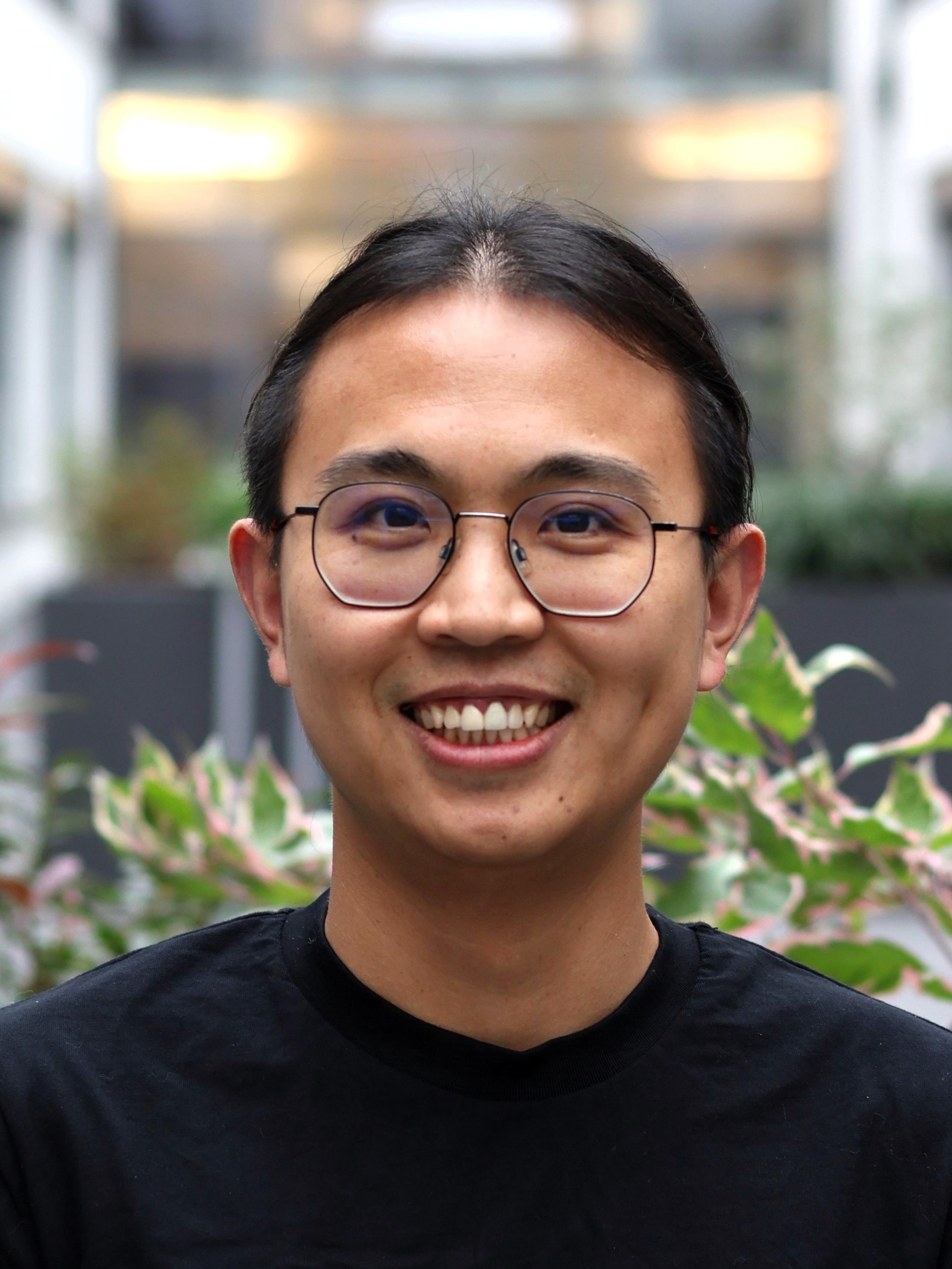}}]
{Yuan Gao}~is a Ph.D. student at the Autonomous Vehicle Systems (AVS) lab at the Technical University of Munich (TUM). He received a B.Sc. (with high distinction) in Mechanical Engineering from Hefei University of Technology in 2017 and two M.Sc. degrees from TUM: Mechatronics and Robotics (with high distinction) and Development, Production, and Management in Mechanical Engineering (with distinction) in 2023.
His research focuses on scenario generation and analysis for autonomous vehicles using FMs.
\end{IEEEbiography}

\vskip -1\baselineskip
% Mattia
\begin{IEEEbiography}[{\includegraphics[width=1in,height=1.1in, clip,keepaspectratio]{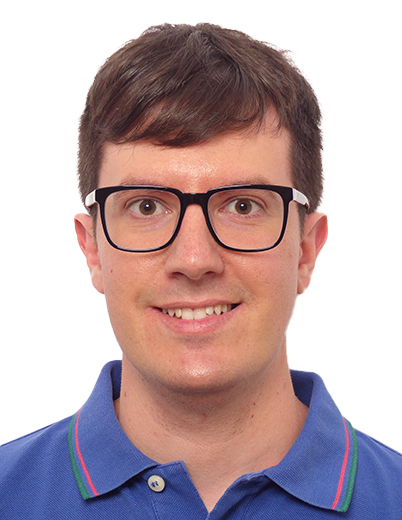}}]
{Mattia Piccinini}~is a Humboldt Post-doctoral Fellow at the Autonomous Vehicle Systems (AVS) lab at the Technical University of Munich (TUM). He received an M.Sc. in mechatronics engineering (cum laude) and a Ph.D. (cum laude) in autonomous systems from the University of Trento, Italy, in 2019 and 2024 respectively. He obtained the TUM Global Post-doctoral Fellowship (2024), the ITSS Best Dissertation Award (2025) and the Humboldt Post-doctoral Fellowship (2025). He was a visiting researcher at the Universit{\"a}t der Bundeswehr Munich and at the Eindhoven University of Technology. He serves as an Associate Editor for the IEEE IROS conference. His research focuses on physics-guided motion generation and control of mobile ground robots.
\end{IEEEbiography}

\vskip -1\baselineskip
%Yuchen
\begin{IEEEbiography}[{\includegraphics[width=1in,height=1.1in, clip,keepaspectratio]{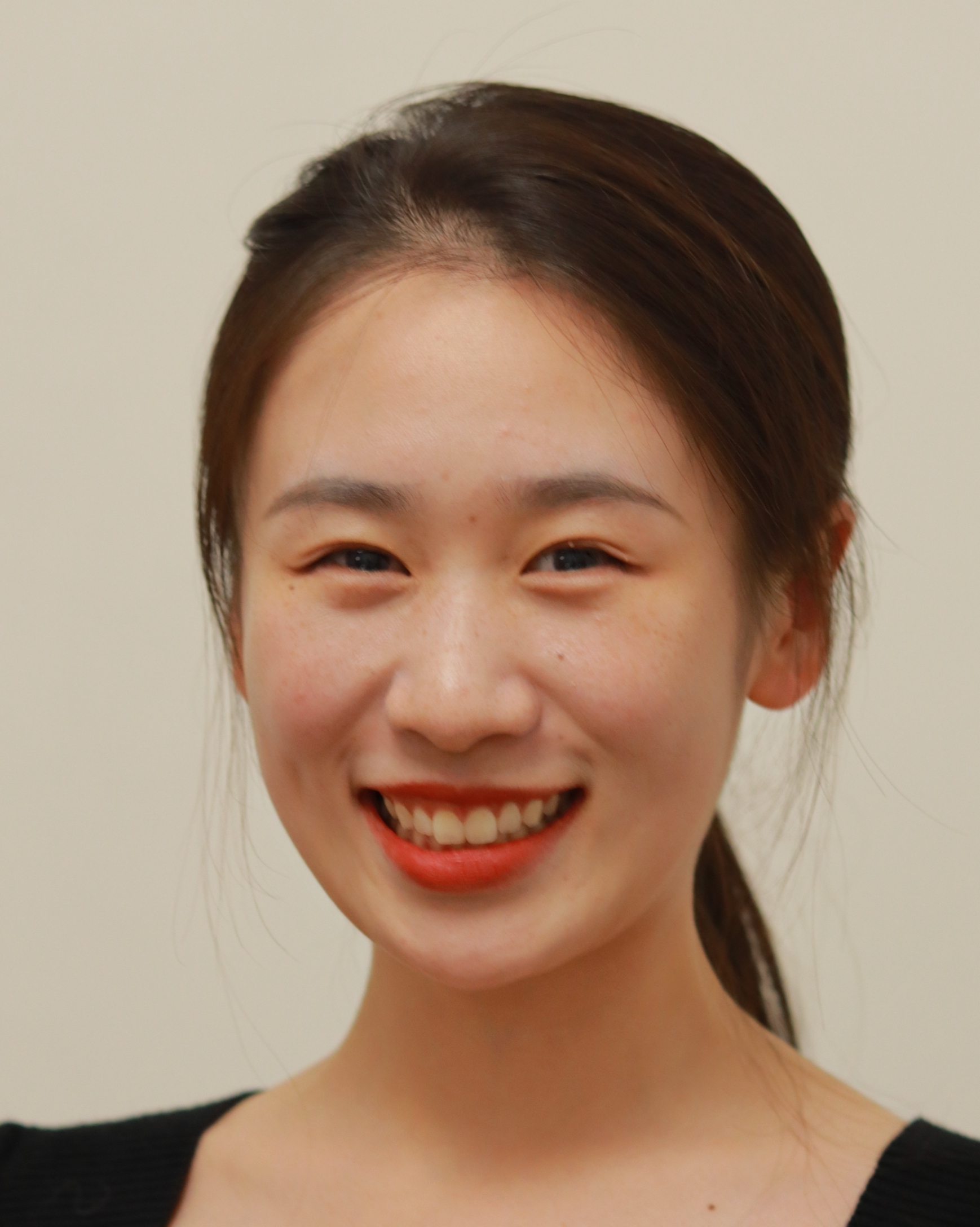}}]
{Yuchen Zhang}~is a Ph.D. student at the Autonomous Vehicle Systems (AVS) lab at the Technical University of Munich (TUM). She holds a bachelor’s degree in Mechanical Engineering from Shanghai University (2021) and a master’s degree in Robotic Systems Engineering from RWTH Aachen University (2023). Since April 2024, she has been part of the AVS Lab, where her research focuses on perception systems for off-road vehicles.
\end{IEEEbiography}
%\vspace{-0.1cm}

\vskip -1\baselineskip % plus -1fil
\begin{IEEEbiography}[{\includegraphics[width=1in,height=1.1in, clip,keepaspectratio]{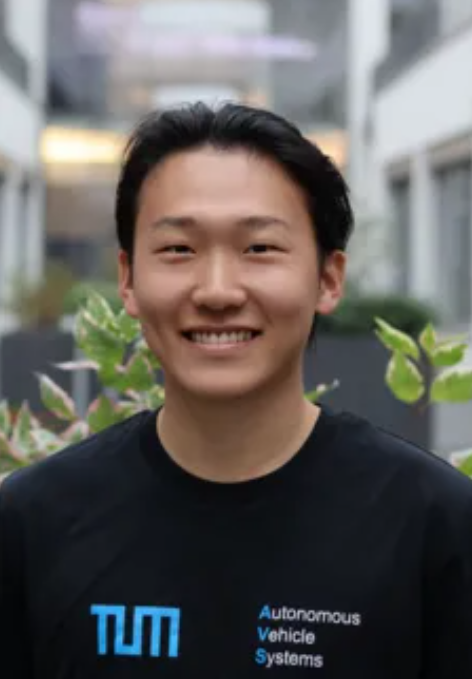}}]
{Dingrui Wang}~is a PhD student at the Autonomous Vehicle Systems (AVS) Lab at the Technical University of Munich (TUM). He received his B.Sc. from Tianjin University in 2021 and M.Sc. from KU Leuven in 2022. His research focuses on investigating the applications of world models and end-to-end learning to develop reliable, data-driven systems capable of handling the complex decision-making processes required for autonomous navigation.
\end{IEEEbiography}

%\vskip -1\baselineskip

\begin{IEEEbiography}[{\includegraphics[width=1in,height=1.25in, clip,keepaspectratio]{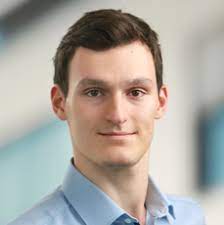}}]
{Korbinian Moller}~received a B.Sc. degree and an M.Sc. degree in mechanical engineering from the Technical University of Munich (TUM) in 2021 and 2023, respectively. He is currently pursuing a Ph.D. degree at the Autonomous Vehicle Systems (AVS) lab at TUM. His research interests include edge-case scenario simulation, the optimization of vehicle behavior, and motion planning in autonomous driving.
\end{IEEEbiography}
%\vspace{-0.1cm}

\vskip -1\baselineskip % plus -1fil
\begin{IEEEbiography}[{\includegraphics[width=1in,height=1.25in, clip,keepaspectratio]{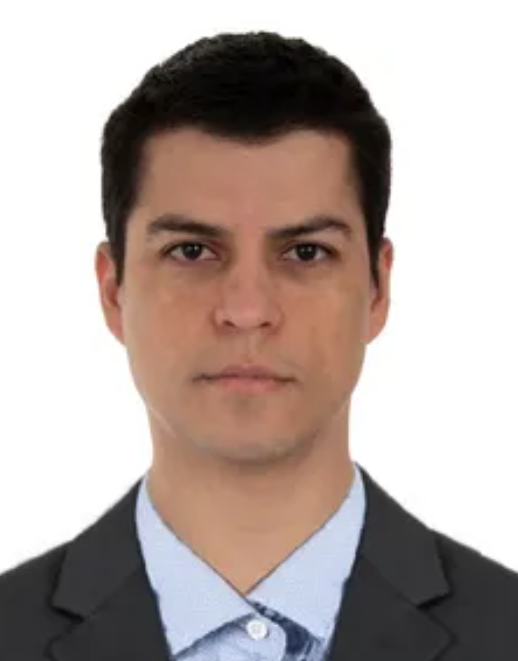}}]
{Roberto Brusnicki}~is a PhD student at the Autonomous Vehicle Systems (AVS) lab at the Technical University of Munich. He received his B.Sc. in 2017 and M.Sc. in 2022 from the Aeronautics Institute of Technology. His research focuses on enhancing autonomous vehicle performance using large language models for perceptual accuracy, scene understanding, and decision-making in ambiguous scenarios. 
%He also explores their use in planning and behavior prediction to improve trajectory optimization and motion planning.
\end{IEEEbiography}
%\vspace{-0.1cm}

\vskip -1\baselineskip % plus -1fil
\begin{IEEEbiography}[{\includegraphics[width=1in,height=1.25in,clip,keepaspectratio]{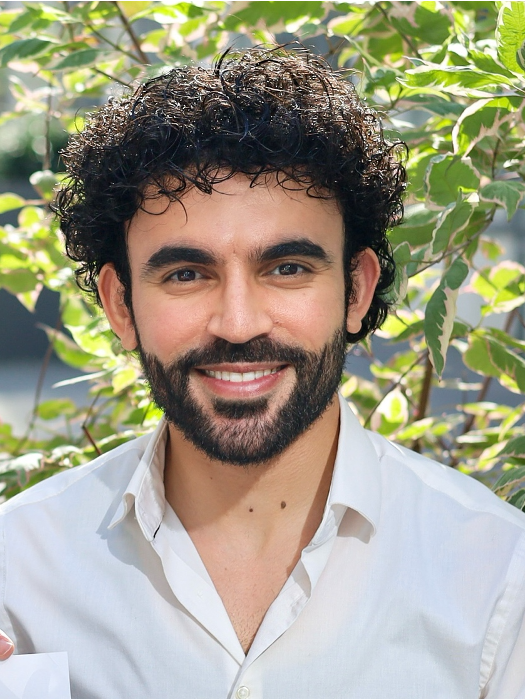}}]
{Baha Zarrouki}~is a PhD Researcher at the Autonomous Vehicle Systems (AVS) Lab at the Technical University of Munich (TUM). He received a B.Sc. and an M.Sc. in Electrical Engineering from TU Berlin in 2018 and 2020, respectively. His PhD research focused on fusing Deep Reinforcement Learning and Model Predictive Control for Nonlinear Motion Control of AV Systems. His current work explores World Model Predictive Control for autonomous driving.
\end{IEEEbiography}
\vskip -1\baselineskip % plus -1fil
%\vspace{-0.1cm}

\begin{IEEEbiography}[{\includegraphics[width=1in,height=1.25in,clip,keepaspectratio]{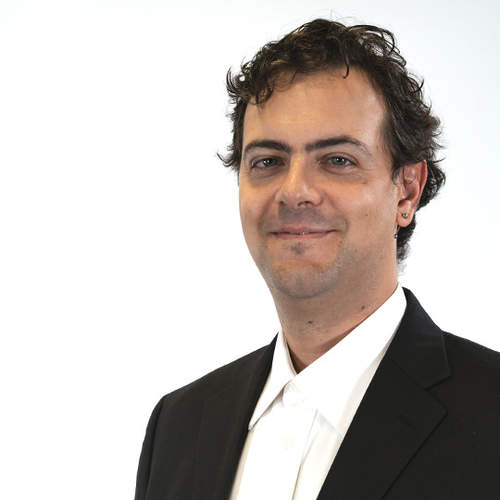}}]
{Alessio Gambi}~is a Scientist at the DSS center of the Austrian Institute of Technology. His research focuses on automated software testing of autonomous systems. He was funded by the European Union. Views and opinions expressed are however those of the authors only and do not necessarily reflect those of the European Union or the European Health and Digital Executive Agency (HADEA). Neither the European Union nor the granting authority can be held responsible for them. RobustifAI project, ID 101212818.
% He is a recipient of a Facebook Testing and Verification Award (2019) and the Best Paper Award of the International Conference on Web Engineering (ICWE 2010).
%
He serves on the program committees of flagship software engineering conferences (e.g., ASE, FSE, ISSTA, ICST) and reviews for top-tier journals (e.g., TSE, TOSEM, TAAS). 

\end{IEEEbiography}
%\vspace{-0.1cm}
\vskip -1\baselineskip % plus -1fil

\begin{IEEEbiography}[{\includegraphics[width=1in,height=1.25in,clip,keepaspectratio]{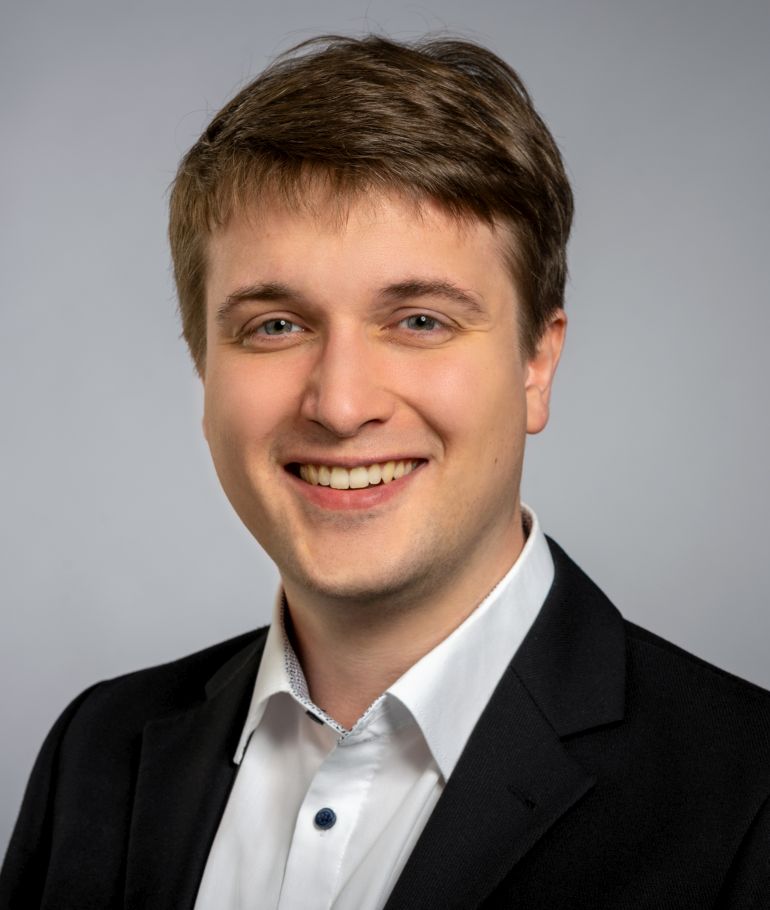}}]
{Jan Frederik Totz}~completed his PhD (Dr. rer. nat.) in Theoretical Physics at the Technical University of Berlin in 2017 on the topic of neuromorphic spiking neural networks. He received the Springer Thesis Award, the Chorafas prize, and the Carl-Ramsauer Award for his thesis. He continued with a postdoctoral position in the Departments of Mathematics and Mechanical Engineering at the Massachusetts Institute of Technology, funded by a Feodor Lynen Research Fellowship of the Humboldt Foundation. Currently, he serves as an R\&D Engineer at Audi, specializing in Autonomous Driving.
\end{IEEEbiography}
%\vspace{-0.1cm}
\vskip -1\baselineskip % plus -1fil

\begin{IEEEbiography}[{\includegraphics[width=1in,height=1.25in,clip,keepaspectratio]{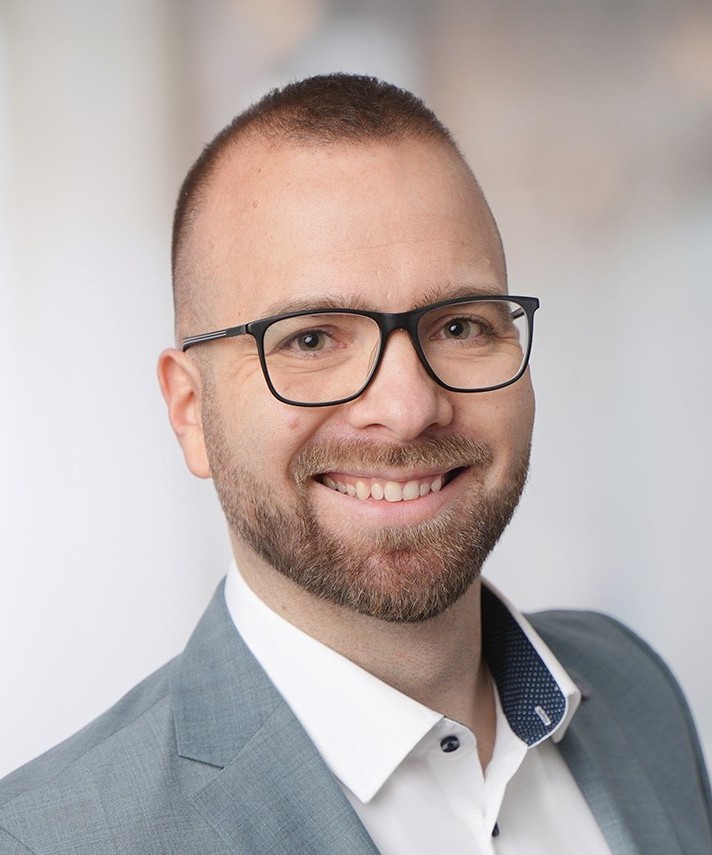}}]
{Kai Storms}~studied Mechanical and Process Engineering at the Technical University of Darmstadt, where he received the M.Sc. in 2020 and completed his PhD (Dr.-Ing.) degree in February 2024. 
He is currently the chief engineer and vice head of the Institute of Automotive Engineering at the Technical University of Darmstadt, Germany. 
His topic was context-aware data reduction for highly automated driving.
His research interests include the verification and validation of automated vehicles.
\end{IEEEbiography}
%\vspace{-0.1cm}
\vskip -1\baselineskip % plus -1fil

\begin{IEEEbiography}[{\includegraphics[width=1in,height=1.25in,clip,keepaspectratio]{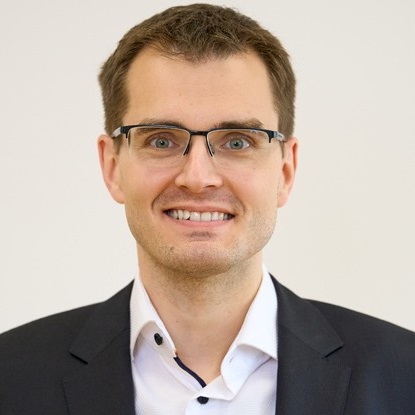}}]
{Steven Peters}~was born in 1987 and received his PhD (Dr.-Ing.) in 2013, at Karlsruhe Institute of Technology, Karlsruhe, Baden-Württemberg, Germany.
From 2016 to 2022 he worked as Manager of Artificial Intelligence Research at Mercedes-Benz AG in Germany. 
He is a Full Professor at the Technical University of Darmstadt, Darmstadt, Germany and heads the Institute of Automotive Engineering in the Department of Mechanical Engineering since 2022.
\end{IEEEbiography}
%\vspace{-0.1cm}
\vskip -1\baselineskip % plus -1fil

\begin{IEEEbiography}[{\includegraphics[width=1in,height=1.25in,clip,keepaspectratio]{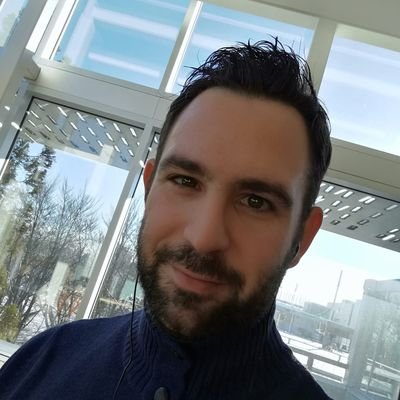}}]
{Andrea Stocco}~is an Assistant Professor at the Technical University of Munich at the Chair of Software Engineering for Data-intensive Applications. He is also the head of the Automated Software Testing unit at fortiss. His research focuses on the interface between software engineering and deep learning with the goals of improving the robustness, reliability, and dependability of data-intensive software systems. 
%He is the recipient of several awards, including two Distinguished Paper Awards at the 18th International Conference on Software Testing, Verification and Validation (ICST 2025), the Best Paper Award at the 16th International Conference on the Quality of Information and Communications Technology (QUATIC 2023) and the Best Student Paper Award at the 16th International Conference on Web Engineering (ICWE 2016). 
He has received multiple best and distinguished paper awards at leading international conferences, including ICST (2025), QUATIC (2023), and ICWE (2016).
He serves on the program committees of top-tier software engineering conferences (e.g., ICSE, FSE, ISSTA, ICST) and reviews for software engineering journals (e.g., TSE, TOSEM, EMSE, JSS, IST).
\end{IEEEbiography}
%\vspace{-0.1cm}
\vskip -1\baselineskip % plus -1fil

\begin{IEEEbiography}[{\includegraphics[width=1in,height=1.25in,clip,keepaspectratio]{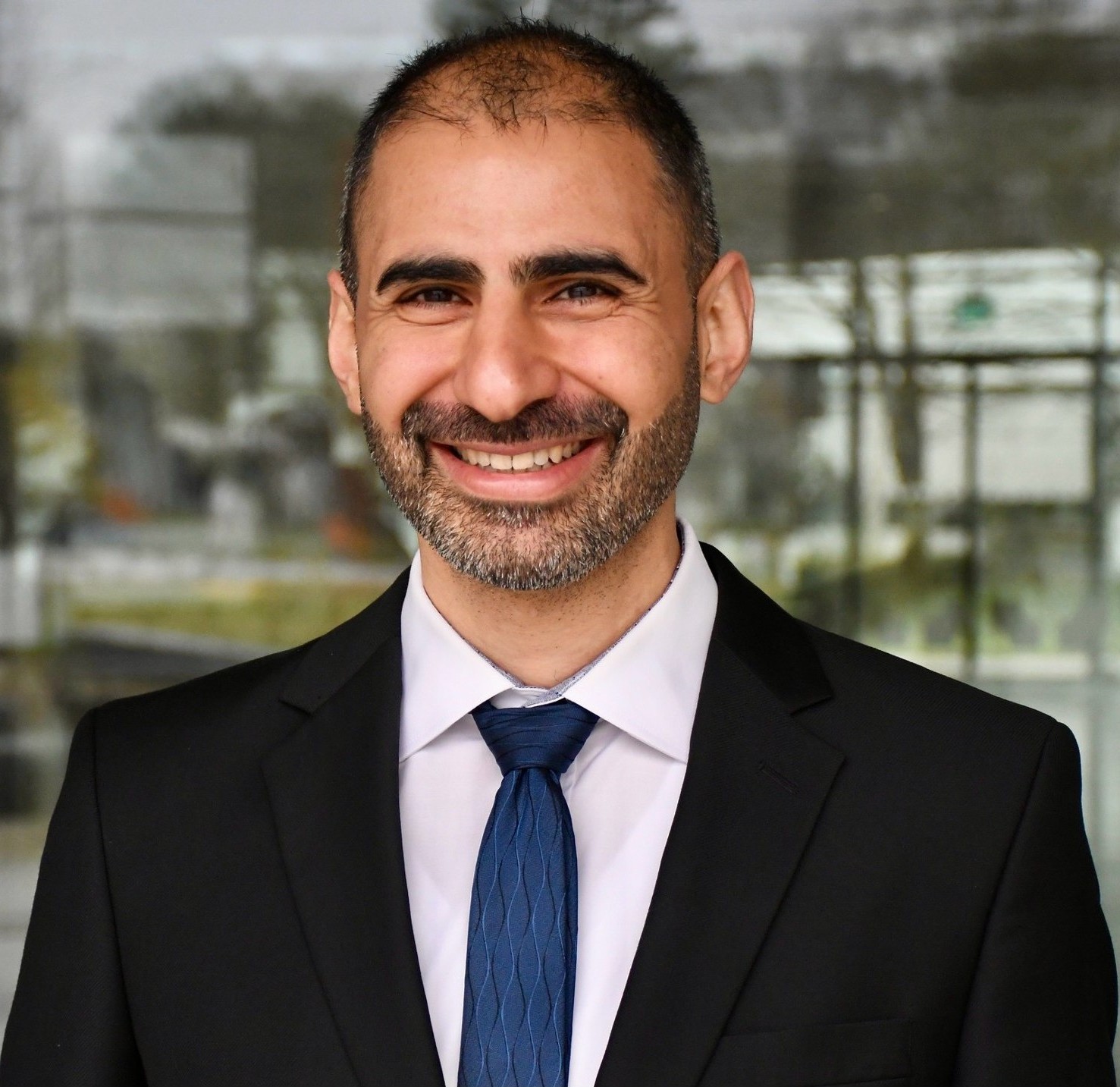}}]
{Bassam Alrifaee}~is a professor at the University of the Bundeswehr Munich, directs the Professorship for Adaptive Behavior of Autonomous Vehicles. Formerly at RWTH Aachen University, he founded the Cyber-Physical Mobility (CPM) group and the CPM Lab (2017–2024). He held a Visiting Scholar role at the University of Delaware in 2023. His research focuses on distributed control, service-oriented architectures, and connected and automated vehicles. Prof. Alrifaee secured grants and received awards for his advisory and editorial work. He holds Senior Member status at IEEE.
\end{IEEEbiography}
%\vspace{-0.1cm}
\vskip -1\baselineskip % plus -1fil

\begin{IEEEbiography}[{\includegraphics[width=1in,height=1.25in,clip,keepaspectratio]{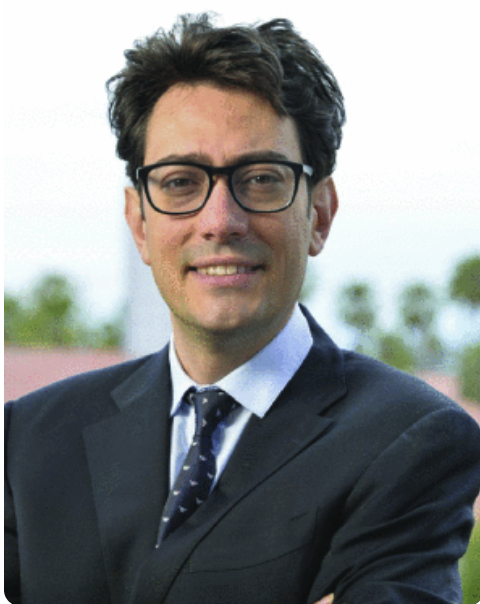}}]
{Marco Pavone}~is an Associate Professor of Aeronautics and Astronautics at Stanford University, where he directs the Autonomous Systems Laboratory and the Center for Automotive Research at Stanford. He is also a Distinguished Research Scientist at NVIDIA where he leads autonomous vehicle research. Before joining Stanford, he was a Research Technologist within the Robotics Section at the NASA Jet Propulsion Laboratory. He received a Ph.D. degree in Aeronautics and Astronautics from the Massachusetts Institute of Technology in 2010. His main research interests are in the development of methodologies for the analysis, design, and control of autonomous systems, with an emphasis on self-driving cars, autonomous aerospace vehicles, and future mobility systems. He is a recipient of a number of awards, including a Presidential Early Career Award for Scientists and Engineers from President Barack Obama.
\end{IEEEbiography}

\begin{IEEEbiography}[{\includegraphics[width=1in,height=1.25in,clip,keepaspectratio]{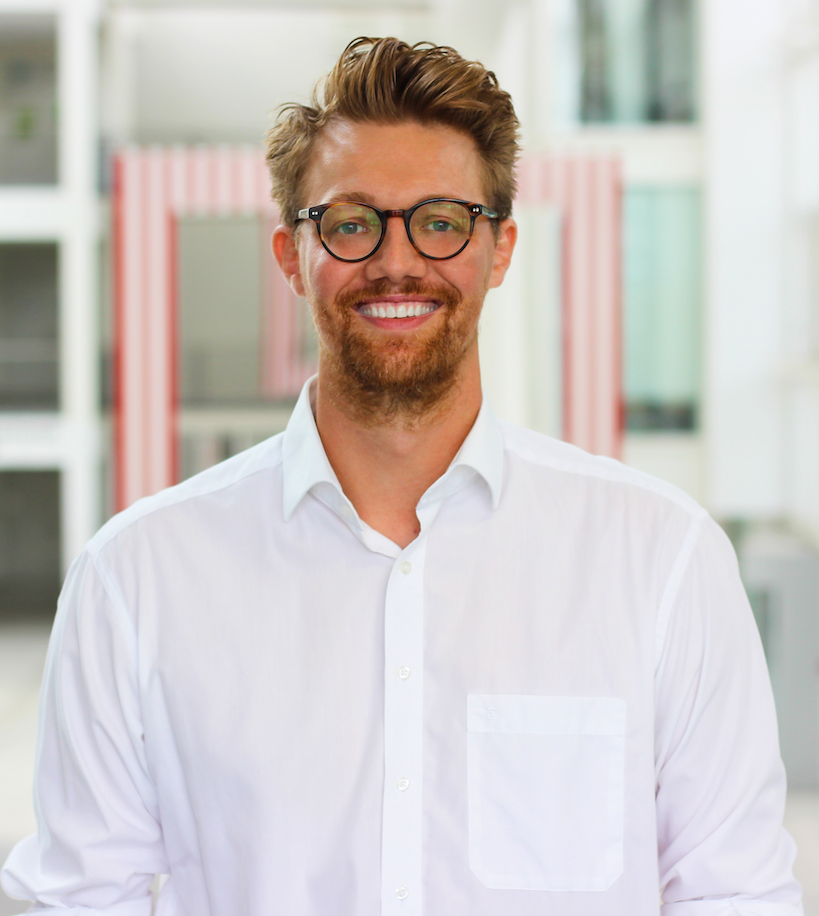}}]
{Johannes Betz}~is an assistant professor in the Department of Mobility Systems Engineering at the Technical University of Munich (TUM), where he is leading the Autonomous Vehicle Systems (AVS) lab. He is one of the founders of the TUM Autonomous Motorsport team. His research focuses on developing adaptive dynamic path planning and control algorithms, decision-making algorithms that work under high uncertainty in multi-agent environments, and validating the algorithms on real-world robotic systems. Johannes earned a B.Eng. (2011) from the University of Applied Science Coburg, an M.Sc. (2012) from the University of Bayreuth, an M.A. (2021) in philosophy from TUM, and a Ph.D. (2019) from TUM. 
\end{IEEEbiography}

\end{document}